\documentclass{article} 
\usepackage{iclr2026_conference,times}

\usepackage{amsmath,amsfonts,bm}

\def\eqref#1{equation~\ref{#1}}

\def\1{\bm{1}}

\def\vs{{\bm{s}}}

\DeclareMathAlphabet{\mathsfit}{\encodingdefault}{\sfdefault}{m}{sl}
\SetMathAlphabet{\mathsfit}{bold}{\encodingdefault}{\sfdefault}{bx}{n}

\usepackage{hyperref}
\usepackage{url}

\usepackage{algorithm}

\usepackage{algorithmicx}
\usepackage{algpseudocode}
\usepackage{xspace}
\usepackage{graphicx}   
\usepackage{booktabs}   
\usepackage{array} 
\usepackage[table]{xcolor}
\usepackage{wrapfig}
\usepackage{enumitem}
\usepackage{multirow}
\usepackage{titletoc}
\usepackage{makecell}   
\usepackage{amssymb}

\usepackage{booktabs}
\usepackage{xcolor}
\usepackage{tabularx}

\newcommand{\benchmark}{\mbox{\textsc{RTSGameBench}}\xspace}
\newcommand{\method}{\mbox{\textsc{RTSGameAgent}}\xspace}

\definecolor{cired}{rgb}{0.75,0.1,0.1}
\definecolor{cigreen}{rgb}{0.0,0.45,0.1}

\newcommand{\ci}[3]{$#1^{\textcolor{cired}{#3}}_{\textcolor{cigreen}{#2}}$}

\makeatletter
\DeclareRobustCommand\onedot{\futurelet\@let@token\@onedot}
\def\@onedot{\ifx\@let@token.\else.\null\fi\xspace}

\def\eg{\emph{e.g}\onedot} 
\def\ie{\emph{i.e}\onedot} 
\def\cf{\emph{cf}\onedot} 
 \def\vs{\emph{vs}\onedot}

\makeatother

\usepackage{pifont}

\definecolor{softgreen}{RGB}{51, 200, 51}  
\definecolor{softred}{RGB}{200, 51, 51}    

\newcommand{\cmark}{\textcolor[HTML]{036400}{\ding{52}}}
\newcommand{\xmark}{\textcolor[HTML]{CB0000}{\ding{56}}}
\providecommand{\pmark}{\ensuremath{\triangle}}

\title{\benchmark: Evaluating VLMs for Strategic Reasoning in Game Environments}

\author{%
{\bfseries
San Kim\thanks{Equal contribution.} \quad
Daechul Ahn\footnotemark[1] \quad
Reokyoung Kim
}\\[-0.1em]
{\bfseries
Hyeonbeom Choi \quad
Seungyeon Jwa \quad
Jonghyun Choi\thanks{JC is with ECE, ASRI and IPAI in SNU and a corresponding author.}
}\\[0.4em]
Seoul National University\\[0.2em]
}

\iclrfinalcopy 
\begin{document}

\maketitle
\pagestyle{fancy}
\fancyhf{}

\fancyhead[L]{\small Preprint}
\fancyfoot[C]{\thepage}

\renewcommand{\headrulewidth}{0.4pt}
\setlength{\headheight}{14pt}

\thispagestyle{fancy}

\begin{abstract}
    Vision-Language Models (VLMs) often struggle with \emph{strategic reasoning} in a multi-agent setting, \ie, deciding under uncertainty while coordinating with and adapting to other agents.
    Real-time strategy (RTS) games can be a natural testbed for this capability, requiring long-horizon planning, coordination with allies, and adaptation to opponents under partial observability.
    But existing RTS benchmarks cover limited agent configurations, lack controlled evaluation across distinct strategic demands, and remain confined to fixed scenarios.
    To address these gaps, we propose \benchmark, built on \textit{Beyond All Reason}, an RTS game whose larger scale broadens the strategic space beyond prior testbeds.
    In \benchmark, we offer {holistic} evaluation across diverse full-game match-ups, {diagnostic} mini-games each emphasizing a focal strategic demand, and extensible coverage via a self-evolving framework that turns free-form queries into new mini-games, improving with accumulated experience.
    We further propose \method, an agent harness that combines scalable group-level trajectory planning with short- and long-term memory and outperforms existing agents. 
    We find that state-of-the-art VLMs struggle where humans succeed, particularly when they must coordinate with allies or contend with multiple opponents.
\end{abstract}

\section{Introduction}
\label{sec:introduction}

Vision-Language Models (VLMs) have achieved remarkable success across diverse tasks~\citep{alayrac2022flamingo,liu2023llava}, yet deploying them in complex, evolving environments
requiring long-horizon decisions in the presence of other agents remains challenging~\citep{paglieri2025balrog}.
A key capability in such environments is \emph{strategic reasoning}, which involves deciding under uncertainty while coordinating with and adapting to other agents, both competitors and collaborators~\citep{gandhi2023strategic, zhang2024llm}.
We argue that real-time strategy (RTS) games are a natural testbed as they require long-horizon decision-making under partial observability, including resource allocation, multi-unit coordination, adaptation to opponents, and cooperation with allies, within a measurable and reproducible simulator~\citep{ontanon2013survey, robertson2014review}.

While StarCraft II (SC2) has been widely used as an RTS testbed for AI research~\citep{vinyals2019grandmaster,ma2024large,li2025llmpysc2}, we build upon \textit{Beyond All Reason} (BAR)~\citep{bar2024beyondallreason} featuring larger battles with more units and larger maps than SC2 (Tab.~\ref{tab:sc2_vs_bar_extreme_scale}; see Appendix~\ref{supp:sc2_bar}).
This expanded scale naturally broadens the strategic space, requiring longer-horizon planning over many interacting units, coordination across allied groups, and reasoning about enemy formations~\citep{ontanon2013survey}.
Meanwhile, BAR partially automates low-level execution, from target selection and pathfinding to energy conversion (Appendix~\ref{supp:bar_game_characteristic}), reducing low-level overhead while leaving the agent to decide how to allocate units, when and where to engage, and how to coordinate spatial maneuvers.
Together, BAR's scale and partial low-level automation make it a suitable platform for evaluating VLMs' strategic reasoning in RTS.

\begin{table}[t]
\caption{\textbf{Quantitative comparison of StarCraft II (SC2) \vs \ BAR.}
\textit{Unit Variety}: unique units and buildings across all factions.
\textit{Supply Cap}: per-player unit limit.
\textit{Unit Capacity}: total unit limit across all players.
$^{*}$SC2 uses a weighted population system, so actual unit counts are lower than this.}
\label{tab:sc2_vs_bar_extreme_scale}
\begin{center}
\small
\setlength{\tabcolsep}{2pt}
\begin{tabular*}{\linewidth}{@{\extracolsep{\fill}}lccccc@{}}
\toprule
& Unit Variety & Supply Cap & Unit Capacity & Map Size & Player Limit \\
\midrule
StarCraft II (SC2)      & 96 & $200^{*}$ & $1{,}600^{*}$ & 1$\times$ & 8 \\
Beyond All Reason (BAR) & \textbf{554} & \textbf{2,000} & \textbf{32,000} & \textbf{64$\times$} & \textbf{100} \\
\bottomrule
\end{tabular*}
\end{center}
\vspace{-1.5em}
\end{table}

However, the platform alone does not guarantee a rigorous evaluation.
Strategic reasoning in RTS is inherently multi-faceted~\citep{buro2003real}, spanning resource management, spatial and temporal reasoning, opponent modeling, collaboration, adversarial planning, and decision making under uncertainty.
Moreover, these strategic demands vary with the number, roles, and relationships of agents.
But existing benchmarks cover this space only partially. 
They consider limited agent configurations, lack controlled evaluations across distinct strategic demands, and remain confined to fixed scenarios~\citep{ma2024large,ahn2025hima,li2025llmpysc2} (see Sec.\ref{sec:related} and Appendix~\ref{supp:benchmark_comparison} for detailed comparisons).
To this end, we argue that a rigorous RTS benchmark should be:
(i)~\textit{holistic}, measuring integrated strategic performance in complete games across diverse matchup structures;
(ii)~\textit{diagnostic}, using controlled scenarios to probe different strategic demands;
and (iii)~\textit{extensible}, allowing diagnostic coverage to expand beyond a fixed scenario set via automated, on-demand generation.

\begin{figure}[t]
    \centering
    \includegraphics[width=0.85\linewidth]{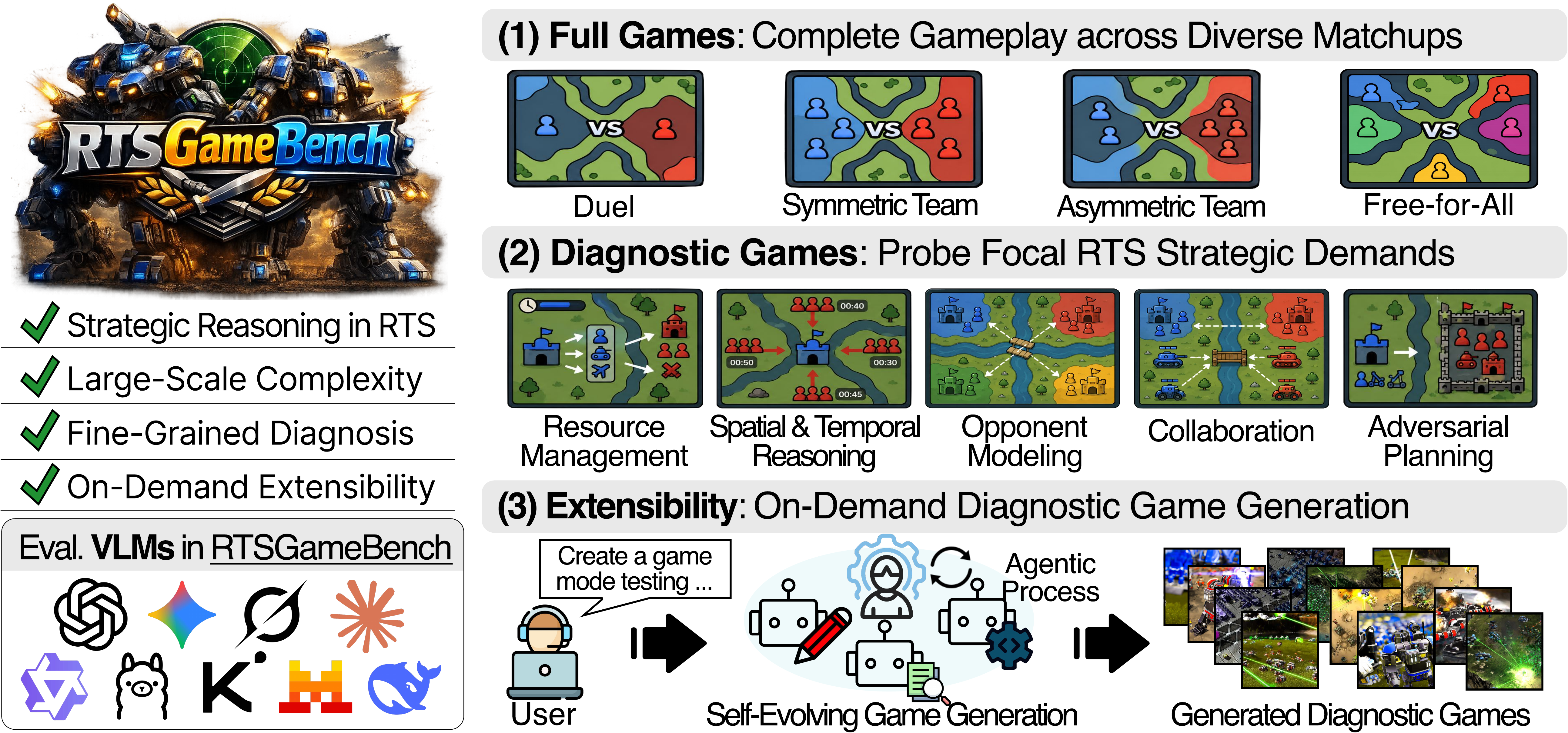}
    \vspace{-1em}
    \caption{\textbf{Overview of \benchmark.} The benchmark comprises (1)~\textit{Full Game Evaluation} across diverse matchups, (2)~\textit{Diagnostic Mini-Games} designed for different focal strategic demands, and (3)~a \textit{Self-Evolving Game Generation Framework} for on-demand diagnostic game generation.}
    \label{fig:overview}
    \vspace{-1.0em}
\end{figure}

To address the desiderata, we propose \textbf{\benchmark}, a benchmark with three components (Fig.~\ref{fig:overview}):
(i)~\textit{Full Games}, evaluating the agent against and alongside built-in AIs across diverse match-ups;
(ii)~\textit{Diagnostic Mini-Games}, controlled scenarios organized around focal strategic demands, grounded in a taxonomy of RTS AI challenges~\citep{buro2003real};
and (iii)~a \textit{Game Generation Framework}, converting free-form user queries into diagnostic mini-games.
As this framework expands diagnostic coverage via repeated generation, treating each generation cycle independently would discard the validated artifacts and experience accumulated in earlier cycles, leading to redundant regeneration and recurring failures.
We therefore introduce \textit{experience-driven evolution}, which makes the framework self-evolving by reusing these artifacts and experiences to progressively improve generation efficiency and quality.

In addition, despite BAR's partial low-level automation, two challenges remain for deploying VLMs at this scale.
Growing combat forces make per-unit control increasingly costly, a challenge compounded by sparse VLM calls that limit how often unit behavior can be revised.
Meanwhile, long-horizon play requires integrating recent events with longer-term context.
To address these, we propose \textbf{\method}, a baseline VLM agent harness that organizes combat units into groups and plans trajectories of group commands to sustain execution between VLM calls while maintaining temporal context via short- and long-term memory.
\method outperforms competing agents under matched settings.
With it, we find state-of-the-art (SoTA) VLMs struggle where humans succeed, especially when coordinating with allies or contending with multiple opponents, and mini-games suggest that their relative strengths vary across strategic demands.
Finally, experience-driven evolution improves generated games while reducing token usage.


\section{Related Work}
\label{sec:related}

\vspace{-0.5em}\paragraph{Game-based evaluation of LLM and VLM agents.}
Games serve as effective testbeds for evaluating the decision-making capabilities of LLM- and VLM-based agents~\citep{paglieri2025balrog, hu2025lmgame, park2025orak}.
Early benchmarks relied on textual observations in single-agent~\citep{hu2024pokellmon} or multi-agent strategic settings~\citep{qi2024civrealm}, while multimodal benchmarks such as EscapeCraft~\citep{wang2025escapecraft} and MCU~\citep{zheng2025mcu} remain largely single-agent.
As full-game outcomes can obscure the sources of success or failure~\citep{lin2025gamebot}, recent studies adopt scenario-level evaluation~\citep{tang2025dsgbench, zheng2025v} or separate strategic reasoning from decision-making~\citep{xu2025vs}, mostly over predefined scenarios.
Scenario generation has been explored for single-agent tasks~\citep{zheng2025mcu} and text-based two-player games~\citep{verma2025ggbench}, but without targeting strategic demands specified by users.
\benchmark instead evaluates VLM agents in a large-scale multi-agent RTS environment and extends its diagnostic mini-games on demand from free-form user queries.

\vspace{-0.5em}\paragraph{RTS game benchmarks.}
Most RTS benchmarks for LLM-based agents are built on SC2: TextStarCraft~II~\citep{ma2024large}, \textsc{TextSCII-All}~\citep{ahn2025hima}, and HIVE~\citep{anne2025harnessing} use textual observations, while AVACraft~\citep{ma2025ava}, LLM-PySC2~\citep{li2025llmpysc2}, and \citet{kim2026multimodal} introduce multimodal scenario- or full-game evaluation.
Most closely related, \citet{li2026winning} introduce a diagnostic long-horizon RTS benchmark for LLM agents in Red Alert.
\benchmark differs by evaluating multimodal VLMs across diverse multiplayer match-ups and taxonomy-grounded strategic demands, combining complete games with targeted diagnostics while enabling on-demand diagnostic game generation beyond a fixed scenario set; building on BAR also supports substantially larger-scale evaluation (see Appendix~\ref{supp:benchmark_comparison} for a detailed comparison).

\vspace{-0.5em}\paragraph{LLM-based game and environment generation.}
LLMs have automated game development with role-specialized multi-agent frameworks~\citep{chen2023gamegpt} and generated executable environments for open-ended learning~\citep{faldor2025omniepic} or cross-environment agent learning~\citep{autoenv2025}.
Our framework similarly builds on role-specialized agents~\citep{hong2024metagpt,chen2023gamegpt}, but targets diagnostic evaluation, designing each game around a focal strategic demand in a complex RTS engine and validating it via staged rubric checks and simulation tests.
It also evolves across generations. 
Unlike archives that seed novel tasks~\citep{faldor2025omniepic} or agents that self-improve on the tasks they solve~\citep{shinn2023reflexion,zhao2024expel}, it reuses validated designs and rule code to skip redundant stages and refines its rubrics through retrospectives.



\section{\benchmark}
\label{sec:rtsbench}

As discussed in Sec.~\ref{sec:introduction}, rigorous evaluation of strategic reasoning in RTS requires a holistic, diagnostic, and extensible benchmark, which existing benchmarks partially meet.
To address this, we introduce \benchmark, built on BAR, with three components (Fig.~\ref{fig:overview}): (i)~\textit{Full Games} (Sec.~\ref{sec:fullgame_desc}), (ii)~\textit{Diagnostic Mini-Games} (Sec.~\ref{sec:minigame_bench}), and (iii)~\textit{Game Generation Framework} (Sec.~\ref{sec:minigame_gen}).

\vspace{-0.5em}\paragraph{Interface.}
All inferences follow an observe--decide--act loop.
The agent uses game knowledge~$\mathcal{K}$, including the scenario description, available units and buildings, and team configuration.
At each interval~$t$, visual observations~$v_t$ (global minimap and local camera views) are captured from game state~$s_t$, while a Python wrapper~$\mathcal{W}$ extracts textual state~$\mathcal{W}(s_t)$ to form multimodal observation~$o_t$.
With Fog-of-War (FoW)\footnote{A game mechanic that hides areas outside allied units' line of sight, introducing partial observability.} enabled, the observation~$o_t$ is restricted to allied line of sight.
Given~$\mathcal{K}$ and~$o_t$, the agent's VLM policy~$\pi$ produces action~$a_t$, which updates the environment state:
\begin{equation}
    o_t = (v_t, \mathcal{W}(s_t)), \quad
    a_t = \pi(o_t \mid \mathcal{K}), \quad
    s_{t+1} \leftarrow \text{Env}(s_t, a_t).
    \label{eq:rtsbench_policy}
    \vspace{-0.2em}
\end{equation}
The action space comprises \textit{building construction}, \textit{unit production}, and \textit{unit action}.
The agent decides which building produces each unit, which unit constructs each building, and where to build or move on a $(0,0)$--$(100,100)$ grid, while~$\text{Env}$ handles low-level execution.
The game runs for a fixed interval between decisions and pauses during inference, so evaluation reflects strategic decision quality rather than reaction speed.
Full interface and~$\mathcal{K}$ details are provided in Appendix~\ref{supp:interface}.


\begin{table}[t]
\vspace{-1.0em}
\caption{\textbf{Evaluation settings in \benchmark.}
\textit{Top:} Full-game matchups.
\textit{Bottom:} Diagnostic mini-games designed to make different focal strategic demands particularly salient~\citep{buro2003real}.
Action types (Sec.\ref{sec:rtsbench}): Build = building construction, Prod. = unit production, Action = unit action.}
\label{tab:eval_overview}
\vspace{-1.0em}

\begin{center}
\resizebox{\linewidth}{!}{%
\begin{tabular}{llllc}
\toprule
\multicolumn{5}{l}{\textit{Full Game Match-ups}} \\
\midrule
\textbf{Mode} & \textbf{Config} & \textbf{Strategic Demand} & \textbf{Action Type} & \textbf{FoW} \\
\cmidrule(lr){1-5}
Duel            & 1v1     & Individual decision-making                 & Build + Prod. + Action & On \\
Symmetric Team  & 2v2     & Allied coordination                        & Build + Prod. + Action & On \\
Asymmetric Team & 3v4     & Coordination under numerical disadvantage & Build + Prod. + Action & On \\
Free-for-All    & 1v1v1v1 & Multi-polar threat prioritization          & Build + Prod. + Action & On \\
\midrule
\multicolumn{5}{l}{\textit{Diagnostic Mini-Games}} \\
\midrule
\textbf{Focal Strategic Demand} & \textbf{Game} & \textbf{Task} & \textbf{Action Type} & \textbf{FoW} \\
\cmidrule(lr){1-5}
Resource Management (RM) & TCP & Meet production and economy targets & Build + Prod. + Action & On \\
Spatial\&Temporal Reasoning (S\&T) & MFD & Defend multiple objectives from staggered attacks & Action & Off \\
Opponent Modeling (OM) & FBA & Analyze and destroy a fortified enemy base & Build + Prod. + Action & On \\
Collaboration (COL) & ATS & Form a team with allied forces and fight cooperatively & Prod. + Action & Off \\
Adversarial Planning (AP) & AWD & Survive enemy waves whose composition is known in advance & Build + Prod. + Action & Off \\

\bottomrule
\end{tabular}%
}
\end{center}
\vspace{-2em}
\end{table}

\subsection{Full Game Match-ups}
\label{sec:fullgame_desc}

In full-game evaluation, an agent plays complete BAR matches.
While existing RTS benchmarks mainly focus on 1v1 settings~\citep{ma2024large,li2025llmpysc2}, different player configurations impose distinct competitive and cooperative demands~\citep{buro2003real}.
We thus evaluate four match-ups (Tab.~\ref{tab:eval_overview}, top): \textit{Duel} tests individual capability, \textit{Symmetric Team} adds allied coordination, \textit{Asymmetric Team} requires coordination from the smaller team, and \textit{Free-for-All} requires prioritizing threats among multiple opponents.
In all settings, the evaluated agent occupies one slot and all remaining slots are filled by built-in AIs at fixed difficulty levels.


\subsection{Diagnostic Mini-Games}
\label{sec:minigame_bench}

\begin{wrapfigure}{r}{0.28\linewidth}
\vspace{-1.0em}
\centering
\includegraphics[width=\linewidth]{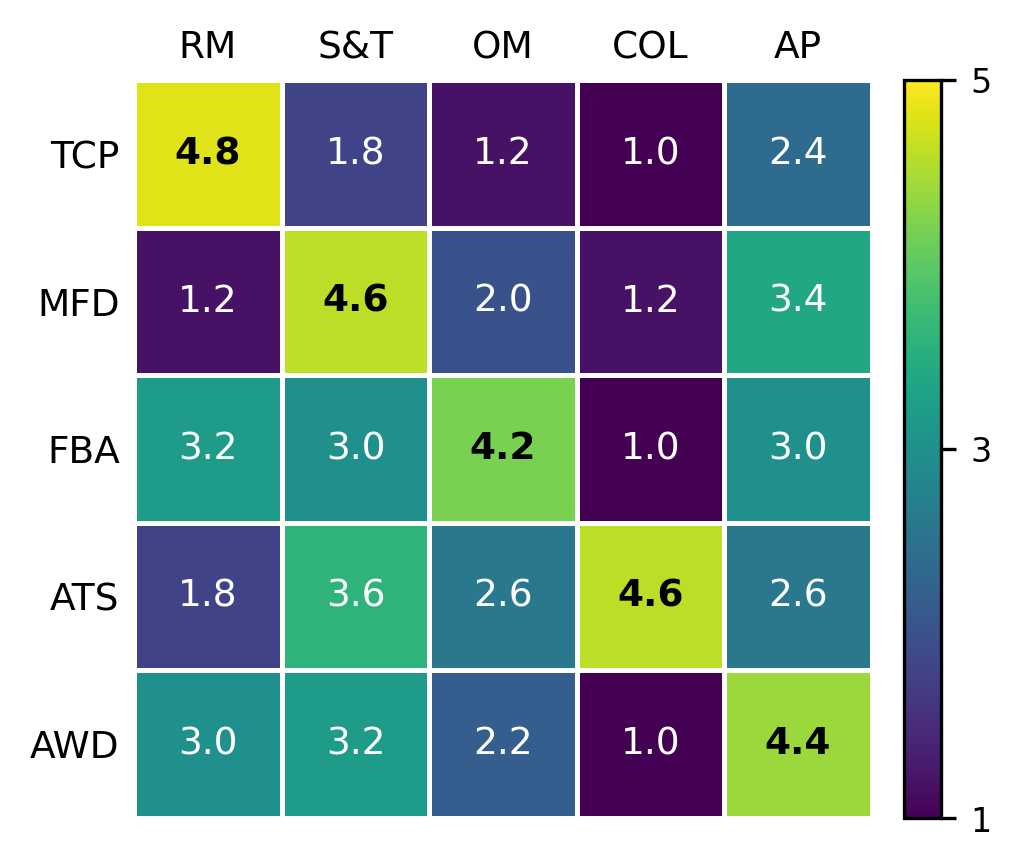}
\vspace{-1.5em}
\caption{\textbf{Strategic demands of each mini-game.} Mean human ratings (1--5).}
\label{fig:minigame_demands}
\vspace{-1.0em}
\end{wrapfigure}

Full-game evaluation requires multiple strategic demands simultaneously, making it difficult to attribute performance patterns to any particular demand.
We thus introduce controlled diagnostic mini-games (Tab.~\ref{tab:eval_overview}, bottom), each organized around a focal demand from a taxonomy of RTS AI challenges~\citep{buro2003real}, at three difficulty levels.
Rather than strictly isolating capabilities, these scenarios make one demand particularly salient while preserving the coupled decision-making inherent to RTS gameplay.
Human ratings support this, as each scores highest on its focal demand while still eliciting others (Fig.~\ref{fig:minigame_demands}; Appendix~\ref{supp:demand_ratings}).
We thus interpret them as targeted probes of behavior under particular demands, rather than isolated measurements of latent capabilities, providing a finer-grained characterization of model behavior than full-game outcomes alone, as follows:

\vspace{-0.3em}
\noindent(1)~\textit{Resource Management: Time-Constrained Production (TCP).}
The agent must meet economy and production targets within a time limit.
Production dependencies~\citep{bar2024qualitipedia} force sequential build decisions, testing how it balances economic growth against unit production.

\vspace{-0.3em}
\noindent(2)~\textit{Spatial \& Temporal Reasoning: Multi-Front Defense (MFD).}
With fixed forces and no production, the agent defends multiple buildings against attacks arriving from different directions at staggered times, placing emphasis on maneuvering forces across fronts as threats emerge.

\vspace{-0.3em}
\noindent(3)~\textit{Opponent Modeling: Fortified Base Assault (FBA).}
The agent must destroy a key structure inside a fortified enemy base whose defenses are revealed only via scouting, placing emphasis on inferring the enemy's defensive setup and adapting forces and approach accordingly.

\vspace{-0.3em}
\noindent(4)~\textit{Collaboration: Allied Team Skirmish (ATS).}
With building construction disabled, the agent fights alongside a built-in AI ally in a 2v2 battle, emphasizing coordination with this ally; the agent wins if its team's score, combining damage dealt and forces preserved, exceeds that of the opposing team.

\vspace{-0.3em}
\noindent(5)~\textit{Adversarial Planning: Anticipatory Wave Defense (AWD).}
The agent must protect its commander from successive waves whose unit compositions are known in advance, emphasizing anticipatory counter-planning for future enemy attacks.

Another challenge in the taxonomy, \textit{Decision Making under Uncertainty}, is incorporated selectively via FoW (Tab.~\ref{tab:eval_overview}, FoW column).
See Appendix~\ref{supp:minigames} for mini-game details and difficulty levels.


\subsection{Game Generation Framework}
\label{sec:minigame_gen}

Each RTS strategic demand can arise in diverse scenarios that a fixed mini-game set cannot fully cover.
Expanding this coverage manually is costly, as game creation spans stages such as scenario design, engine-level implementation, and simulation-based validation, each requiring distinct expertise~\citep{hendrikx2013procedural}; this motivates generating diagnostic games automatically and on demand.
Moreover, experience from earlier generation cycles can be reused to avoid redundant effort and recurring failures.
With both in mind, we propose a \textit{self-evolving} game generation framework that builds mini-games from free-form queries and progressively improves its efficiency and quality through accumulated experience (Fig.~\ref{fig:minigame_gen}; Alg.~\ref{alg:self_evolving} in Appendix~\ref{supp:self_evolving_algo}).

\vspace{-0.5em}\paragraph{Generation pipeline.}
Handling all stages in a single agent would mix heterogeneous contexts~\citep{liu2024lost,shi2023large} and obscure where failures originate.
We therefore take a \emph{divide-and-conquer} approach with a \emph{role-specialized multi-agent system}~\citep{hong2024metagpt}: a \textit{project manager} (PM) for orchestration, a \textit{designer} for game design, a \textit{developer} for implementation, and an \textit{analyst} for validation.
The PM coordinates four stages, rolling back to the stage responsible for each validation failure and aborting once a fixed rollback budget is exhausted (Appendix~\ref{supp:self_evolving_algo}).
Before each validation, the PM retrieves relevant \emph{validation rubrics} from a shared knowledge database.
In \textit{Stage~1}, the designer interacts with the user to produce a scenario brief specifying game composition, enemy behavior, and win/loss conditions; if the database contains a matching game design document (GDD)~\citep{fullerton2014game} and rule set, the pipeline skips to \textit{Stage~4}.
In \textit{Stage~2}, the designer expands the brief into a GDD defining the focal demand, game rules, and a checklist for the final game, and the analyst validates it.
In \textit{Stage~3}, the PM retrieves stored rules, the developer writes the rest in Lua~\citep{ierusalimschy2006lua} with rule-based simulation tests, and the analyst validates the rule set against the tests and rubrics.
In \textit{Stage~4}, the developer builds an executable game script with tests, which the analyst validates against the tests, rubrics, and GDD checklist.
The validated script is then scaled to three difficulty levels, using the mini-games in Sec.\ref{sec:minigame_bench} as references.

\vspace{-0.5em}\paragraph{Experience-driven evolution.}
Rather than treating each generation independently, the framework accumulates and reuses experience in two ways (validated in Sec.\ref{sec:eval_self-evol}).
First, validated GDDs and rule sets are stored in the shared knowledge database and reused when applicable, bypassing redundant stages.
Second, after each successful generation, the PM performs a \textit{retrospective analysis} of its generation history, including failed attempts, to update the rubrics.
However, because generated games vary widely in objectives and mechanics, a single flat rubric set either remains too general~\citep{jonsson2007use} or grows overly complex as new criteria accumulate.
We therefore organize each stage's rubrics in a \emph{two-level hierarchy} that groups individual rubrics under broad categories (Fig.~\ref{fig:minigame_gen}, right).
The PM retrieves and updates only the rubrics relevant to each game, so validation experience accumulates within each category instead of growing a monolithic rubric.

\begin{figure}[t]
\centering
\includegraphics[width=1\linewidth]{figures/fig2_pipeline_v3.pdf}
\vspace{-1.5em}
\caption{\textbf{Self-evolving game generation.}
The PM coordinates four stages with rubric retrieval, retries, and rollback, while shared experience enables reuse and rubric updates across generations.}
\label{fig:minigame_gen}
\vspace{-1.0em}
\end{figure}



\section{\method}
\label{sec:rtsagent}
Deploying VLM agents at BAR's scale poses two challenges.
First, \emph{unit action} requires per-unit decisions that grow with the number of combat units, while sparse VLM calls limit how often unit behavior is revised.
Second, long-horizon play requires integrating recent events with historical context~\citep{park2023generative,packer2023memgpt}.
To address both, we introduce \method, a baseline agent harness (Fig.~\ref{fig:gameagent}) combining group-level control with short- and long-term memory.

\vspace{-0.5em}\paragraph{Scalable group-level control.}
To scale control with the number of combat units, we decompose \emph{unit action} into two group-level action types, \textit{group assignment} and \textit{group command} (Fig.~\ref{fig:gameagent}).
With group assignment, the VLM dynamically organizes combat units into groups according to the current strategy and game state, unlike prior approaches that rely on predefined combat groups~\citep{li2025llmpysc2}.
For each group, a \textit{group command} specifies one of three modes, \texttt{move}, \texttt{move\_force}, or \texttt{stop}, together with a target coordinate when needed.
Upon enemy contact, a group automatically engages (\texttt{fight}) and then resumes its command, except under \texttt{move\_force}, which continues toward the target.
Together, group assignment and group commands enable the VLM to express diverse group-level maneuvers within a compact control space.

To bridge sparse VLM calls, we further introduce \textit{trajectory planning} over group commands.
Instead of a single command, the VLM outputs for each group a condition-dependent command trajectory, represented as a directed acyclic graph (DAG), to sustain adaptive execution between calls.
Transitions between nodes are triggered by predefined conditions on the group's own execution state; when a condition is satisfied, the corresponding follow-up command is activated without another VLM call.
Because transitions are group-local, changes beyond a group's own state, including other groups' progress, are handled only at VLM calls.
For critical changes requiring an immediate response, predefined triggers therefore initiate \textit{replanning} before the next scheduled call, invoking the VLM with the updated game state to generate new trajectories.
Detailed transition conditions and replanning triggers are provided in Appendix~\ref{supp:agent_trajectory}.

\begin{figure}[t]
    \centering
    \includegraphics[width=1\linewidth]{figures/fig4_rts_agent.pdf}
    \vspace{-1em}
    \caption{\textbf{Inference loop of \method.} The \textit{memory phase} (left) consolidates short-term logs~$\mathcal{S}_t$ and long-term memory~$\mathcal{L}_{t-1}$ into context~$m_t$ and updated memory~$\mathcal{L}_t$; the \textit{decision phase} (right) feeds~$m_t$, observation~$o_t$, and game knowledge~$\mathcal{K}$ to the VLM policy~$\pi$ to output action~$a_t$.}
    \label{fig:gameagent}
    \vspace{-1.2em}
\end{figure}

\vspace{-0.5em}\paragraph{Memory-augmented inference.}
To integrate recent events with historical context, we augment each inference with short- and long-term memory, following their cognitive distinction~\citep{atkinson1968human,squire1992memory}.
\textit{Short-term memory}~$\mathcal{S}_t$ holds event-triggered logs collected between VLM calls, such as enemy sightings within allied line of sight and battle events, whereas \textit{long-term memory}~$\mathcal{L}_t$ persists throughout the game, summarizing game progress and past battles, such as the resulting changes in allied and enemy forces.
At each VLM call~$t$, a memory phase and a decision phase run in sequence (Fig.~\ref{fig:gameagent}).
In the \textit{memory phase}, the policy VLM, prompted with text only, consolidates $\mathcal{S}_t$ and $\mathcal{L}_{t-1}$ into $\mathcal{L}_t$ and extracts decision-relevant context~$m_t$, \ie, $(m_t,\mathcal{L}_t)=\mathrm{VLM}(\mathcal{S}_t,\mathcal{L}_{t-1})$, after which $\mathcal{S}_t$ is cleared.
In the \textit{decision phase}, the VLM receives $m_t$ with the observation~$o_t$, whose local camera views are centered on strategically relevant regions selected by heuristics.
The base policy in Eq.~\ref{eq:rtsbench_policy} thus becomes $a_t=\pi(o_t,m_t\mid\mathcal{K})$, where $a_t$ comprises building construction, unit production, group assignment, and group-command trajectories.
Appendix~\ref{supp:agent_algo} details the inference loop, camera-view heuristics, and event-logging conditions.

\section{Experiments}
\label{sec:exp}

\vspace{-0.5em}\paragraph{Evaluation protocol.}
On full- and mini-games (Tabs.~\ref{tab:main_results} and~\ref{tab:minigame_results}), we evaluate three agent baselines~\citep{ma2024large,shao2024swarmbrain,li2025llmpysc2} adapted to BAR (Appendix~\ref{supp:adaptation_agents}), proprietary and open-weight VLMs, and a human baseline, all under the same interface and observations.
Following \citet{li2025llmpysc2}, we use three difficulty levels (Lv.1--3; Details in Appendices~\ref{supp:full_game} and~\ref{supp:minigames}).
Each setting is run 30 times (10 for humans), and Elo is computed from 20 round-robin matches per pair.
The inference interval is 1\,min of game time for full games, TCP, and FBA, and 3\,s for MFD, ATS, and AWD (Appendix~\ref{supp:interval}).
Self-evolving games are generated with Gemini-3.8-Flash.

\vspace{-0.5em}\paragraph{Metrics.}
We primarily use win rate (WR) for full- and mini-game evaluation.
As WR reflects only the final outcome, we additionally report six metrics in four dimensions to analyze how agents play:
\emph{resources} (gathered, RG; spent, RS),
\emph{army} (strength, AS; unit quality, UQ),
\emph{battle} (enemies destroyed, ED),
and \emph{visibility} (vision coverage, VC).
Each metric is the agent's share relative to other players, scaled so that 0.5 denotes an even share.
AS measures the total value of combat units an agent maintains, whereas UQ measures the average value per produced combat unit, reflecting how advanced its composition is (Appendix~\ref{supp:metrics}).
To evaluate generated games (Sec.\ref{sec:eval_self-evol}), we use four criteria, detailed in Appendix~\ref{supp:eval_generated_games}: \emph{playability}, the fraction of generation attempts that yield a launchable game; \emph{behavioral fidelity}, the fraction of expected behaviors, derived from the query and GDD, that are observed in play; \emph{token usage} per generation; and pairwise \emph{human preference}.

\begin{table}[t]
\vspace{-1.0em}
\caption{\textbf{Full-game evaluation.} WR at built-in AIs (Lv.1--3). The evaluated agent, or a human using the same interface, occupies one slot; others are built-in AIs. Best in \textbf{bold}; second best \underline{underlined}.}
\vspace{-1em}
\label{tab:main_results}
\begin{center}
\newcolumntype{C}{>{\centering\arraybackslash}p{2.3em}}
\resizebox{\linewidth}{!}{%
\begin{tabular}{l *{12}{C}}
\toprule
\multirow{2.5}{*}{\textbf{Model}}
& \multicolumn{3}{c}{\textit{\textbf{Duel} (1v1)}}
& \multicolumn{3}{c}{\textit{\textbf{Symmetric Team} (2v2)}}
& \multicolumn{3}{c}{\textit{\textbf{Asymm. Team} (3v4)}}
& \multicolumn{3}{c}{\textit{\textbf{Free-for-All} (1v1v1v1)}} \\
\cmidrule(lr){2-4}
\cmidrule(lr){5-7}
\cmidrule(lr){8-10}
\cmidrule(lr){11-13}

& Lv.1 & Lv.2 & Lv.3
& Lv.1 & Lv.2 & Lv.3
& Lv.1 & Lv.2 & Lv.3
& Lv.1 & Lv.2 & Lv.3 \\

\midrule
\multicolumn{13}{l}{\textit{Agent Comparison (Backbone = Gemini-3.7-Flash)}} \\
\midrule

CoS~\citep{ma2024large}
    & 0.17 & 0.00 & 0.00
    & 0.20 & 0.07 & 0.00
    & 0.00 & 0.00 & 0.00
    & 0.00 & 0.00 & 0.00 \\

LLM-PySC2~\citep{li2025llmpysc2}
    & 0.43 & 0.00 & 0.00
    & 0.23 & 0.13 & \underline{0.10}
    & 0.00 & 0.00 & 0.00
    & 0.00 & 0.00 & 0.00 \\

SwarmBrain~\citep{shao2024swarmbrain}
    & \underline{0.80} & \underline{0.23} & \underline{0.13}
    & \underline{0.57} & \underline{0.33} & \textbf{0.20}
    & 0.00 & 0.00 & 0.00
    & 0.00 & 0.00 & 0.00 \\

\rowcolor{gray!12}
\textsc{RTSGameAgent} (\textbf{Ours})
    & \textbf{1.00} & \textbf{0.43} & \textbf{0.30}
    & \textbf{0.83} & \textbf{0.40} & \textbf{0.20}
    & \textbf{0.17} & 0.00 & 0.00
    & \textbf{0.20} & \textbf{0.07} & 0.00 \\

\midrule
\multicolumn{13}{l}{\textit{VLM Backbone Comparison (Agent = \textsc{RTSGameAgent})}} \\
\midrule

GPT-5.6-Luna
    & 0.20 & 0.00 & 0.00
    & 0.13 & 0.10 & 0.00
    & 0.00 & 0.00 & 0.00
    & 0.00 & 0.00 & 0.00 \\

Qwen3.8-27B
    & 0.23 & 0.00 & 0.00
    & 0.33 & 0.00 & 0.00
    & 0.00 & 0.00 & 0.00
    & 0.00 & 0.00 & 0.00 \\

Qwen3.5-397B
    & 0.47 & 0.00 & 0.00
    & 0.10 & 0.07 & 0.03
    & 0.00 & 0.00 & 0.00
    & 0.00 & 0.00 & 0.00 \\

Muse-Spark-1.2
    & 0.40 & 0.00 & 0.00
    & 0.27 & \underline{0.20} & \underline{0.07}
    & 0.00 & 0.00 & 0.00
    & 0.00 & 0.00 & 0.00 \\

GPT-5.6-Terra
    & 0.57 & 0.07 & 0.00
    & 0.30 & 0.03 & \underline{0.07}
    & 0.00 & 0.00 & 0.00
    & \underline{0.13} & \textbf{0.07} & 0.00 \\

GLM-5.3-Flash
    & 0.67 & \underline{0.13} & \underline{0.10}
    & \underline{0.40} & 0.13 & \underline{0.07}
    & 0.00 & 0.00 & 0.00
    & 0.03 & \underline{0.03} & 0.00 \\

Grok-4.6
    & \underline{0.73} & 0.00 & 0.00
    & \textbf{0.83} & 0.10 & 0.00
    & 0.00 & 0.00 & 0.00
    & 0.07 & 0.00 & 0.00 \\

Gemini-3.7-Flash
    & \textbf{1.00} & \textbf{0.43} & \textbf{0.30}
    & \textbf{0.83} & \textbf{0.40} & \textbf{0.20}
    & \textbf{0.17} & 0.00 & 0.00
    & \textbf{0.20} & \textbf{0.07} & 0.00 \\

\midrule
Human
    & 1.00 & 1.00 & 1.00
    & 1.00 & 1.00 & 1.00
    & 1.00 & 1.00 & 0.80
    & 1.00 & 1.00 & 0.70 \\

\bottomrule
\end{tabular}%
}
\end{center}
\vspace{-1em}
\end{table}

\begin{table}[t]
\vspace{-1.0em}
\caption{\textbf{Mini-game evaluation.} WR over 30 runs per setting across Lv.1--3. CoS is evaluated only on TCP, as its heuristic army-wide action lacks the spatial unit control required by the other mini-games. Best in \textbf{bold}; second best \underline{underlined}.}
\vspace{-1em}
\label{tab:minigame_results}
\begin{center}
\resizebox{\linewidth}{!}{%
\begin{tabular}{lccccccccccccccc}
\toprule
\multirow{2.5}{*}{\textbf{Model}}
& \multicolumn{3}{c}{\textbf{TCP}}
& \multicolumn{3}{c}{\textbf{MFD}}
& \multicolumn{3}{c}{\textbf{FBA}}
& \multicolumn{3}{c}{\textbf{ATS}}
& \multicolumn{3}{c}{\textbf{AWD}} \\
\cmidrule(r){2-4}
\cmidrule(lr){5-7}
\cmidrule(lr){8-10}
\cmidrule(lr){11-13}
\cmidrule(l){14-16}

& Lv.1 & Lv.2 & Lv.3
& Lv.1 & Lv.2 & Lv.3
& Lv.1 & Lv.2 & Lv.3
& Lv.1 & Lv.2 & Lv.3
& Lv.1 & Lv.2 & Lv.3 \\

\midrule
\multicolumn{16}{l}{\textit{Agent Comparison (Backbone = Gemini-3.7-Flash)}} \\
\midrule

CoS~\citep{ma2024large}
  & 0.63 & 0.10 & 0.00
  & -- & -- & --
  & -- & -- & --
  & -- & -- & --
  & -- & -- & -- \\

LLM-PySC2~\citep{li2025llmpysc2}
  & \underline{0.93} & \underline{0.43} & 0.03
  & 0.13 & \underline{0.10} & 0.00
  & 0.73 & 0.10 & 0.00
  & 0.37 & 0.30 & \underline{0.10}
  & \underline{0.27} & 0.00 & 0.00 \\

SwarmBrain~\citep{shao2024swarmbrain}
  & \textbf{0.97} & 0.40 & \underline{0.07}
  & \underline{0.67} & 0.00 & 0.00
  & \textbf{1.00} & \underline{0.33} & \textbf{0.30}
  & \underline{0.87} & \textbf{0.43} & \textbf{0.13}
  & 0.10 & 0.00 & 0.00 \\

\rowcolor{gray!12}
\method (\textbf{Ours})
  & \underline{0.93} & \textbf{0.67} & \textbf{0.20}
  & \textbf{0.93} & \textbf{0.60} & \textbf{0.37}
  & \underline{0.90} & \textbf{0.47} & \underline{0.20}
  & \textbf{1.00} & \underline{0.40} & \textbf{0.13}
  & \textbf{0.77} & \textbf{0.43} & \textbf{0.13} \\

\midrule
\multicolumn{16}{l}{\textit{VLM Backbone Comparison (Agent = \textsc{RTSGameAgent})}} \\
\midrule

GPT-5.6-Luna
  & 0.47 & 0.07 & 0.00
  & 0.50 & 0.10 & 0.03
  & 0.83 & 0.40 & 0.00
  & 0.50 & 0.27 & 0.00
  & 0.00 & 0.00 & 0.00 \\

Qwen3.8-27B
  & 0.60 & 0.00 & 0.00
  & 0.63 & 0.13 & 0.00
  & 0.83 & 0.03 & 0.10
  & 0.20 & 0.17 & 0.00
  & 0.20 & 0.03 & 0.00 \\

Qwen3.5-397B
  & 0.53 & 0.17 & 0.00
  & 0.67 & 0.37 & 0.13
  & 0.87 & 0.20 & 0.10
  & 0.27 & 0.13 & 0.07
  & 0.03 & 0.00 & 0.00 \\

Muse-Spark-1.2
  & \textbf{1.00} & 0.13 & 0.07
  & 0.73 & \textbf{0.60} & 0.00
  & 0.80 & 0.20 & 0.00
  & 0.27 & 0.17 & \underline{0.10}
  & 0.23 & \underline{0.10} & 0.00 \\

GPT-5.6-Terra
  & 0.73 & 0.20 & 0.00
  & 0.37 & 0.23 & \underline{0.17}
  & \underline{0.90} & 0.37 & \underline{0.33}
  & 0.30 & 0.20 & 0.07
  & 0.13 & 0.00 & 0.00 \\

GLM-5.3-Flash
  & \underline{0.93} & \underline{0.77} & 0.07
  & \underline{0.83} & \underline{0.47} & 0.00
  & \textbf{1.00} & 0.37 & \textbf{0.40}
  & \underline{0.53} & \underline{0.30} & 0.00
  & \underline{0.27} & 0.00 & 0.00 \\

Grok-4.6
  & \textbf{1.00} & \textbf{0.83} & \textbf{0.23}
  & 0.80 & 0.13 & 0.00
  & \textbf{1.00} & \textbf{0.57} & 0.30
  & 0.47 & 0.23 & 0.07
  & 0.23 & \underline{0.10} & 0.00 \\

Gemini-3.7-Flash
  & \underline{0.93} & 0.67 & \underline{0.20}
  & \textbf{0.93} & \textbf{0.60} & \textbf{0.37}
  & \underline{0.90} & \underline{0.47} & 0.20
  & \textbf{1.00} & \textbf{0.40} & \textbf{0.13}
  & \textbf{0.77} & \textbf{0.43} & \textbf{0.13} \\

\midrule
Human
  & 1.00 & 1.00 & 0.60
  & 1.00 & 1.00 & 0.60
  & 1.00 & 1.00 & 0.70
  & 1.00 & 0.80 & 0.70
  & 1.00 & 0.80 & 0.60 \\

\bottomrule
\end{tabular}%
}
\end{center}
\vspace{-1.5em}
\end{table}

\subsection{Results}
\label{sec:exp_quant}

\paragraph{Full game evaluation.}
Tab.~\ref{tab:main_results} reports full-game results (90\% confidence intervals in Appendix~\ref{supp:fullgame_ci}).
For both agents and backbones, WR declines as the built-in AI level increases and drops sharply in the most interaction-demanding match-ups.
Gemini-3.7-Flash matches or exceeds all other backbones in every setting, with Grok-4.6 and GLM-5.3-Flash competitive only in selected settings, so we use it for the agent comparison, which shows similar trends with other backbones (Appendix Tab.~\ref{tab:agent_backbones}).
\method matches or outperforms the baselines in every setting and is the only agent to win any Asymmetric Team or Free-for-All game, supporting the benefit of its components (Sec.\ref{sec:detailed_anal}).
Nevertheless, no VLM agent exceeds 0.20 WR in these match-ups, even at Lv.1, where success requires tight ally coordination under numerical disadvantage and reasoning over multiple opponents, respectively.
Humans under the same interface reach at least 0.70, indicating limitations of current VLM agents rather than infeasible settings.

\vspace{-0.5em}\paragraph{Diagnostic mini-game evaluation.}
Tab.~\ref{tab:minigame_results} reports mini-game results (90\% confidence intervals in Appendix~\ref{supp:minigame_ci}).
Among agents, \method ranks first or second in every setting, with the largest margins on MFD and AWD; as both require committing forces against successive attacks, we conjecture that these margins stem from group-level control and the components built on it, the most critical component in our ablation (Sec.\ref{sec:detailed_anal}).
Across backbones, although Gemini-3.7-Flash leads in full games, Grok-4.6 and GLM-5.3-Flash mostly match or exceed it on TCP and FBA but fall behind on MFD, ATS, and AWD at every level.
We conjecture that they are at least as strong at economic throughput and scouting, central to TCP and FBA, but weaker at handling forces as engagements unfold, central to the other three, a difference less visible in full-game results.
ATS results are also consistent with those in Symmetric Team full games (Tab.~\ref{tab:main_results}), which likewise require allied coordination, with Gemini-3.7-Flash best or tied for best in both at every level; while not a one-to-one measure of collaboration, this agreement suggests that ATS captures behavior relevant to coordination.
Humans under the same interface achieve near-perfect WR at Lv.1--2 and 0.60--0.70 at Lv.3, whereas no model exceeds 0.23 on TCP, ATS, and AWD at Lv.3; the largest gaps, on ATS and AWD, suggest that collaboration and anticipatory planning may be common weaknesses of current VLM agents, consistent with their failures in team full games.

\begin{figure*}[t]
    \centering
    \includegraphics[width=0.75\linewidth]{figures/fig5_detail_anal.pdf}
    \vspace{-1em}
    \caption{\textbf{Metric analysis of full-game performance.} Evaluated against Lv.1 built-in AIs.}
    \label{fig:metric_analysis}
    \vspace{-1.5em}
\end{figure*}

\begin{table}[t]
\centering
\begin{minipage}[t]{0.28\linewidth}
\caption{\textbf{Head-to-head Elo} from duels among Gemini-3.7-Flash-based agents.}
\vspace{0.2em}
\label{tab:elo}
\end{minipage}\hfill
\begin{minipage}[t]{0.69\linewidth}
\caption{\textbf{Ablation study of \method.} We ablate each component and the visual observation, and report WR and detailed gameplay metrics in full-game duels; higher is better for all metrics.}
\vspace{0.2em}
\label{tab:ablation}
\end{minipage}
\par\vspace{2pt}
\begin{minipage}[c]{0.28\linewidth}   
\fontsize{7.5}{9}\selectfont
\renewcommand{\arraystretch}{1.0}
\begin{tabularx}{\linewidth}{Xc}
\toprule
\textbf{Agent} & \textbf{Elo} \\
\midrule
CoS        & 1307 \\
LLM-PySC2  & 1501 \\
SwarmBrain & \underline{1562} \\
\rowcolor{gray!12}
\method    & \textbf{1628} \\
\bottomrule
\end{tabularx}
\end{minipage}\hfill
\begin{minipage}[c]{0.69\linewidth}   
\fontsize{7.5}{9}\selectfont
\renewcommand{\arraystretch}{1.0}
\setlength{\tabcolsep}{2.5pt}
\begin{tabularx}{\linewidth}{l*{7}{>{\centering\arraybackslash}X}}
\toprule
\textbf{Variant}
& \textbf{WR}~$\uparrow$ & \textbf{RG}~$\uparrow$ & \textbf{RS}~$\uparrow$ & \textbf{AS}~$\uparrow$ & \textbf{UQ}~$\uparrow$ & \textbf{ED}~$\uparrow$ & \textbf{VC}~$\uparrow$ \\
\midrule
\rowcolor{gray!12}
\method
& \textbf{0.43} & 0.39 & \textbf{0.44} & \textbf{0.44} & \textbf{0.58} & \textbf{0.49} & 0.33 \\
~~w/o Group-Level Control
& 0.07 & 0.33 & 0.38 & 0.38 & 0.57 & 0.30 & 0.29 \\
~~w/o Trajectory Planning
& 0.13 & 0.34 & 0.39 & 0.40 & \textbf{0.58} & 0.37 & 0.27 \\
~~w/o Replanning
& 0.33 & \textbf{0.40} & \textbf{0.44} & 0.43 & 0.57 & 0.45 & \textbf{0.34} \\
~~w/o Memory
& 0.20 & 0.37 & 0.42 & 0.40 & \textbf{0.58} & 0.34 & 0.30 \\
~~w/o Visual Observation
& 0.20 & 0.37 & 0.38 & 0.39 & 0.57 & 0.39 & 0.20 \\
\bottomrule
\end{tabularx}
\end{minipage}%
\vspace{-1.5em}
\end{table}

\begin{figure*}[t]
    \centering
    \includegraphics[width=1\linewidth]{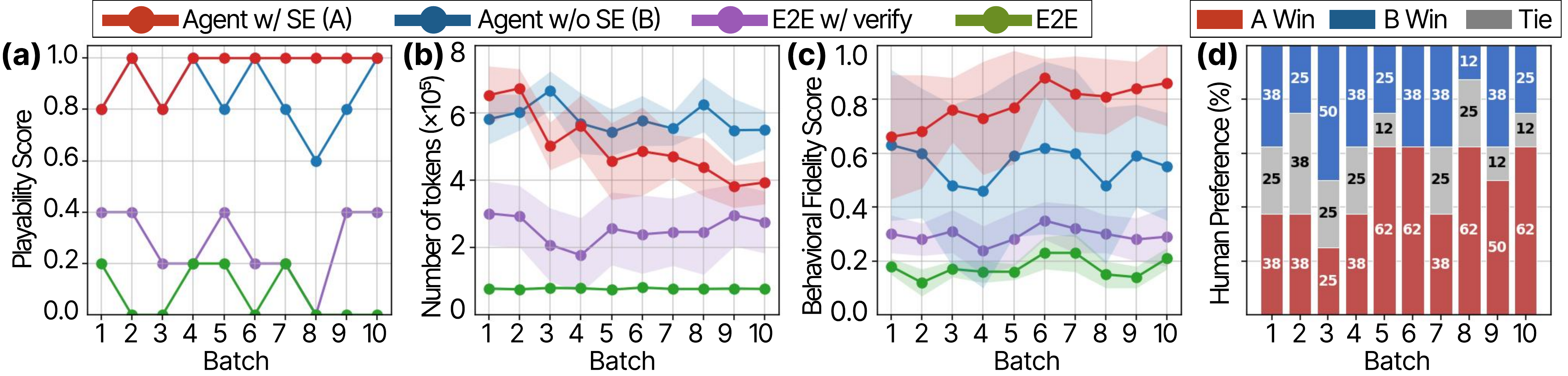}
    \vspace{-1em}
    \caption{\textbf{Self-evolution over successive generation.} (a) Playability, (b) Token Usage, (c) Behavioral Fidelity, and (d) Human Preference between A and B across ten batches of five generations.}
    \label{fig:self_evolution}
    \vspace{-1.0em}
\end{figure*}

\subsection{Detailed Analyses}
\label{sec:detailed_anal}

\vspace{-0.5em}\paragraph{Metric analysis.}
Fig.~\ref{fig:metric_analysis} reports additional metrics for (a) agents under Gemini-3.7-Flash and (b) the four strong backbones in full games (Tab.~\ref{tab:main_results}).
We measure them at Lv.1, where most models win some games.
SwarmBrain nearly matches \method in RG, RS, and VC in Duel, and Grok-4.6 matches or exceeds Gemini-3.7-Flash in RG, RS, AS, and VC in Duel and Symmetric Team, yet both score lower in UQ and ED.
We conjecture that, beyond economy and vision, winning requires fielding stronger units and committing them to battle.
The latter is consistent with the largest mini-game gains of \method and Gemini-3.7-Flash on MFD and AWD, both of which require committing forces against successive attacks.
In Asymmetric Team and Free-for-All, where WR barely separates models, nearly all shares fall below 0.5 and AS drops to about 0.3 for every model, suggesting that agents fail to sustain combat forces against multiple opponents.

\vspace{-1em}\paragraph{Comparing agents.}
To compare agents, we perform head-to-head round-robin duels, whose Elo ratings (Tab.~\ref{tab:elo}) match the built-in-AI ranking.
In particular, \method wins 14 of 20 duels against SwarmBrain~\citep{shao2024swarmbrain}, the strongest baseline, with a 90\% Wilson interval above 0.5.
This agreement supports built-in AIs as a consistent reference for ranking agents.

\vspace{-0.5em}\paragraph{Ablation study.}
We ablate each component of \method and visual observation in full-game duels against Lv.2 built-in AIs, whose intermediate difficulty lets each ablation reveal its effect (Tab.~\ref{tab:ablation}).
Removing \emph{group-level control}, which also disables trajectory planning and replanning built on group commands, causes the largest drops in WR and ED, mainly hurting coordination in engagements.
Removing \emph{trajectory planning} lowers WR, ED, and VC, suggesting that condition-dependent follow-up commands sustain execution and map control between sparse VLM calls.
Without \emph{replanning}, resource, army, and vision metrics stay comparable or even increase, yet ED and WR decrease; we conjecture that it helps respond to critical events in time rather than build strength.
Removing \emph{memory} lowers WR and ED with modest changes elsewhere, suggesting that temporal context turns similar strength into more effective engagements.
Overall, ED varies most among the auxiliary metrics and largely follows WR, whereas UQ stays nearly constant, indicating that the components mainly improve how forces are committed rather than which units are fielded (\cf Fig.~\ref{fig:metric_analysis}).
Removing \emph{vision}, \ie, dropping $v_t$ from $o_t$, reduces VC the most and also lowers ED and WR, suggesting that visual observations help maintain map control, which in turn supports engagements.

\subsection{Evaluating Generated Games}
\label{sec:eval_self-evol}

Fig.~\ref{fig:self_evolution} plots the four criteria as experience accumulates over ten sequential batches of five queries.
We compare the multi-agent pipeline with and without self-evolution (SE), \textit{Agent w/ SE}~(A) and \textit{Agent w/o SE}~(B), with an end-to-end baseline that generates the entire game in one pass, with and without iterative re-verification (\textit{E2E w/ verify} and \textit{E2E}; Appendix~\ref{supp:e2e_impl}).
Appendix~\ref{supp:selfevolving_examples} shows an example generation cycle, rule and rubric updates, and representative generated games, and Appendix~\ref{supp:exp_generated_games} reports a preliminary agent evaluation on generated games.

\noindent\textbf{Multi-agent pipeline \vs end-to-end generation.}
Both Agent variants yield launchable games in nearly every generation and achieve far higher behavioral fidelity than the E2E variants, even without self-evolution.
The E2E variants consume fewer tokens only by skipping stage-wise design.
Repeated verification improves E2E only modestly (\textit{E2E w/ verify}), so it cannot replace this design.

\noindent\textbf{Effect of self-evolution.}
Self-evolution improves quality and efficiency.
Over batches, the behavioral fidelity of \textit{Agent w/ SE} rises and its token usage falls, whereas \textit{Agent w/o SE} shows no such trend.
\textit{Agent w/ SE} also reaches full playability after a few batches, whereas \textit{Agent w/o SE} yields unlaunchable games in some batches.
Human preference between the Agent variants, the only ones with high playability, shifts toward \textit{Agent w/ SE} in later batches.
Removing either database reuse or rubric updates weakens the fidelity and token trends (Appendix~\ref{supp:abl_self_evol_gen_framework}), showing that each contributes.

\section{Conclusion}
\label{sec:conc}
We propose \benchmark, a benchmark built on \textit{Beyond All Reason} that provides holistic evaluation via full games across diverse match-up structures, diagnostic assessment via mini-games each emphasizing a focal strategic demand, and extensible coverage via a self-evolving game generation framework.
For VLMs to operate in large-scale RTS games, we also introduce \method, a baseline agent harness combining group-level trajectory planning with short- and long-term memory.
Using \method, we evaluate SoTA VLMs and find that their performance declines as built-in AIs strengthen and falls short of humans, particularly when they must coordinate with allies or contend with multiple opponents.
Mini-games further suggest that the relative strengths of VLMs vary across strategic demands.
In game generation, experience-driven evolution improves game quality while reducing token usage over successive cycles.

\section*{AI Use Statement}

We utilized AI solely to polish the writing, such as improving the grammar, phrasing, and readability of author-written drafts. 
All AI-suggested edits were reviewed and revised by the authors.

\bibliography{iclr2026_conference}

@article{gandhi2023strategic,
  title={Strategic Reasoning with Language Models},
  author={Gandhi, Kanishk and Sadigh, Dorsa and Goodman, Noah D},
  journal={arXiv preprint arXiv:2305.19165},
  year={2023}
}

@article{verma2025ggbench,
  title={Measuring General Intelligence with Generated Games},
  author={Verma, Vivek and Huang, David and Chen, William and Klein, Dan and Tomlin, Nicholas},
  journal={arXiv preprint arXiv:2505.07215},
  year={2025}
}

@article{yao2025spin,
  title={{SPIN-Bench}: How Well Do {LLMs} Plan Strategically and Reason Socially?},
  author={Yao, Jianzhu and Wang, Kevin and Hsieh, Ryan and Zhou, Haisu and Zou, Tianqing and Cheng, Zerui and Wang, Zhangyang and Viswanath, Pramod},
  journal={arXiv preprint arXiv:2503.12349},
  year={2025}
}

@article{lin2026omnigamearena,
  title={{OmniGameArena}: A Unified {UE5} Benchmark for {VLM} Game Agents with Improvement Dynamics},
  author={Lin, Mingxian and Qian, Shengju and Liu, Yuqi and Huang, Yi-Hua and Wang, Yiyu and Huang, Wei and Li, Yitang and Zhang, Fan and Hu, Zeyu and Zhu, Lingting and Wang, Xin and Qi, Xiaojuan},
  journal={arXiv preprint arXiv:2606.09826},
  year={2026}
}

@article{chen2023gamegpt,
  title={{GameGPT}: Multi-agent Collaborative Framework for Game Development},
  author={Chen, Dake and Zhang, Haoyang and Wang, Hanbin and Huo, Yunhao and Li, Yuzhao and Wang, Junjie},
  journal={arXiv preprint arXiv:2310.08067},
  year={2023}
}

@article{autoenv2025,
  title={AutoEnv: Automated Environments for Measuring Cross-Environment Agent Learning},
  author={Zhang, Jiayi and Peng, Yiran and Kong, Fanqi and Yang, Cheng and Wu, Yifan and Yu, Zhaoyang and Xiang, Jinyu and Ruan, Jianhao and Wang, Jinlin and Song, Maojia and Liu, HongZhang and Tang, Xiangru and Liu, Bang and Wu, Chenglin and Luo, Yuyu},
  journal={arXiv preprint arXiv:2511.19304},
  year={2025}
}

@article{shen2025sc2arena,
  title={SC2Arena and StarEvolve: Benchmark and Self-Improvement Framework for LLMs in Complex Decision-Making Tasks},
  author={Shen, Pengbo and Wang, Yaqing and Mu, Ni and Luan, Yao and Xie, Runpeng and Yang, Senhao and Wang, Lexiang and Hu, Hao and Xu, Shuang and Yang, Yiqin and others},
  journal={arXiv preprint arXiv:2508.10428},
  year={2025}
}

@article{long2024teamcraft,
  title={Teamcraft: A benchmark for multi-modal multi-agent systems in minecraft},
  author={Long, Qian and Li, Zhi and Gong, Ran and Wu, Ying Nian and Terzopoulos, Demetri and Gao, Xiaofeng},
  journal={arXiv preprint arXiv:2412.05255},
  year={2024}
}

@article{hu2024pokellmon,
  title={Pok{\'e}llmon: A human-parity agent for pok{\'e}mon battles with large language models},
  author={Hu, Sihao and Huang, Tiansheng and Liu, Ling},
  journal={arXiv preprint arXiv:2402.01118},
  year={2024}
}

@article{shao2024swarmbrain,
  title={Swarmbrain: Embodied agent for real-time strategy game starcraft ii via large language models},
  author={Shao, Xiao and Jiang, Weifu and Zuo, Fei and Liu, Mengqing},
  journal={arXiv preprint arXiv:2401.17749},
  year={2024}
}

@article{light2023avalonbench,
  title={Avalonbench: Evaluating llms playing the game of avalon},
  author={Light, Jonathan and Cai, Min and Shen, Sheng and Hu, Ziniu},
  journal={arXiv preprint arXiv:2310.05036},
  year={2023}
}

@article{packer2023memgpt,
  title={Memgpt: Towards llms as operating systems},
  author={Packer, Charles and Wooders, Sarah and Lin, Kevin and Fang, Vivian and Patil, Shishir G and Stoica, Ion and Gonzalez, Joseph E},
  journal={arXiv preprint arXiv:2310.08560},
  year={2023}
}

@inproceedings{lin2025gamebot,
  title={Gamebot: Transparent assessment of llm reasoning in games},
  author={Lin, Wenye and Roberts, Jonathan and Yang, Yunhan and Albanie, Samuel and Lu, Zongqing and Han, Kai},
  booktitle={Proceedings of the Association for Computational Linguistics (ACL)},
  pages={7656--7682},
  year={2025}
}

@inproceedings{ma2025ava,
  title={Ava: Attentive vlm agent for mastering starcraft ii},
  author={Ma, Weiyu and Fu, Yuqian and Zhang, Zecheng and Ghanem, Bernard and Li, Guohao},
  booktitle={Findings of the Association for Computational Linguistics (ACL)},
  pages={4270--4290},
  year={2026}
}

@inproceedings{zheng2025v,
  title={V-mage: A game evaluation framework for assessing vision-centric capabilities in multimodal large language models},
  author={Zheng, Xiangxi and Li, Linjie and Yang, Zhengyuan and Yu, Ping and Wang, Alex Jinpeng and Yan, Rui and Yao, Yuan and Wang, Lijuan},
  booktitle={Findings of the Association for Computational Linguistics (ACL)},
  pages={17707--17758},
  year={2026}
}

@article{liu2024lost,
  title={Lost in the Middle: How Language Models Use Long Contexts},
  author={Liu, Nelson F. and Lin, Kevin and Hewitt, John and Paranjape, Ashwin and Bevilacqua, Michele and Petroni, Fabio and Liang, Percy},
  journal={Transactions of the Association for Computational Linguistics (TACL)},
  pages={157--173},
  year={2024}
}

@inproceedings{tang2025dsgbench,
  title={Dsgbench: A diverse strategic game benchmark for evaluating llm-based agents in complex decision-making environments},
  author={Tang, Wenjie and Zhou, Yuan and Cheng, Keyan and Xu, Erqiang and Xiao, Liquan and Li, Minne},
  booktitle={IEEE International Conference on Acoustics, Speech and Signal Processing (ICASSP)},
  pages={16987--16991},
  year={2026}
}

@inproceedings{wang2026towermind,
  title={Towermind: A tower defence game learning environment and benchmark for llm as agents},
  author={Wang, Dawei and Zhou, Chengming and Zhao, Di and Liu, Xinyuan and Ma, Marci Chi and Ushaw, Gary and Davison, Richard},
  booktitle={Proceedings of the AAAI Conference on Artificial Intelligence (AAAI)},
  pages={26151--26159},
  year={2026}
}

@inproceedings{zhao2024expel,
  title={Expel: Llm agents are experiential learners},
  author={Zhao, Andrew and Huang, Daniel and Xu, Quentin and Lin, Matthieu and Liu, Yong-Jin and Huang, Gao},
  booktitle={Proceedings of the AAAI Conference on Artificial Intelligence (AAAI)},
  pages={19632--19642},
  year={2024}
}

@inproceedings{xu2025vs,
  title={VS-BENCH: Evaluating VLMs for Strategic Abilities in Multi-Agent Environments},
  author={Xu, Zelai and Xu, Zhexuan and Yi, Xiangmin and Yuan, Huining and Guang, Mo and Long, Kaiwen and Chen, Xinlei and Wu, Yi and Yu, Chao and Wang, Yu},
  booktitle={Proceedings of the IEEE/CVF Conference on Computer Vision and Pattern Recognition (CVPR)},
  pages={21561--21572},
  year={2026}
}

@inproceedings{khan2025sketchtopia,
  title={Sketchtopia: A dataset and foundational agents for benchmarking asynchronous multimodal communication with iconic feedback},
  author={Khan, Mohd Hozaifa and Sarvadevabhatla, Ravi Kiran},
  booktitle={Proceedings of the IEEE/CVF Conference on Computer Vision and Pattern Recognition (CVPR)},
  pages={18176--18186},
  year={2025}
}

@inproceedings{wang2025escapecraft,
  title     = {How Do Multimodal Large Language Models Handle Complex Multimodal Reasoning? Placing Them in An Extensible Escape Game},
  author    = {Wang, Ziyue and Dong, Yurui and Luo, Fuwen and Ruan, Minyuan and Cheng, Zhili and Chen, Chi and Li, Peng and Liu, Yang},
  booktitle = {Proceedings of the IEEE/CVF International Conference on Computer Vision (ICCV)},
  pages     = {4807--4817},
  year      = {2025}
}

@inproceedings{park2025orak,
  title={Orak: A foundational benchmark for training and evaluating llm agents on diverse video games},
  author={Park, Dongmin and Kim, Minkyu and Choi, Beongjun and Kim, Junhyuck and Lee, Keon and Lee, Jonghyun and Park, Inkyu and Lee, ByeongUk and Hwang, Jaeyoung and Ahn, Jaewoo and others},
  booktitle={International Conference on Learning Representations (ICLR)},
  pages={60684--60744},
  year={2026}
}

@inproceedings{hong2024metagpt,
  title={MetaGPT: Meta programming for a multi-agent collaborative framework},
  author={Hong, Sirui and Zhuge, Mingchen and Chen, Jonathan and Zheng, Xiawu and Cheng, Yuheng and Wang, Jinlin and Zhang, Ceyao and Yau, Steven and Lin, Zijuan and Zhou, Liyang and others},
  booktitle={International Conference on Learning Representations (ICLR)},
  pages={23247--23275},
  year={2024}
}

@inproceedings{hu2025lmgame,
  title={lmgame-bench: How good are llms at playing games?},
  author={Hu, Lanxiang and Huo, Mingjia and Zhang, Yuxuan and Yu, Haoyang and Xing, Eric P and Stoica, Ion and Rosing, Tajana and Jin, Haojian and Zhang, Hao},
  booktitle={International Conference on Learning Representations (ICLR)},
  pages={81308--81356},
  year={2026}
}

@inproceedings{qi2024civrealm,
  title={Civrealm: A learning and reasoning odyssey in civilization for decision-making agents},
  author={Qi, Siyuan and Chen, Shuo and Li, Yexin and Kong, Xiangyu and Wang, Junqi and Yang, Bangcheng and Zhong, Yifan and Zhang, Xiaoyuan and Zhang, Zhaowei and Liu, Nian and others},
  booktitle={International Conference on Learning Representations (ICLR)},
  pages={27995--28026},
  year={2024}
}

@inproceedings{faldor2025omniepic,
  title={Omni-epic: Open-endedness via models of human notions of interestingness with environments programmed in code},
  author={Faldor, Maxence and Zhang, Jenny and Cully, Antoine and Clune, Jeff},
  booktitle={International Conference on Learning Representations (ICLR)},
  pages={85260--85385},
  year={2025}
}

@inproceedings{paglieri2025balrog,
  title={Balrog: Benchmarking agentic llm and vlm reasoning on games},
  author={Paglieri, Davide and Cupia{\l}, Bart{\l}omiej and Coward, Samuel and Piterbarg, Ulyana and Wo{\l}czyk, Maciej and Khan, Akbir and Pignatelli, Eduardo and Kuci{\'n}ski, {\L}ukasz and Pinto, Lerrel and Fergus, Rob and others},
  booktitle = {International Conference on Learning Representations (ICLR)},
  pages={96666--96702},
  year={2025}
}

@inproceedings{li2026winning,
  title={From Winning to Understanding: A Diagnostic Long-Horizon {RTS} Benchmark for {LLMs}},
  author={Li, Jiacheng and Liu, Jiahui and Wang, Yuqing and Cui, Gaochen and Zhang, Xiao and Zhao, Qianchuan and Zhang, Ziyou and Li, Chenghao},
  booktitle={International Conference on Machine Learning (ICML)},
  year={2026}
}

@inproceedings{shi2023large,
  title={Large language models can be easily distracted by irrelevant context},
  author={Shi, Freda and Chen, Xinyun and Misra, Kanishka and Scales, Nathan and Dohan, David and Chi, Ed H and Sch{\"a}rli, Nathanael and Zhou, Denny},
  booktitle={International conference on machine learning (ICML)},
  pages={31210--31227},
  year={2023}
}

@inproceedings{zheng2025mcu,
  title={MCU: An Evaluation Framework for Open-Ended Game Agents},
  author={Zheng, Xinyue and Lin, Haowei and He, Kaichen and Wang, Zihao and Fu, Qiang and Fu, Haobo and Zheng, Zilong and Liang, Yitao},
  booktitle={International Conference on Machine Learning (ICML)},
  year={2025}
}

@inproceedings{liu2023llava,
  title     = {Visual Instruction Tuning},
  author    = {Liu, Haotian and Li, Chunyuan and Wu, Qingyang and Lee, Yong Jae},
  booktitle = {Advances in Neural Information Processing Systems (NeurIPS)},
  year      = {2023}
}

@article{li2025llmpysc2,
  title={Llm-pysc2: Starcraft ii learning environment for large language models},
  author={Li, Zongyuan and Ni, Yanan and Qi, Runnan and Lu, Chang and Jiang, Lumin and Xiaojie, Xu and Liu, Xiangbei and Li, Pengfei and Guo, Yunzheng and Ma, Zhe and others},
  journal={Advances in Neural Information Processing Systems (NeurIPS)},
  pages={127595--127657},
  year={2025}
}

@article{alayrac2022flamingo,
  title={Flamingo: a visual language model for few-shot learning},
  author={Alayrac, Jean-Baptiste and Donahue, Jeff and Luc, Pauline and Miech, Antoine and Barr, Iain and Hasson, Yana and Lenc, Karel and Mensch, Arthur and Millican, Katherine and Reynolds, Malcolm and others},
  journal={Advances in neural information processing systems (NeurIPS)},
  pages={23716--23736},
  year={2022}
}

@article{shinn2023reflexion,
  title={Reflexion: Language agents with verbal reinforcement learning},
  author={Shinn, Noah and Cassano, Federico and Gopinath, Ashwin and Narasimhan, Karthik and Yao, Shunyu},
  journal={Advances in neural information processing systems (NeurIPS)},
  pages={8634--8652},
  year={2023}
}

@article{ma2024large,
  title={Large language models play starcraft ii: Benchmarks and a chain of summarization approach},
  author={Ma, Weiyu and Mi, Qirui and Zeng, Yongcheng and Yan, Xue and Wu, Yuqiao and Lin, Runji and Zhang, Haifeng and Wang, Jun},
  journal={Advances in neural information processing systems (NeurIPS)},
  pages={133386--133442},
  year={2024}
}

@inproceedings{zhang2024llm,
  title={LLM as a Mastermind: A Survey of Strategic Reasoning with Large Language Models},
  author={Zhang, Yadong and Mao, Shaoguang and Ge, Tao and Wang, Xun and de Wynter, Adrian and Xia, Yan and Wu, Wenshan and Song, Ting and Lan, Man and Wei, Furu},
  booktitle={Conference on Language Modeling (COLM)},
  year={2024}
}

@inproceedings{ahn2025hima,
  title={Society of Mind Meets Real-Time Strategy: A Hierarchical Multi-Agent Framework for Strategic Reasoning},
  author={Ahn, Daechul and Kim, San and Choi, Jonghyun},
  booktitle={Conference on Language Modeling (COLM)},
  year={2025}
}

@inproceedings{buro2003real,
  title={Real-Time Strategy Games: A New AI Research Challenge},
  author={Buro, Michael},
  booktitle={International Joint Conference on Artificial Intelligence (IJCAI)},
  pages={1534--1535},
  year={2003}
}

@article{vinyals2019grandmaster,
  title={Grandmaster Level in StarCraft II Using Multi-Agent Reinforcement Learning},
  author={Vinyals, Oriol and Babuschkin, Igor and Czarnecki, Wojciech M and Mathieu, Micha{\"e}l and Dudzik, Andrew and Chung, Junyoung and Choi, David H and Powell, Richard and Ewalds, Timo and Georgiev, Petko and others},
  journal={Nature},
  volume={575},
  number={7782},
  pages={350--354},
  year={2019},
  publisher={Nature Publishing Group}
}

@misc{bar2024beyondallreason,
  title={Beyond All Reason},
  author={{Beyond All Reason Team}},
  year={2024},
  howpublished={\url{https://www.beyondallreason.info/}},
  note={Open-source real-time strategy game}
}

@misc{bar2024qualitipedia,
  title={Beyond All Reason},
  author={{Qualitipedia contributors}},
  year={2024},
  howpublished={\url{https://newqualitipedia.telepedia.net/wiki/Beyond_All_Reason}}
}

@article{anne2025harnessing,
  title={Harnessing language for coordination: A framework and benchmark for llm-driven multiagent control},
  author={Anne, Timoth{\'e}e and Syrkis, Noah and Elhosni, Meriem and Turati, Florian and Legendre, Franck and Jaquier, Alain and Risi, Sebastian},
  journal={IEEE Transactions on Games},
  volume={17},
  number={4},
  pages={933--943},
  year={2025},
  publisher={IEEE}
}

@article{ontanon2013survey,
  title={A Survey of Real-Time Strategy Game AI Research and Competition in StarCraft},
  author={Onta{\~n}{\'o}n, Santiago and Synnaeve, Gabriel and Uriarte, Alberto and Richoux, Florian and Churchill, David and Preuss, Mike},
  journal={IEEE Transactions on Computational Intelligence and AI in Games},
  volume={5},
  number={4},
  pages={293--311},
  year={2013},
  publisher={IEEE}
}

@article{robertson2014review,
  title={A Review of Real-Time Strategy Game AI},
  author={Robertson, Glen and Watson, Ian},
  journal={AI Magazine},
  volume={35},
  number={4},
  pages={75--104},
  year={2014},
  publisher={AAAI}
}

@inproceedings{park2023generative,
  title={Generative Agents: Interactive Simulacra of Human Behavior},
  author={Park, Joon Sung and O'Brien, Joseph C and Cai, Carrie J and Morris, Meredith Ringel and Liang, Percy and Bernstein, Michael S},
  booktitle={ACM Symposium on User Interface Software and Technology (UIST)},
  pages={1--22},
  year={2023}
}

@incollection{atkinson1968human,
  title={Human Memory: A Proposed System and Its Control Processes},
  author={Atkinson, Richard C and Shiffrin, Richard M},
  booktitle={Psychology of Learning and Motivation},
  volume={2},
  pages={89--195},
  year={1968},
  publisher={Academic Press}
}

@article{squire1992memory,
  title={Memory and the Hippocampus: A Synthesis from Findings with Rats, Monkeys, and Humans},
  author={Squire, Larry R},
  journal={Psychological Review},
  volume={99},
  number={2},
  pages={195--231},
  year={1992}
}

@book{fullerton2014game,
  title     = {Game Design Workshop: A Playcentric Approach to Creating Innovative Games},
  author    = {Fullerton, Tracy},
  year      = {2014},
  edition   = {3rd},
  publisher = {CRC Press},
  address   = {Boca Raton, FL}
}

@book{ierusalimschy2006lua,
  title={Programming in Lua},
  author={Ierusalimschy, Roberto},
  year={2006},
  publisher={Lua.org}
}

@misc{recoil,
  author       = {{Beyond All Reason Developers}},
  title        = {Recoil Engine: A Powerful Free Cross-Platform {RTS} Game Engine},
  year         = {2023},
  howpublished = {\url{https://github.com/beyond-all-reason/RecoilEngine}},
  note         = {A hard fork of the SpringRTS engine (version 105 tree)}
}

@misc{bar_repo,
  author       = {{Beyond All Reason Developers}},
  title        = {Beyond All Reason: Main Game Repository},
  howpublished = {\url{https://github.com/beyond-all-reason/Beyond-All-Reason}},
  year         = {2019},
}

@misc{sc2_wiki,
  author       = {{Fandom Contributors}},
  title        = {{StarCraft~II} --- {StarCraft} Wiki},
  howpublished = {\url{https://starcraft.fandom.com/wiki/StarCraft_II}},
  year         = {2024},
  note         = {Accessed: 2025},
}

@article{hendrikx2013procedural,
  title     = {Procedural Content Generation for Games: A Survey},
  author    = {Hendrikx, Mark and Meijer, Sebastiaan and van der Velden, Joeri and Iosup, Alexandru},
  journal   = {ACM Transactions on Multimedia Computing, Communications, and Applications},
  volume    = {9},
  number    = {1},
  pages     = {1:1--1:22},
  year      = {2013},
  publisher = {ACM},
  doi       = {10.1145/2422956.2422957}
}

@article{jonsson2007use,
  title={The Use of Scoring Rubrics: Reliability, Validity and Educational Consequences},
  author={Jonsson, Anders and Svingby, Gunilla},
  journal={Educational Research Review},
  volume={2},
  number={2},
  pages={130--144},
  year={2007},
  doi={10.1016/j.edurev.2007.05.002}
}

@article{kim2026multimodal,
  title={A Multimodal {RTS} Game Benchmark for Evaluating Vision-Language Models},
  author={Kim, San and Ahn, Daechul and Choi, Jonghyun},
  journal={Journal of KIISE},
  volume={53},
  number={4},
  pages={331--338},
  year={2026},
  doi={10.5626/JOK.2026.53.4.331}
}

@misc{krippendorff2011computing,
  author       = {Krippendorff, Klaus},
  title        = {Computing {K}rippendorff's Alpha-Reliability},
  year         = {2011},
  howpublished = {Departmental Papers (ASC) 43, Annenberg School for Communication, University of Pennsylvania},
  url          = {https://repository.upenn.edu/asc_papers/43}
}
\bibliographystyle{iclr2026_conference}

\appendix

\startcontents[appendix]

\clearpage
\section*{Appendix Contents}
\printcontents[appendix]{l}{1}{\setcounter{tocdepth}{2}}
\vspace{1em}

\clearpage


\section{Beyond All Reason: Gameplay and Scalability}
\label{supp:bar_overview}

\subsection{Game Characteristics for Strategic Reasoning}
\label{supp:bar_game_characteristic}

\textit{Beyond All Reason} (BAR)~\citep{bar2024beyondallreason} is an open-source real-time strategy (RTS) game built on the Recoil engine~\citep{recoil}, a fork of SpringRTS.
BAR supports substantially larger battles than StarCraft~II (SC2) in unit count, player count, and battlefield size (Sec.~\ref{supp:sc2_bar}).
Players choose from three factions, \textbf{Armada}, \textbf{Cortex}, and \textbf{Legion}, each with a distinct playstyle.
Armada emphasizes flexible and mobile warfare with cloaking and precision weapons, Cortex favors durable and defensive play with heavily armored units, and Legion adopts an aggressive style centered on resource-intensive offensive units.
In our full games, all players use Armada (Sec.~\ref{supp:full_game}).

\noindent \textbf{Resource management.}
BAR uses two primary resources, \textit{metal} and \textit{energy}, whose production and consumption must be kept in balance.
Metal is extracted from map deposits or reclaimed from wreckage, whereas energy is generated by structures such as solar collectors, wind turbines, geothermal power plants, and fusion reactors.
BAR employs a \textit{streaming economy}, in which resources are generated and consumed as continuous flows rather than managed solely as finite stockpiles.
Production throughput therefore depends on resource income, requiring players to continuously trade off economic expansion, technological progression, and immediate military investment.

\noindent \textbf{Base construction and technology progression.}
BAR enforces explicit construction and technology dependencies, making advanced units and structures available only after their prerequisites are completed.
Effective production therefore requires sequential reasoning over build orders, technology paths, and the allocation of construction capacity across multiple facilities.

\noindent \textbf{Low-level automation.}
BAR delegates much of the per-unit and economic execution to the game engine, reducing routine micro-management~\citep{bar2024beyondallreason}\footnote{\url{https://www.beyondallreason.info/commands-20}}.
In combat, armed units automatically select and fire at enemies within range according to their firing stance (fire at will, return fire, or hold fire), while their movement state (hold position, maneuver, or roam) determines how far they pursue targets.
Given a destination, units plan their own routes through engine-level pathfinding.
An attack-move makes them stop to engage enemies within firing range before continuing, whereas a plain move prioritizes reaching the destination.
On the economic side, construction units execute queued build orders in sequence, assist other constructors and factories, and reclaim or repair everything within a designated area.
Factories can repeat their production queues, and energy converters run only when stored energy exceeds a set threshold.
Consequently, a single command, such as sending a group to a position, expands into many low-level actions without further input.
These mechanisms automate how commands are executed, not which commands to issue.
\method's group commands build on them, with \texttt{move} engaging enemies on contact before resuming and \texttt{move\_force} continuing toward the target (Sec.~\ref{sec:rtsagent}).

\noindent \textbf{Army coordination and strategic maneuvering.}
Effective play therefore centers on coordinating forces at the group level.
Players must decide how to organize units into groups and allocate them across fronts, when and where to engage, and how to maneuver them as battlefield conditions evolve under fog-of-war.
These coordination and planning demands are central to the strategic capabilities that \benchmark is designed to evaluate.

\subsection{Scale Comparison with StarCraft II}
\label{supp:sc2_bar}

\noindent \textbf{Unit variety.}
SC2 and BAR each feature three playable factions.
SC2 contains 96 unique units and buildings across Terran, Protoss, and Zerg, whereas BAR contains 554~\citep{bar_repo}.
BAR further supports a multi-tier progression system with faction-specific units across Tier~1, Tier~2, and Tier~3, as well as multiple movement domains spanning land, air, sea, and amphibious operations.
This broader unit space increases the diversity of possible army compositions, counters, and strategic interactions.

\noindent \textbf{Supply capacity.}
SC2 uses a \textit{weighted} population system in which units consume different amounts of supply~\citep{sc2_wiki}.
For example, a Battlecruiser consumes six supply, whereas a Marine consumes one.
Thus, the standard per-player cap of 200 supply does not correspond to 200 fielded units.
A late-game army with approximately 60 workers, for instance, typically leaves room for roughly 110--130 combat units, depending on army composition.
For this reason, the SC2 supply cap in Tab.~\ref{tab:sc2_vs_bar_extreme_scale} is marked with an asterisk, indicating that nominal supply should not be read directly as unit count.
BAR instead counts each unit equally toward its unit limit regardless of tier or combat strength, with a default per-player cap of 2,000 units~\citep{bar_repo}.
This limit is configurable, and we use the default in our experiments.

\begin{figure}[t]
\centering
\includegraphics[width=\linewidth]{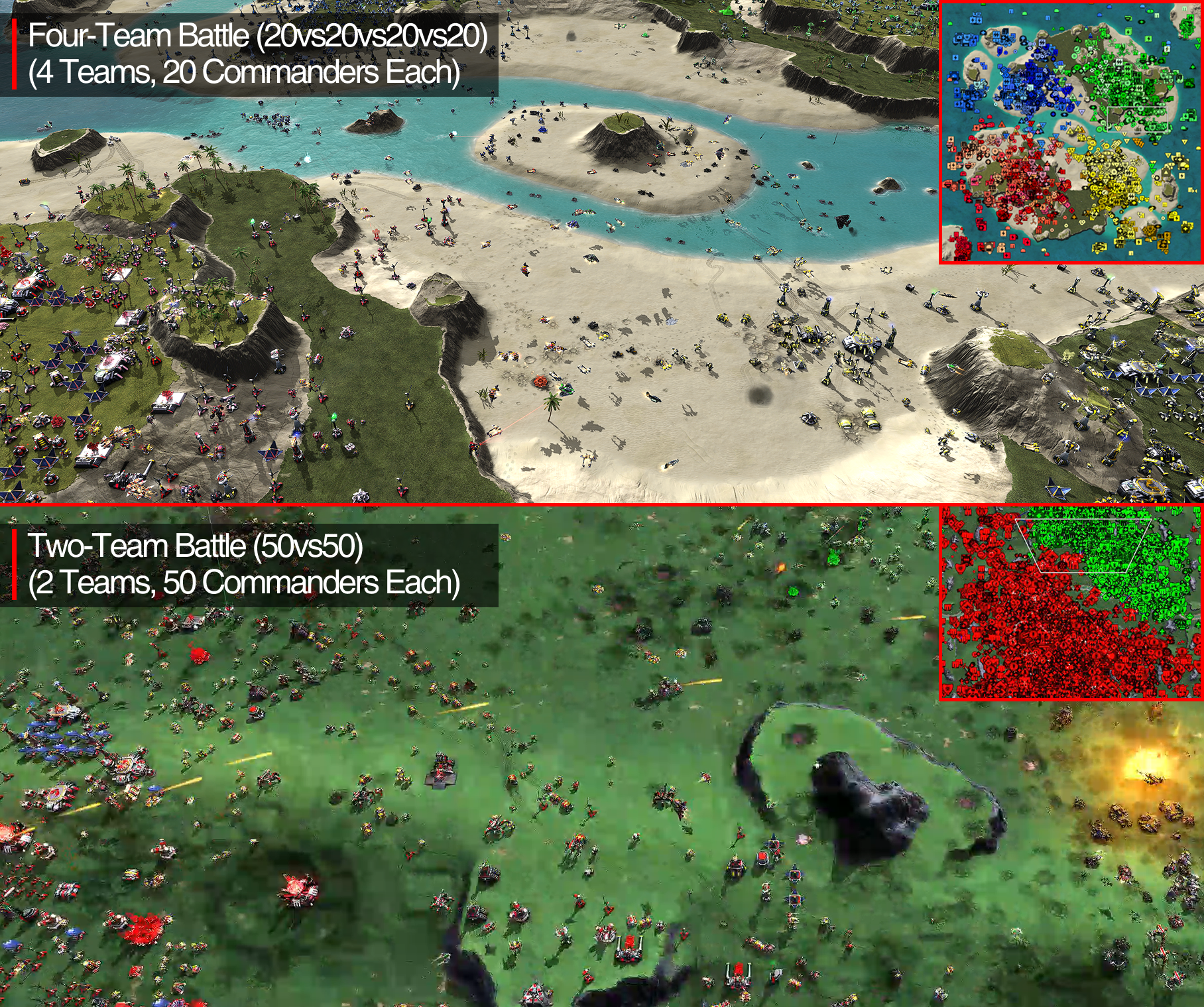}
\vspace{-0.5em}
\caption{\textbf{Large-scale combat environments in BAR.}
\textbf{Top.} A four-team battle with 20 commanders per team (80 in total).
\textbf{Bottom.} A 50 \vs 50 team battle with 50 commanders per side.
Such battles involve large numbers of players and thousands of units within a single battlefield.}
\label{fig:bar_scale}
\vspace{-1.0em}
\end{figure}

\noindent \textbf{Unit capacity.}
Multiplying the eight-player limit by the 200-supply cap gives SC2 a nominal upper bound of 1,600 supply.
Because supply is weighted by unit type and a large fraction typically goes to workers, this value overstates the number of simultaneously fielded combat units; the illustrative late-game range above yields approximately 880--1,040 combat units across eight players.
In contrast, BAR supports up to 32,000 simultaneous units across all players~\citep{bar_repo}.

\noindent \textbf{Map size.}
We compare the largest available maps in each game.
The SC2 map editor limits playable maps to $256 \times 256$ tiles~\citep{sc2_wiki}.
BAR specifies map dimensions in map units of 512 elmos, the engine's distance unit, each spanning 64 squares of 8 elmos.
The largest official BAR map measures $32 \times 32$ map units, \ie, $2{,}048 \times 2{,}048$ squares.
Treating an SC2 tile and a BAR square as comparable grid cells, the largest BAR map covers approximately $64\times$ the area of the largest SC2 map~\citep{bar2024beyondallreason}.

\noindent \textbf{Player limit.}
Standard SC2 multiplayer supports up to eight players~\citep{sc2_wiki}.
BAR officially supports up to 100 players in a single match~\citep{bar2024beyondallreason}, although competitive matches commonly use smaller configurations such as 8v8.

\noindent \textbf{Large-scale engagements in practice.}
Tab.~\ref{tab:sc2_vs_bar_extreme_scale} reports the maximum scale supported by each game, whereas Fig.~\ref{fig:bar_scale} shows how this capacity translates into large multiplayer battles, where coordination complexity grows with the numbers of players, units, and interacting fronts.
This scale motivates our use of BAR as a testbed for strategic reasoning under large-scale multi-agent interaction.


\section{Further Comparison with Existing Game Benchmarks for LLM and VLM Agents}
\label{supp:benchmark_comparison}

We compare against game-based benchmarks for LLM and VLM agents that target strategic or multi-agent decision-making or support scenario generation (Tab.~\ref{tab:benchmark_comparison}), and against the RTS benchmarks for LLM and VLM agents that we are aware of (Tab.~\ref{tab:rts_comparison}).
The first two properties in Tab.~\ref{tab:benchmark_comparison} follow from our definition of strategic reasoning, \ie, deciding under uncertainty (imperfect information) while interacting with other agents (multi-agent), and the remaining three from our focus on VLMs (multimodal input) and the diagnostic and extensible desiderata in Sec.~\ref{sec:introduction} (diagnostic evaluation and scenario generation).

\begin{table}[t]
\centering
\caption{\textbf{Comparison with game-based benchmarks for LLM/VLM agents.}
Single-game benchmarks are listed separately from multi-game suites, for which a property is marked if at least one constituent game or evaluation track supports it; in contrast, \benchmark provides all properties within a single environment.
\textit{Diagnostic Evaluation} indicates evaluation that decomposes performance by specific demands or abilities beyond aggregate outcomes, via targeted scenarios or dimension-wise scores.
\textit{Scenario Generation} indicates automated generation of new evaluation scenarios, \eg, tasks or games specified by an LLM (\cmark), procedural or manual variation of scenarios within predefined task families (\pmark), or a fixed scenario set (\xmark).}
\vspace{0.5em}
\resizebox{\textwidth}{!}{%
\begin{tabular}{llccccc}
\toprule
\textbf{Benchmark}
& \makecell[c]{\textbf{Game(s)}}
& \makecell[c]{\textbf{Imperfect}\\ \textbf{Information}}
& \makecell[c]{\textbf{Multi-}\\ \textbf{Agent}}
& \makecell[c]{\textbf{Multimodal}\\ \textbf{Input}}
& \makecell[c]{\textbf{Diagnostic}\\ \textbf{Evaluation}}
& \makecell[c]{\textbf{Scenario}\\ \textbf{Generation}} \\
\midrule
\multicolumn{7}{l}{\textit{Single-game benchmarks}} \\
EscapeCraft~\citep{wang2025escapecraft} & Room Escape Game & \xmark & \xmark & \cmark & \xmark & \pmark \\
AvalonBench~\citep{light2023avalonbench} & Avalon & \cmark & \cmark & \xmark & \xmark & \xmark \\
Sketchtopia~\citep{khan2025sketchtopia} & Sketchtopia & \cmark & \cmark & \cmark & \xmark & \xmark \\
CivRealm~\citep{qi2024civrealm} & FreeCiv & \cmark & \cmark & \xmark & \cmark & \xmark \\
TeamCraft~\citep{long2024teamcraft} & Minecraft & \cmark & \cmark & \cmark & \xmark & \pmark \\
MCU~\citep{zheng2025mcu} & Minecraft & \cmark & \xmark & \cmark & \cmark & \cmark \\
\midrule
\multicolumn{7}{l}{\textit{Multi-game suites}} \\
SPIN-Bench~\citep{yao2025spin} & Multiple & \cmark & \cmark & \xmark & \cmark & \xmark \\
DSGBench~\citep{tang2025dsgbench} & Multiple & \cmark & \cmark & \xmark & \cmark & \pmark \\
gg-bench~\citep{verma2025ggbench} & Generated & \cmark & \cmark & \xmark & \xmark & \cmark \\
VS-Bench~\citep{xu2025vs} & Multiple & \cmark & \cmark & \cmark & \cmark & \xmark \\
OmniGameArena~\citep{lin2026omnigamearena} & Multiple & \cmark & \cmark & \cmark & \cmark & \xmark \\
\midrule
\makecell[l]{\textbf{\benchmark} (Ours)} & Beyond All Reason & \cmark & \cmark & \cmark & \cmark & \cmark \\
\bottomrule
\end{tabular}%
}
\label{tab:benchmark_comparison}
\end{table}

\noindent \textbf{Game-based benchmarks.}
As shown in Tab.~\ref{tab:benchmark_comparison}, single-game benchmarks cover only subsets of the five properties, and multi-game suites broaden coverage by combining games with complementary properties.
MCU~\citep{zheng2025mcu} generates tasks with an LLM and scores trajectories along multiple dimensions, but targets single-agent play in Minecraft.
VS-Bench~\citep{xu2025vs} is the closest in scope, evaluating VLMs in cooperative, competitive, and mixed-motive environments and separating strategic reasoning, measured by next-action prediction, from decision-making; its environments, however, are a fixed set of small-scale board, card, and video games.
DSGBench~\citep{tang2025dsgbench} scores LLM agents along five capability dimensions across six text-based strategic games, including \textit{StarCraft~II}, and supports manual customization of scenario variables such as opponent strategies.
SPIN-Bench~\citep{yao2025spin} varies action spaces, state complexity, and the number of interacting agents across planning, board, card, and negotiation tasks, relying on textual observations.
OmniGameArena~\citep{lin2026omnigamearena} evaluates VLM agents in newly built solo, competitive, and cooperative games, each designed around a subset of embodied capabilities such as spatial navigation and reaction, and focuses on how performance improves through reflection.
gg-bench~\citep{verma2025ggbench} is the closest to our generation framework, using an LLM to generate new two-player turn-based games as a data-generating benchmark; its games are text-based, and its pipeline samples novel games rather than targeting strategic demands specified by users.
Among the compared benchmarks, only \benchmark fully supports all five properties, and it does so within a single large-scale RTS environment.

\noindent \textbf{RTS-specific benchmarks.}
Prior RTS benchmarks evaluate agents in simplified hand-crafted environments, in \textit{StarCraft~II}, or, more recently, in \textit{Red Alert} (Tab.~\ref{tab:rts_comparison}).
Several track behavioral metrics beyond win rate~\citep{ma2024large,shen2025sc2arena,tang2025dsgbench,li2025llmpysc2,li2026winning}.
The \textit{Red Alert} benchmark of \citet{li2026winning} logs economy, combat-loss, and visibility statistics and ranks models by head-to-head Elo ratings, in the same spirit as our auxiliary metrics and head-to-head evaluation (Sec.~\ref{sec:detailed_anal}).
It differs from \benchmark in that agents receive structured textual states, play one-on-one matches against rule-based opponents of distinct playstyles or other LLMs, and are diagnosed through opponent-style and rule variations and language interventions rather than scenarios organized around strategic demands.
Unlike these one-on-one metrics, our auxiliary metrics are designed for multi-player match-ups, normalizing per-player quantities by alive span and reporting odds-normalized shares (Sec.~\ref{supp:metrics}).
Some benchmarks also probe specific abilities through controlled scenarios~\citep{anne2025harnessing,ma2025ava,li2025llmpysc2}.
HIVE and AVACraft do not pair these scenarios with full-game evaluation, and LLM-PySC2 pairs its mini-scenarios only with one-on-one games.
None of them organizes its scenarios around a taxonomy of RTS strategic demands whose alignment is verified by human ratings (Sec.~\ref{supp:demand_ratings}).
Overall, no prior RTS benchmark in Tab.~\ref{tab:rts_comparison} combines the following properties, and none provides (ii) or (iv):
(i) multimodal observations that support direct evaluation of VLM agents;
(ii) full games across diverse match-up structures, including team and free-for-all settings;
(iii) diagnostic mini-games grounded in a taxonomy of RTS AI challenges~\citep{buro2003real}, each emphasizing a focal strategic demand;
(iv) automated generation of new diagnostic scenarios from free-form user queries; and
(v) an action space spanning building construction, unit production, and unit action.
Moreover, BAR provides substantially larger battles, maps, and player configurations than \textit{StarCraft~II} (Tab.~\ref{tab:sc2_vs_bar_extreme_scale}; Sec.~\ref{supp:sc2_bar}), enabling evaluation under larger-scale coordination and longer-horizon strategic demands.

\begin{table}[t]
\centering
\caption{\textbf{Comparison with RTS-specific benchmarks.}
\textit{Full-Game Match-ups} lists the player configurations of full-game evaluation (--: scenario-based evaluation only; Solo: a single player against scripted enemy waves); match-ups refer to player-level configurations rather than the number of LLM agents controlling a single player.
\textit{Diagnostic Scenarios} indicates controlled scenarios designed around specific strategic demands or abilities (\pmark: controlled variations not organized around strategic demands, \eg, opponent-strategy or rule perturbations); unlike \textit{Diagnostic Evaluation} in Tab.~\ref{tab:benchmark_comparison}, it does not count dimension-wise scores, which fall under \textit{Behavioral Metrics}.
\textit{Behavioral Metrics} indicates in-game statistics that characterize gameplay beyond task outcome or its graded progress, \eg, resource, combat, or visibility statistics.
\textit{Scenario Generation} follows Tab.~\ref{tab:benchmark_comparison}.
Build, Prod., and Action denote building construction, unit production, and unit actions, respectively.
$^{\dagger}$The \textit{StarCraft~II} component of a multi-game suite.}
\vspace{0.5em}
\resizebox{\textwidth}{!}{%
\begin{tabular}{llcccccc}
\toprule
\textbf{Benchmark}
& \makecell[c]{\textbf{Game}}
& \makecell[c]{\textbf{Multimodal}\\ \textbf{Input}}
& \makecell[c]{\textbf{Full-Game}\\ \textbf{Match-ups}}
& \makecell[c]{\textbf{Diagnostic}\\ \textbf{Scenarios}}
& \makecell[c]{\textbf{Behavioral}\\ \textbf{Metrics}}
& \makecell[c]{\textbf{Scenario}\\ \textbf{Generation}}
& \makecell[c]{\textbf{Action Space}\\ \textbf{Categories}} \\
\midrule
HIVE~\citep{anne2025harnessing} & Hand-Crafted Game & \xmark & -- & \cmark & \xmark & \xmark & Action \\
TowerMind~\citep{wang2026towermind} & Hand-Crafted Game & \cmark & Solo & \xmark & \xmark & \pmark & Build + Prod. + Action \\
TextStarCraftII~\citep{ma2024large} & StarCraft II & \xmark & 1v1 & \xmark & \cmark & \xmark & Build + Prod. \\
TextSCII-All~\citep{ahn2025hima} & StarCraft II & \xmark & 1v1 & \xmark & \xmark & \xmark & Build + Prod. \\
SC2Arena~\citep{shen2025sc2arena} & StarCraft II & \xmark & 1v1 & \xmark & \cmark & \xmark & Build + Prod. + Action \\
DSGBench~\citep{tang2025dsgbench} & StarCraft II$^{\dagger}$ & \xmark & 1v1 & \pmark & \cmark & \pmark & Build + Prod. \\
AVACraft~\citep{ma2025ava} & StarCraft II & \cmark & -- & \cmark & \xmark & \xmark & Action \\
LLM-PySC2~\citep{li2025llmpysc2} & StarCraft II & \cmark & 1v1 & \cmark & \cmark & \xmark & Build + Prod. + Action \\
Kim et al.~\citep{kim2026multimodal} & StarCraft II & \cmark & 1v1 & \xmark & \xmark & \xmark & Build + Prod. + Action \\
Li et al.~\citep{li2026winning} & Red Alert & \xmark & 1v1 & \pmark & \cmark & \xmark & Build + Prod. + Action \\
\midrule
\makecell[l]{\textbf{\benchmark} (Ours)} & Beyond All Reason (BAR) & \cmark & 1v1, Team, FFA & \cmark & \cmark & \cmark & Build + Prod. + Action \\
\bottomrule
\end{tabular}%
}
\label{tab:rts_comparison}
\end{table}

\noindent \textbf{Automated game and environment generation.}
LLMs have also been used to generate games and environments, mostly for purposes other than diagnostic evaluation.
GameGPT~\citep{chen2023gamegpt} automates game development with a role-specialized multi-agent framework; OMNI-EPIC~\citep{faldor2025omniepic} and AutoEnv~\citep{autoenv2025} generate executable environments in code to drive open-ended learning and to study agent learning across heterogeneous environments, respectively; and gg-bench~\citep{verma2025ggbench} generates games to benchmark general reasoning.
Our framework similarly adopts role-specialized agents~\citep{hong2024metagpt,chen2023gamegpt}, but targets diagnostic evaluation: it generates scenarios on demand from free-form queries within a complex RTS engine, designs each around a focal strategic demand, and validates it through staged rubric checks and simulation tests.
Moreover, whereas task archives condition the proposal of novel tasks~\citep{faldor2025omniepic}, our framework reuses validated design documents and rule code to bypass redundant stages and accumulates validation experience in hierarchical rubrics (Sec.~\ref{sec:minigame_gen}).


\section{Game Interface and Static Game Knowledge \texorpdfstring{$\mathcal{K}$}{K}}
\label{supp:interface}

\begin{figure}[t]
\centering
\includegraphics[width=0.75\linewidth]{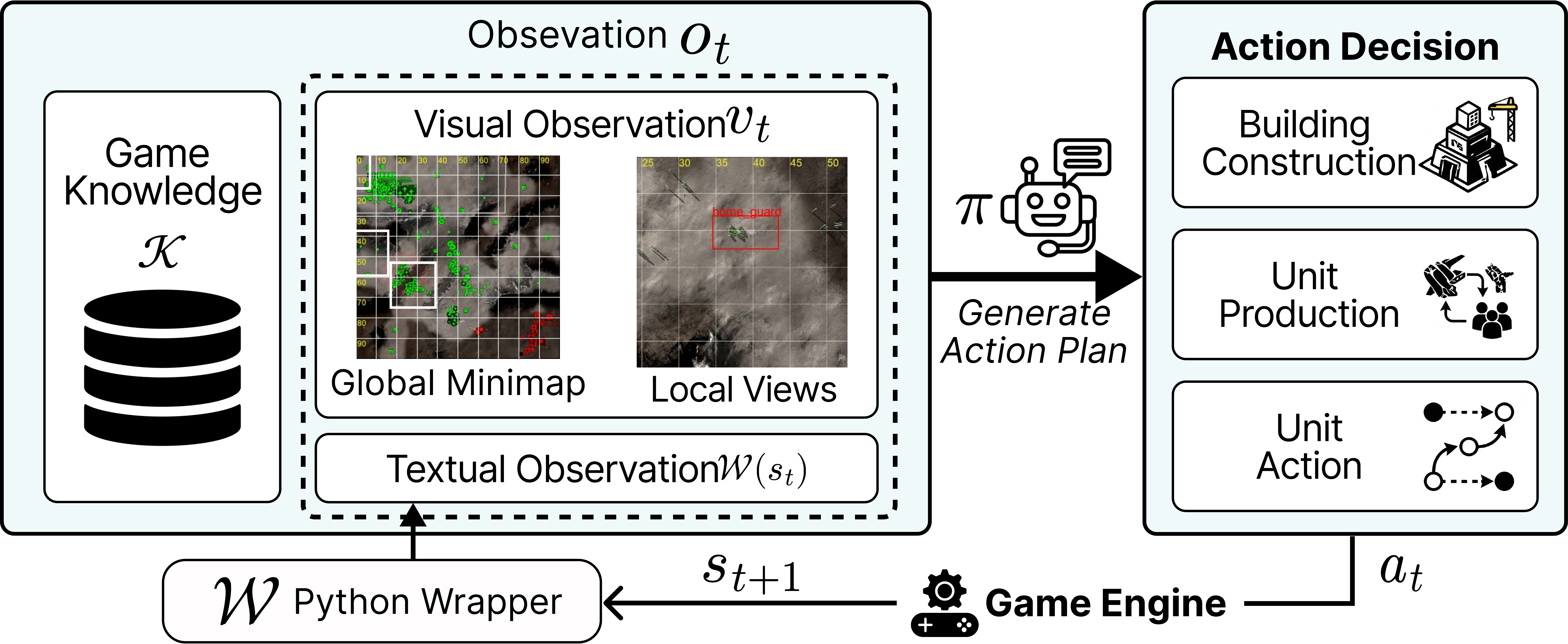}
\caption{\textbf{Overview of the base agent interface.}
At each decision step~$t$, the agent receives a multimodal observation~$o_t$ consisting of visual observations~$v_t$ (a global minimap and up to four local camera views) and a structured textual observation~$\mathcal{W}(s_t)$ extracted from the game state by a Python wrapper.
Together with static game knowledge~$\mathcal{K}$, the VLM policy~$\pi$ generates an action~$a_t$ over three action categories: building construction, unit production, and unit action.
The action is executed by the game environment, producing the next state~$s_{t+1}$ and continuing the observe--decide--act loop.}
\label{fig:agent_interface}
\end{figure}

Fig.~\ref{fig:agent_interface} illustrates the base agent interface shared across all evaluation settings in \benchmark.
This section describes the interface (Sec.~\ref{supp:interface:core}), covering the observe--decide--act loop, the observation space, and the action space, followed by the static game knowledge~$\mathcal{K}$ provided to the agent (Sec.~\ref{supp:interface:knowledge}).

\subsection{Game Interface}
\label{supp:interface:core}

\paragraph{Observe--decide--act loop.}
Every scenario in \benchmark follows a common observe--decide--act loop.
Before each game begins, the agent receives static game knowledge~$\mathcal{K}$ (Sec.~\ref{supp:interface:knowledge}), which contains the scenario description, available units and buildings, and team configuration.
At each decision step~$t$, the environment constructs a multimodal observation
\begin{equation}
o_t = \bigl(v_t,\;\mathcal{W}(s_t)\bigr),
\end{equation}
where $v_t$ denotes the visual observations and $\mathcal{W}(s_t)$ denotes the structured textual observation extracted from the engine state~$s_t$ by a Python wrapper~$\mathcal{W}$.
The agent policy~$\pi$, instantiated by a VLM, then selects an action conditioned on the current observation and static game knowledge:
\begin{equation}
a_t = \pi(o_t \mid \mathcal{K}), \qquad
s_{t+1} \leftarrow \mathrm{Env}(s_t, a_t).
\end{equation}
This process repeats at fixed decision intervals, and the environment pauses during model inference so that evaluation reflects strategic decision quality rather than inference latency or reaction speed.

\paragraph{Visual observation $v_t$.}
The visual input comprises one \emph{global minimap} and up to four \emph{local camera views}.
The minimap provides a bird's-eye overview of the battlefield, while the local views provide higher-resolution observations of strategically relevant regions selected per decision step.
Candidate camera locations are determined according to the following priority order:
(i)~locations of recent combat events,
(ii)~the commander's position when its health falls below 50\%,
(iii)~visible enemy force clusters,
(iv)~allied unit clusters,
(v)~the commander's position when it is not in danger, and
(vi)~the allied home base.
Candidates are processed in this order, and spatially similar locations are merged to avoid redundant observations.
The four highest-priority distinct locations are then selected as local camera views, or fewer when fewer than four distinct candidates are available.
When Fog-of-War (FoW) is enabled, both the minimap and the local views are restricted to information available within allied line of sight, which prevents the visual observations from revealing hidden enemy activity and introduces partial observability into the agent's decision process.

\paragraph{Textual observation $\mathcal{W}(s_t)$.}
The Python wrapper~$\mathcal{W}$ converts the engine state~$s_t$ into a structured textual representation suitable for the VLM.
It includes metal and energy levels; the type, position, and health of allied and currently visible enemy units; ongoing construction and production queues; and the current game time.
The two channels are therefore complementary, as the visual channels capture spatial battlefield structure while the textual channel provides structured game-state information for decision-making.

\paragraph{Action space.}
The base action space consists of three categories:
\begin{itemize}[leftmargin=1.5em]
\item \textbf{Building construction}: the agent selects a constructor, a building type, and a target location.
\item \textbf{Unit production}: the agent selects a production building and a unit type to produce.
\item \textbf{Unit action}: the agent selects one or more units and specifies an action and, when required, a target location.
\end{itemize}
All spatial locations are represented on a normalized $(0,0)$--$(100,100)$ coordinate grid.
Once an action is issued, the game engine handles low-level execution, including pathfinding, collision avoidance, and other unit-level mechanics.

\subsection{Static Game Knowledge $\mathcal{K}$}
\label{supp:interface:knowledge}

The static game knowledge~$\mathcal{K}$ is provided to the agent once before each game and remains fixed throughout the episode.
It consists of three components:
(i)~a scenario description specifying the task objective and win/loss conditions,
(ii)~a team configuration specifying allied and enemy factions, and
(iii)~a unit and building encyclopedia describing the entities available to the agent.

All experiments in this work use the \textit{Armada} faction for the evaluated agent.
Accordingly, the encyclopedia in~$\mathcal{K}$ contains only Armada-side entities.
The opposing \textit{Cortex} faction may appear as an enemy in the evaluated scenarios, but its unit and building encyclopedia is not included in~$\mathcal{K}$.

\noindent \textbf{Unit and building encyclopedia.}
The encyclopedia catalogs the entities available to the Armada faction and organizes them into eight categories:
\textit{Bot}, \textit{Vehicle}, \textit{Air}, \textit{Sea}, \textit{Hover}, \textit{Factory}, \textit{Defense}, and \textit{Building}, where \textit{Building} includes economy and utility structures.
Each entry contains the entity name, internal engine code, technology tier (T1, T2, or T3), and a natural-language description of its role and capabilities.
For constructor units and production buildings, each entry provides a set of \emph{build options} specifying the structures or units that the entity can construct or produce.

\begin{table}[t]
\centering
\caption{\textbf{Summary of the unit and building encyclopedia in $\mathcal{K}$.}
The table reports the number of Armada entities in each category and technology tier.}
\vspace{0.5em}
\label{tab:knowledge_summary}
\small
\begin{tabular}{lcccr}
\toprule
\textbf{Category} & \textbf{T1} & \textbf{T2} & \textbf{T3} & \textbf{Total} \\
\midrule
Bot       & 9  & 17 & 4 & 30 \\
Vehicle   & 10 & 12 & 1 & 23 \\
Air       & 13 & 10 & 0 & 23 \\
Sea       & 7  & 12 & 0 & 19 \\
Hover     & 5  & 0  & 1 & 6  \\
Factory   & 7  & 5  & 1 & 13 \\
Defense   & 17 & 12 & 0 & 29 \\
Building  & 18 & 16 & 0 & 34 \\
\midrule
\textbf{Total} & \textbf{86} & \textbf{84} & \textbf{7} & \textbf{177} \\
\bottomrule
\end{tabular}
\end{table}

\noindent \textbf{Technology tiers.}
Units and buildings are organized into three technology tiers corresponding to BAR's progression system:
\begin{itemize}[leftmargin=1.5em]
\item \textbf{T1 (Tech~1)}: Basic units and structures available during the early stage of the game. They are relatively inexpensive and quick to produce, forming the foundation of early-game armies and economies (\eg, Pawn infantry bot, Solar Collector, and Bot Lab).
\item \textbf{T2 (Tech~2)}: Advanced units and structures unlocked through higher-tier factories and constructors. They are generally more capable and specialized but require greater resource investment (\eg, Sharpshooter sniper bot, Fusion Reactor, and Advanced Bot Lab).
\item \textbf{T3 (Tech~3)}: Experimental units produced through the Experimental Gantry. These units provide substantial combat capability at high resource cost (\eg, Titan, Thor, and Razorback).
\end{itemize}

\noindent \textbf{Category coverage.}
Table~\ref{tab:knowledge_summary} summarizes the number of entities in each category and technology tier.
The encyclopedia contains 177 Armada entities in total, spanning combat units, constructors, factories, defensive structures, and economy and utility buildings.
The complete encyclopedia, including natural-language descriptions and build options, is provided to the agent as part of~$\mathcal{K}$.

\noindent \textbf{Factories and production chains.}
A key component of the encyclopedia is the production dependency structure encoded through each entity's build options.
Each factory specifies the units it can produce.
For example, the Bot Lab (T1) produces basic bot units such as Pawn, Tick, and Mace, whereas the Advanced Bot Lab (T2) provides access to more advanced units such as Gunslinger, Sharpshooter, and Fatboy.
At the highest tier, the Experimental Gantry (T3) produces six experimental units: Vanguard, Titan, Marauder, and Razorback as bots, Thor as a vehicle, and Lunkhead as a hover unit.

Constructor units similarly specify the buildings and structures they can construct.
For example, the Construction Bot and Advanced Construction Vehicle provide access to different portions of the building technology tree.
Together, these factory and constructor dependencies define the production and technology progression that the agent must navigate when planning economic development, unit production, and technological advancement.


\section{Details of Full Games}
\label{supp:full_game}

\subsection{Evaluation Setup}
\label{supp:fullgame_setup}

\noindent \textbf{Faction configuration.}
All players in every matchup use the Armada faction.
Although the environment also supports Cortex, using a single faction removes inter-faction asymmetries, so that performance differences reflect agent behavior rather than faction-specific advantages.

\begin{figure}[t]
    \centering
    \includegraphics[width=0.9\linewidth]{supple_figures/fig9_map_layout.pdf}
    \caption{\textbf{Map layouts and starting positions.}
    Blue markers indicate the starting positions of the agent's team, including allied AIs, and markers in other colors indicate enemy starting positions.}
    \label{fig:map_layout}
\end{figure}

\noindent \textbf{Built-in AI difficulty.}
All slots other than the evaluated agent, both allied and enemy, are filled by BAR's built-in AI, BARbarian~\citep{bar_repo}.
\benchmark defines three difficulty levels, Lv.1, Lv.2, and Lv.3, which correspond to its Easy, Normal, and Hard settings, respectively.

\noindent \textbf{Map configurations.}
Each matchup is played on a fixed map (Fig.~\ref{fig:map_layout}), with map sizes given in BAR's standard map-size units.
The 1v1 matchup uses \textit{BarR 1.1} ($8\times8$), the 2v2 matchup uses \textit{Pinewood Derby v1} ($12\times6$), the 3v4 matchup uses \textit{Wanderlust 2.1} ($10\times8$), and the 1v1v1v1 matchup uses \textit{Center Command BAR v1.0} ($16\times8$).

\begin{figure}[t]
    \centering
    \includegraphics[width=\linewidth]{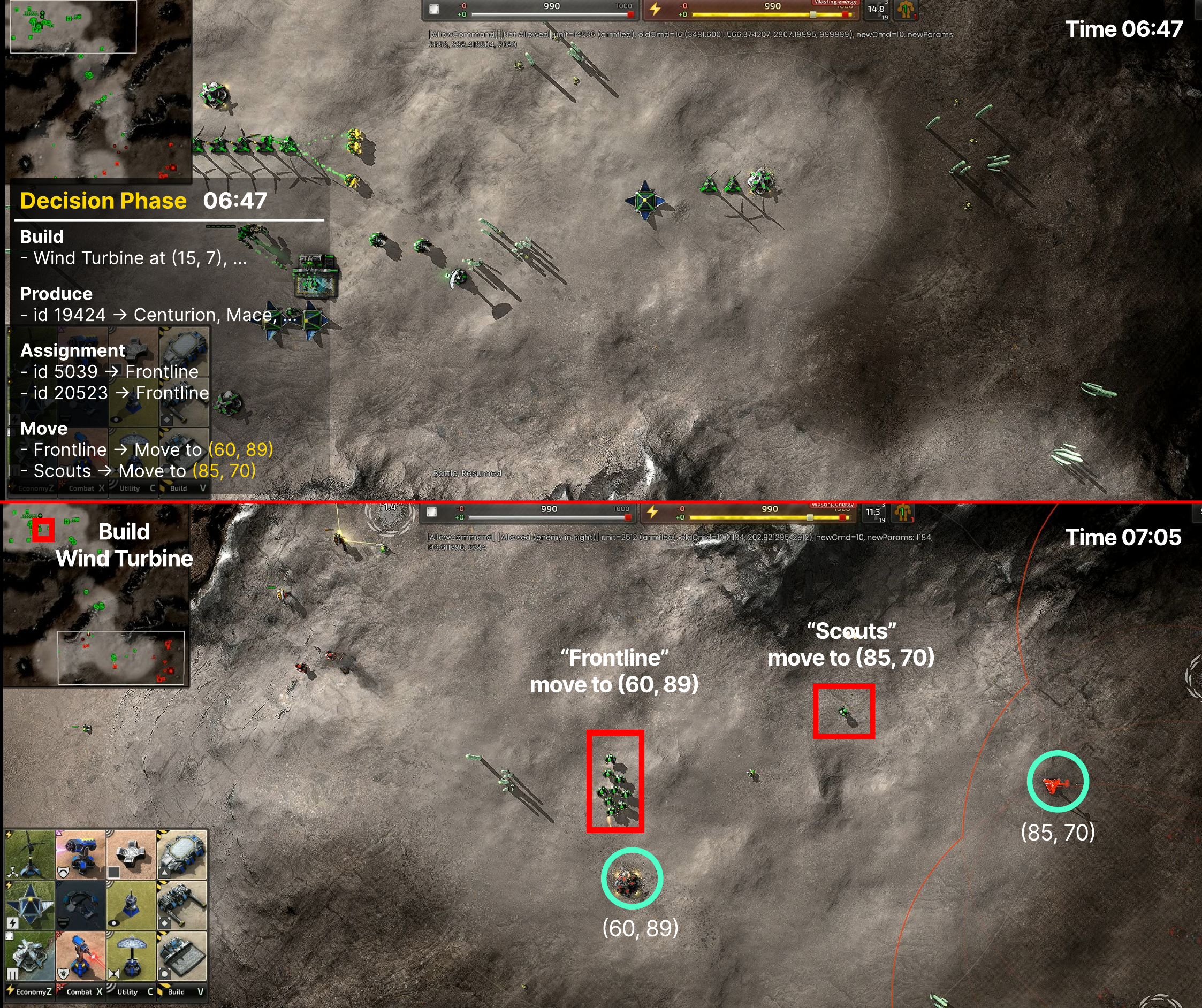}
    \caption{\textbf{Example of decision execution in a full game.}
    At 06:47 (top), the agent issues high-level decisions, including building construction, unit production, and group movement commands with target coordinates.
    At 07:05 (bottom), the frontline units advance toward $(60, 89)$ and the scouts move toward $(85, 70)$, confirming that the issued commands are executed in game.}
    \label{fig:fullgame_quali}
\end{figure}

\begin{table}[p]
\centering
\caption{\textbf{Full-game evaluation results with 90\% Wilson score confidence intervals.}
Each cell reports the win rate over 30 runs; the upper limit of the interval is shown in red superscript and the lower limit in green subscript. Human results are omitted since ten runs yield intervals too wide to be informative.}
\vspace{0.5em}
\label{tab:fullgame_results_ci}
\footnotesize
\resizebox{\ifdim\width>\linewidth\linewidth\else\width\fi}{!}{%
\begin{tabular}{lcccc}
\toprule
\textbf{Model} & \textbf{Duel} (1v1) & \textbf{Sym.\ Team} (2v2) & \textbf{Asymm.\ Team} (3v4) & \textbf{FFA} (1v1v1v1) \\
\midrule
\multicolumn{5}{l}{\textit{Lv.1}} \\
\cmidrule(lr){1-5}
CoS~\citep{ma2024large} & \ci{0.17}{0.08}{0.30} & \ci{0.20}{0.11}{0.34} & \ci{0.00}{0.00}{0.08} & \ci{0.00}{0.00}{0.08} \\
LLM-PySC2~\citep{li2025llmpysc2} & \ci{0.43}{0.30}{0.58} & \ci{0.23}{0.13}{0.38} & \ci{0.00}{0.00}{0.08} & \ci{0.00}{0.00}{0.08} \\
SwarmBrain~\citep{shao2024swarmbrain} & \ci{0.80}{0.66}{0.89} & \ci{0.57}{0.42}{0.70} & \ci{0.00}{0.00}{0.08} & \ci{0.00}{0.00}{0.08} \\
\method{} (Ours) & \ci{1.00}{0.92}{1.00} & \ci{0.83}{0.70}{0.92} & \ci{0.17}{0.08}{0.30} & \ci{0.20}{0.11}{0.34} \\
\cmidrule(lr){1-5}
GPT-5.6-Luna & \ci{0.20}{0.11}{0.34} & \ci{0.13}{0.06}{0.27} & \ci{0.00}{0.00}{0.08} & \ci{0.00}{0.00}{0.08} \\
Qwen3.8-27B & \ci{0.23}{0.13}{0.38} & \ci{0.33}{0.21}{0.48} & \ci{0.00}{0.00}{0.08} & \ci{0.00}{0.00}{0.08} \\
Qwen3.5-397B & \ci{0.47}{0.33}{0.61} & \ci{0.10}{0.04}{0.23} & \ci{0.00}{0.00}{0.08} & \ci{0.00}{0.00}{0.08} \\
Muse-Spark-1.2 & \ci{0.40}{0.27}{0.55} & \ci{0.27}{0.16}{0.41} & \ci{0.00}{0.00}{0.08} & \ci{0.00}{0.00}{0.08} \\
GPT-5.6-Terra & \ci{0.57}{0.42}{0.70} & \ci{0.30}{0.18}{0.45} & \ci{0.00}{0.00}{0.08} & \ci{0.13}{0.06}{0.27} \\
GLM-5.3-Flash & \ci{0.67}{0.52}{0.79} & \ci{0.40}{0.27}{0.55} & \ci{0.00}{0.00}{0.08} & \ci{0.03}{0.01}{0.14} \\
Grok-4.6 & \ci{0.73}{0.59}{0.84} & \ci{0.83}{0.70}{0.92} & \ci{0.00}{0.00}{0.08} & \ci{0.07}{0.02}{0.18} \\
Gemini-3.7-Flash & \ci{1.00}{0.92}{1.00} & \ci{0.83}{0.70}{0.92} & \ci{0.17}{0.08}{0.30} & \ci{0.20}{0.11}{0.34} \\
\midrule
\multicolumn{5}{l}{\textit{Lv.2}} \\
\cmidrule(lr){1-5}
CoS~\citep{ma2024large} & \ci{0.00}{0.00}{0.08} & \ci{0.07}{0.02}{0.18} & \ci{0.00}{0.00}{0.08} & \ci{0.00}{0.00}{0.08} \\
LLM-PySC2~\citep{li2025llmpysc2} & \ci{0.00}{0.00}{0.08} & \ci{0.13}{0.06}{0.27} & \ci{0.00}{0.00}{0.08} & \ci{0.00}{0.00}{0.08} \\
SwarmBrain~\citep{shao2024swarmbrain} & \ci{0.23}{0.13}{0.38} & \ci{0.33}{0.21}{0.48} & \ci{0.00}{0.00}{0.08} & \ci{0.00}{0.00}{0.08} \\
\method{} (Ours) & \ci{0.43}{0.30}{0.58} & \ci{0.40}{0.27}{0.55} & \ci{0.00}{0.00}{0.08} & \ci{0.07}{0.02}{0.18} \\
\cmidrule(lr){1-5}
GPT-5.6-Luna & \ci{0.00}{0.00}{0.08} & \ci{0.10}{0.04}{0.23} & \ci{0.00}{0.00}{0.08} & \ci{0.00}{0.00}{0.08} \\
Qwen3.8-27B & \ci{0.00}{0.00}{0.08} & \ci{0.00}{0.00}{0.08} & \ci{0.00}{0.00}{0.08} & \ci{0.00}{0.00}{0.08} \\
Qwen3.5-397B & \ci{0.00}{0.00}{0.08} & \ci{0.07}{0.02}{0.18} & \ci{0.00}{0.00}{0.08} & \ci{0.00}{0.00}{0.08} \\
Muse-Spark-1.2 & \ci{0.00}{0.00}{0.08} & \ci{0.20}{0.11}{0.34} & \ci{0.00}{0.00}{0.08} & \ci{0.00}{0.00}{0.08} \\
GPT-5.6-Terra & \ci{0.07}{0.02}{0.18} & \ci{0.03}{0.01}{0.14} & \ci{0.00}{0.00}{0.08} & \ci{0.07}{0.02}{0.18} \\
GLM-5.3-Flash & \ci{0.13}{0.06}{0.27} & \ci{0.13}{0.06}{0.27} & \ci{0.00}{0.00}{0.08} & \ci{0.03}{0.01}{0.14} \\
Grok-4.6 & \ci{0.00}{0.00}{0.08} & \ci{0.10}{0.04}{0.23} & \ci{0.00}{0.00}{0.08} & \ci{0.00}{0.00}{0.08} \\
Gemini-3.7-Flash & \ci{0.43}{0.30}{0.58} & \ci{0.40}{0.27}{0.55} & \ci{0.00}{0.00}{0.08} & \ci{0.07}{0.02}{0.18} \\
\midrule
\multicolumn{5}{l}{\textit{Lv.3}} \\
\cmidrule(lr){1-5}
CoS~\citep{ma2024large} & \ci{0.00}{0.00}{0.08} & \ci{0.00}{0.00}{0.08} & \ci{0.00}{0.00}{0.08} & \ci{0.00}{0.00}{0.08} \\
LLM-PySC2~\citep{li2025llmpysc2} & \ci{0.00}{0.00}{0.08} & \ci{0.10}{0.04}{0.23} & \ci{0.00}{0.00}{0.08} & \ci{0.00}{0.00}{0.08} \\
SwarmBrain~\citep{shao2024swarmbrain} & \ci{0.13}{0.06}{0.27} & \ci{0.20}{0.11}{0.34} & \ci{0.00}{0.00}{0.08} & \ci{0.00}{0.00}{0.08} \\
\method{} (Ours) & \ci{0.30}{0.18}{0.45} & \ci{0.20}{0.11}{0.34} & \ci{0.00}{0.00}{0.08} & \ci{0.00}{0.00}{0.08} \\
\cmidrule(lr){1-5}
GPT-5.6-Luna & \ci{0.00}{0.00}{0.08} & \ci{0.00}{0.00}{0.08} & \ci{0.00}{0.00}{0.08} & \ci{0.00}{0.00}{0.08} \\
Qwen3.8-27B & \ci{0.00}{0.00}{0.08} & \ci{0.00}{0.00}{0.08} & \ci{0.00}{0.00}{0.08} & \ci{0.00}{0.00}{0.08} \\
Qwen3.5-397B & \ci{0.00}{0.00}{0.08} & \ci{0.03}{0.01}{0.14} & \ci{0.00}{0.00}{0.08} & \ci{0.00}{0.00}{0.08} \\
Muse-Spark-1.2 & \ci{0.00}{0.00}{0.08} & \ci{0.07}{0.02}{0.18} & \ci{0.00}{0.00}{0.08} & \ci{0.00}{0.00}{0.08} \\
GPT-5.6-Terra & \ci{0.00}{0.00}{0.08} & \ci{0.07}{0.02}{0.18} & \ci{0.00}{0.00}{0.08} & \ci{0.00}{0.00}{0.08} \\
GLM-5.3-Flash & \ci{0.10}{0.04}{0.23} & \ci{0.07}{0.02}{0.18} & \ci{0.00}{0.00}{0.08} & \ci{0.00}{0.00}{0.08} \\
Grok-4.6 & \ci{0.00}{0.00}{0.08} & \ci{0.00}{0.00}{0.08} & \ci{0.00}{0.00}{0.08} & \ci{0.00}{0.00}{0.08} \\
Gemini-3.7-Flash & \ci{0.30}{0.18}{0.45} & \ci{0.20}{0.11}{0.34} & \ci{0.00}{0.00}{0.08} & \ci{0.00}{0.00}{0.08} \\
\bottomrule
\end{tabular}%
}
\end{table}

\begin{figure}[t]
    \centering
    \includegraphics[width=\linewidth]{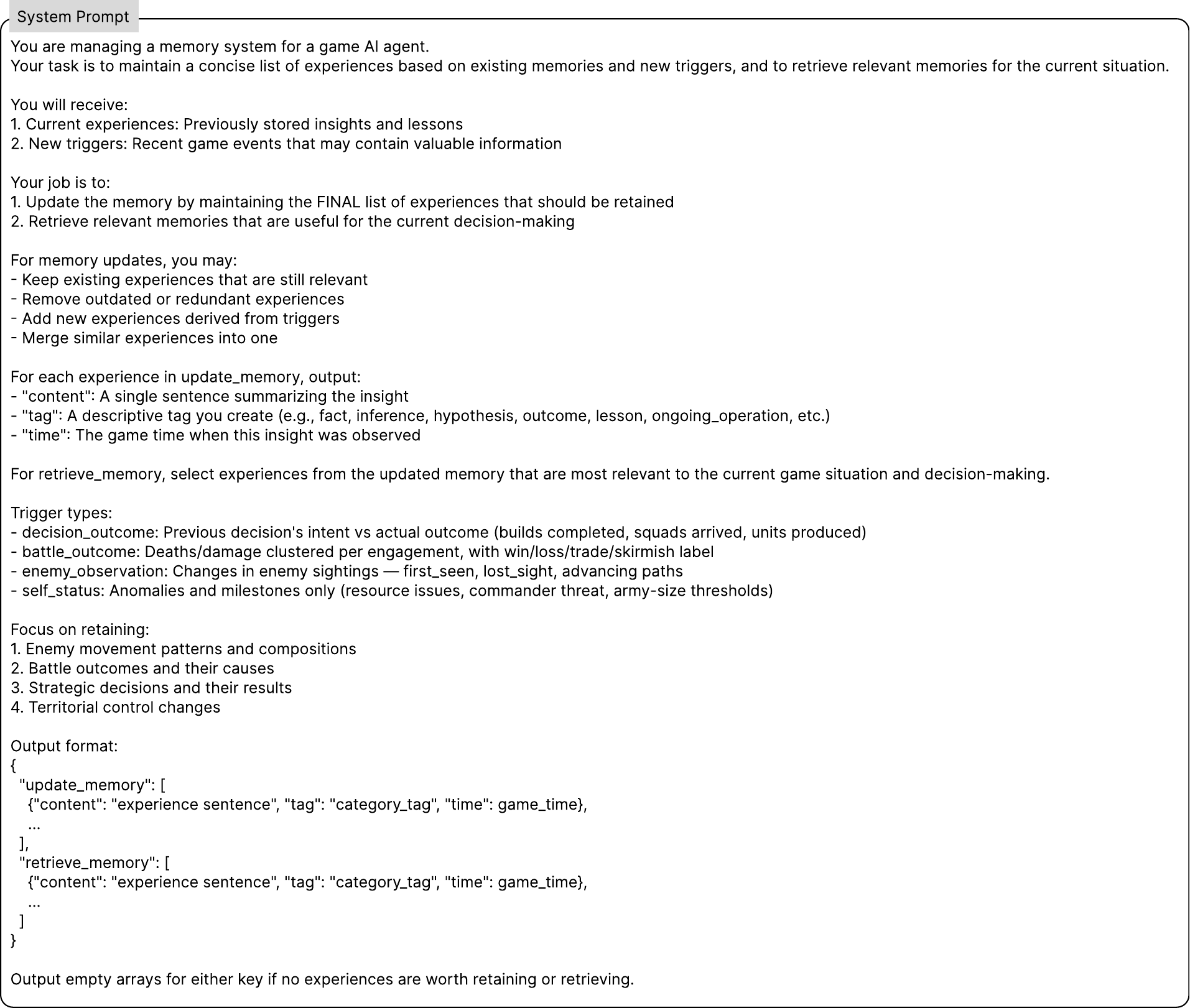}
    \vspace{-1.0em}
    \caption{\textbf{System prompt for the memory phase.} The system prompt used in the memory phase to update the agent's memory and retrieve relevant experiences before the decision phase.}
    \label{fig:memory_system}
\end{figure}

\begin{figure}[t]
    \centering
    \includegraphics[width=\linewidth]{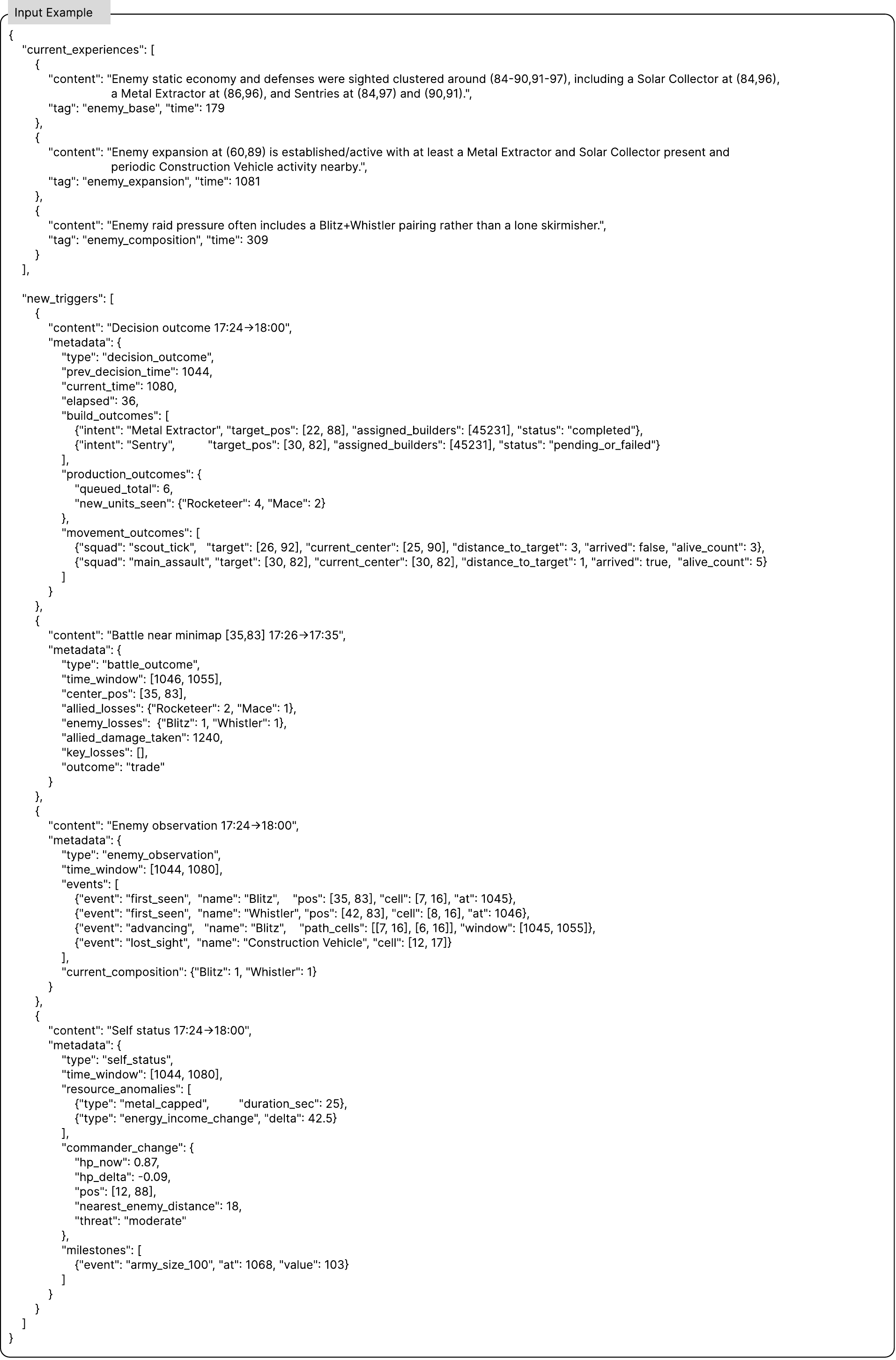}
    \vspace{-1.0em}
    \caption{\textbf{Input example for the memory phase.} The input to the memory phase consists of the currently retained experiences and newly emitted triggers since the last decision — including decision outcomes, battle outcomes, enemy observations, and self-status changes.}
    \label{fig:memory_input}
\end{figure}

\begin{figure}[t]
    \centering
    \includegraphics[width=\linewidth]{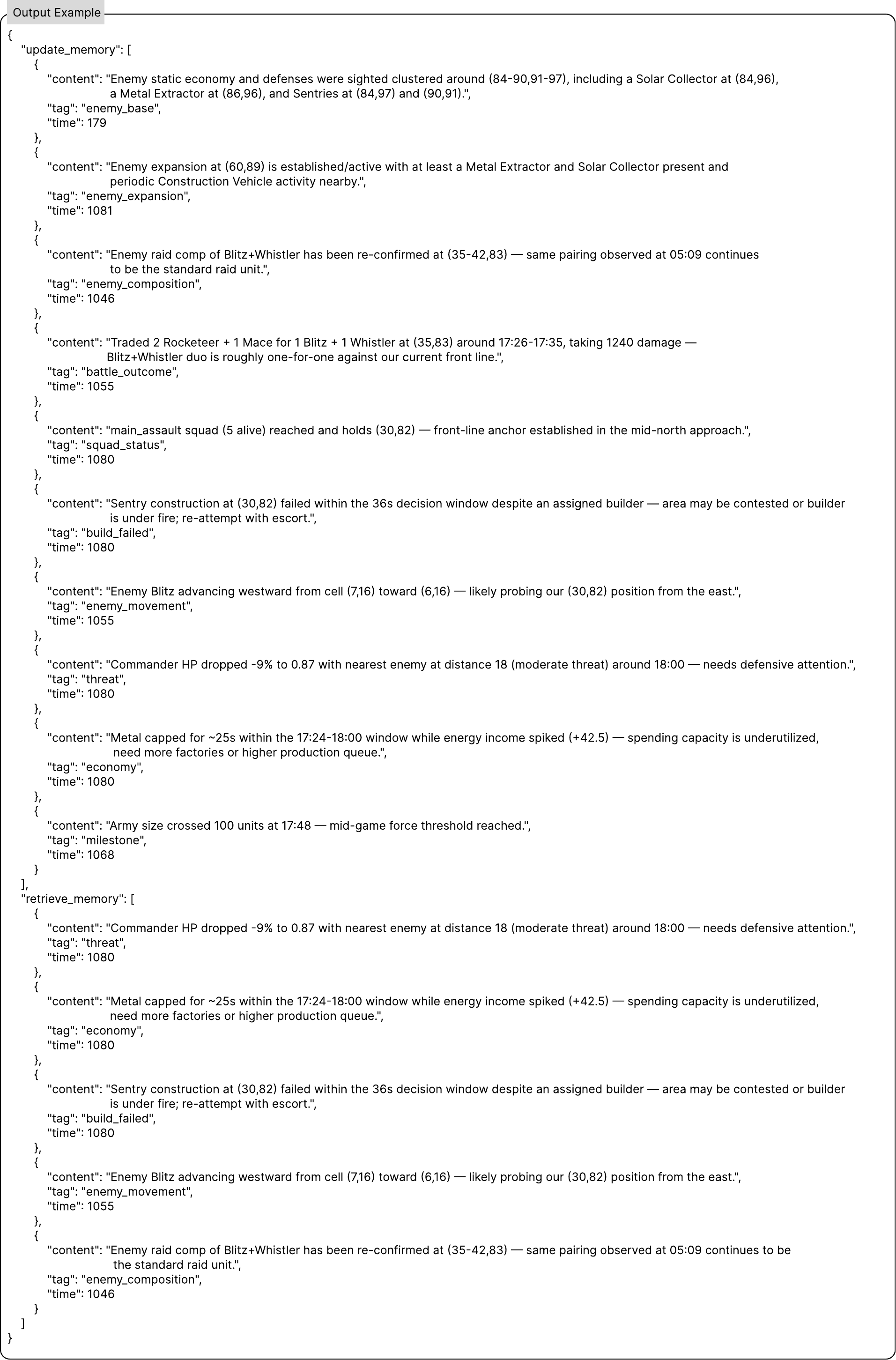}
    \vspace{-1.0em}
    \caption{\textbf{Output example for the memory phase.} The output of the memory phase consists of an updated memory state that incorporates the new triggers and a subset of those experiences retrieved for the upcoming decision.}
    \label{fig:memory_output}
\end{figure}

\begin{figure}[h]
    \centering
    \includegraphics[width=0.96\linewidth]{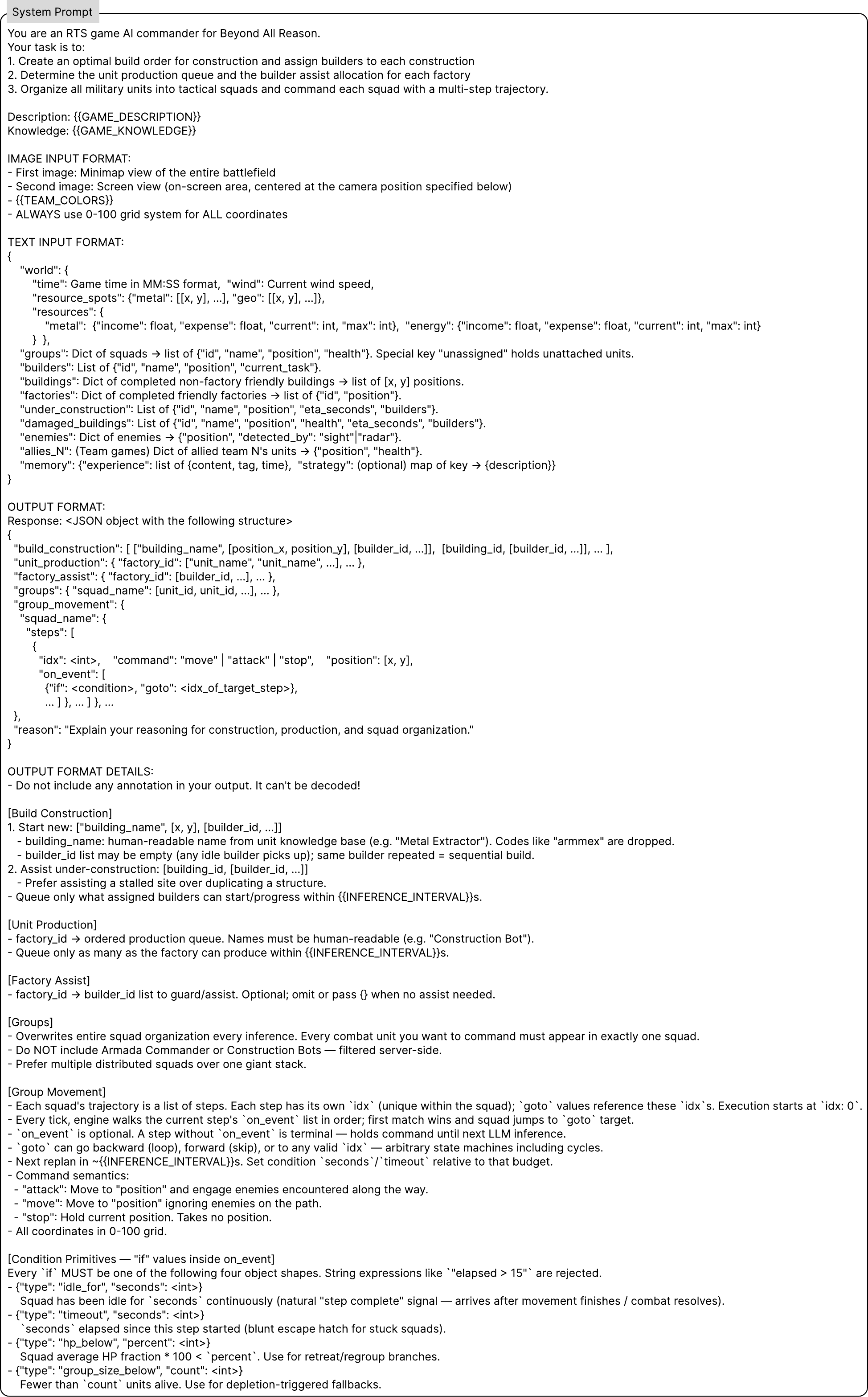}
    \vspace{-1.0em}
    \caption{\textbf{System prompt for the decision phase.} Placeholders (denoted with double braces \{\{...\}\}) in the system prompt are replaced with the game description, unit knowledge, team color specification, and inference interval.}
    \label{fig:decision_system}
\end{figure}

\begin{figure}[t]
    \centering
    \includegraphics[width=\linewidth]{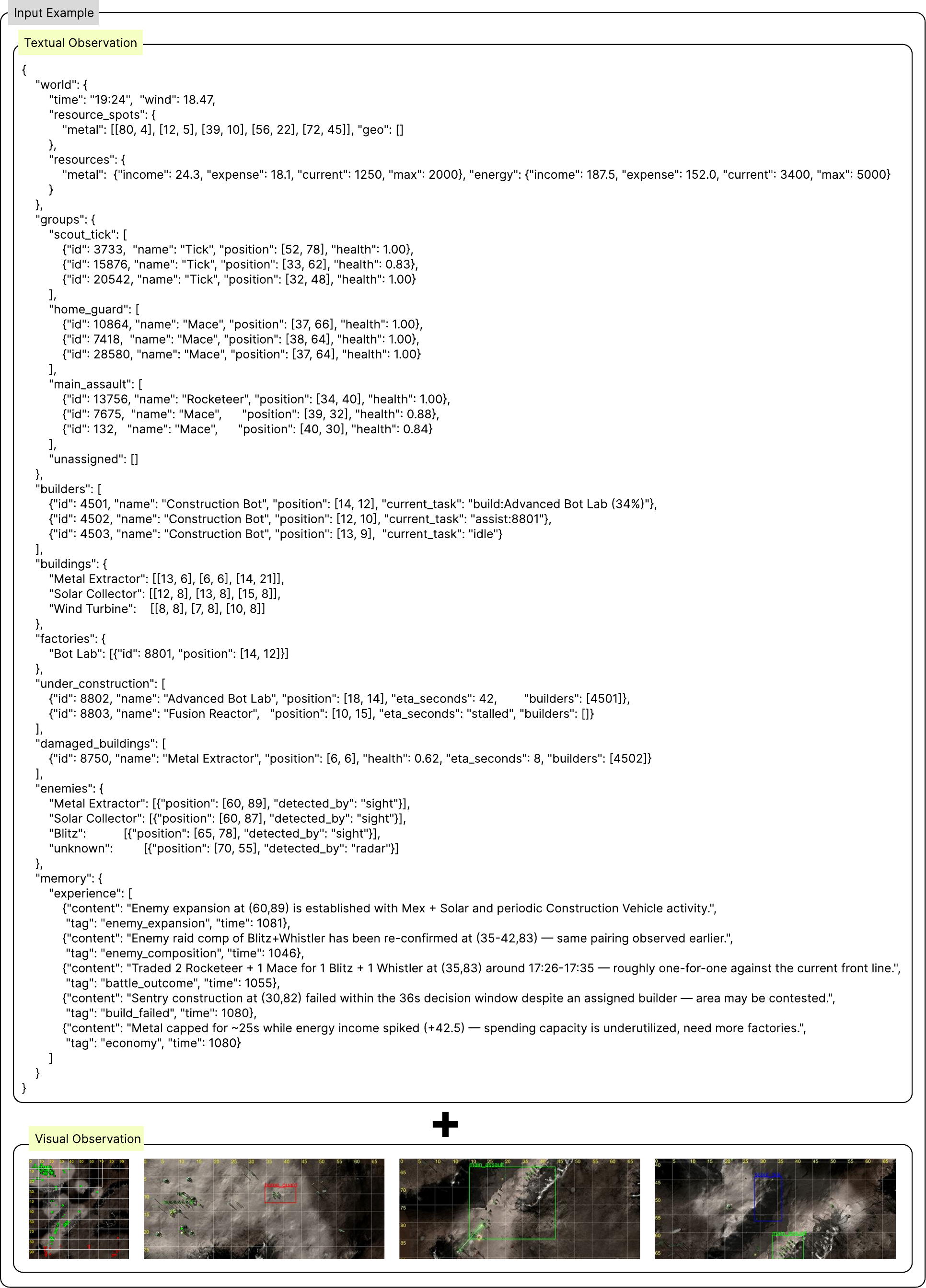}
    \vspace{-1.0em}
    \caption{\textbf{Input example for the decision phase.} The input to the decision phase consists of textual and visual observations of the current game state, together with experiences retrieved by the memory phase.}
    \label{fig:decision_input}
\end{figure}

\begin{figure}[t]
    \centering
    \includegraphics[width=\linewidth]{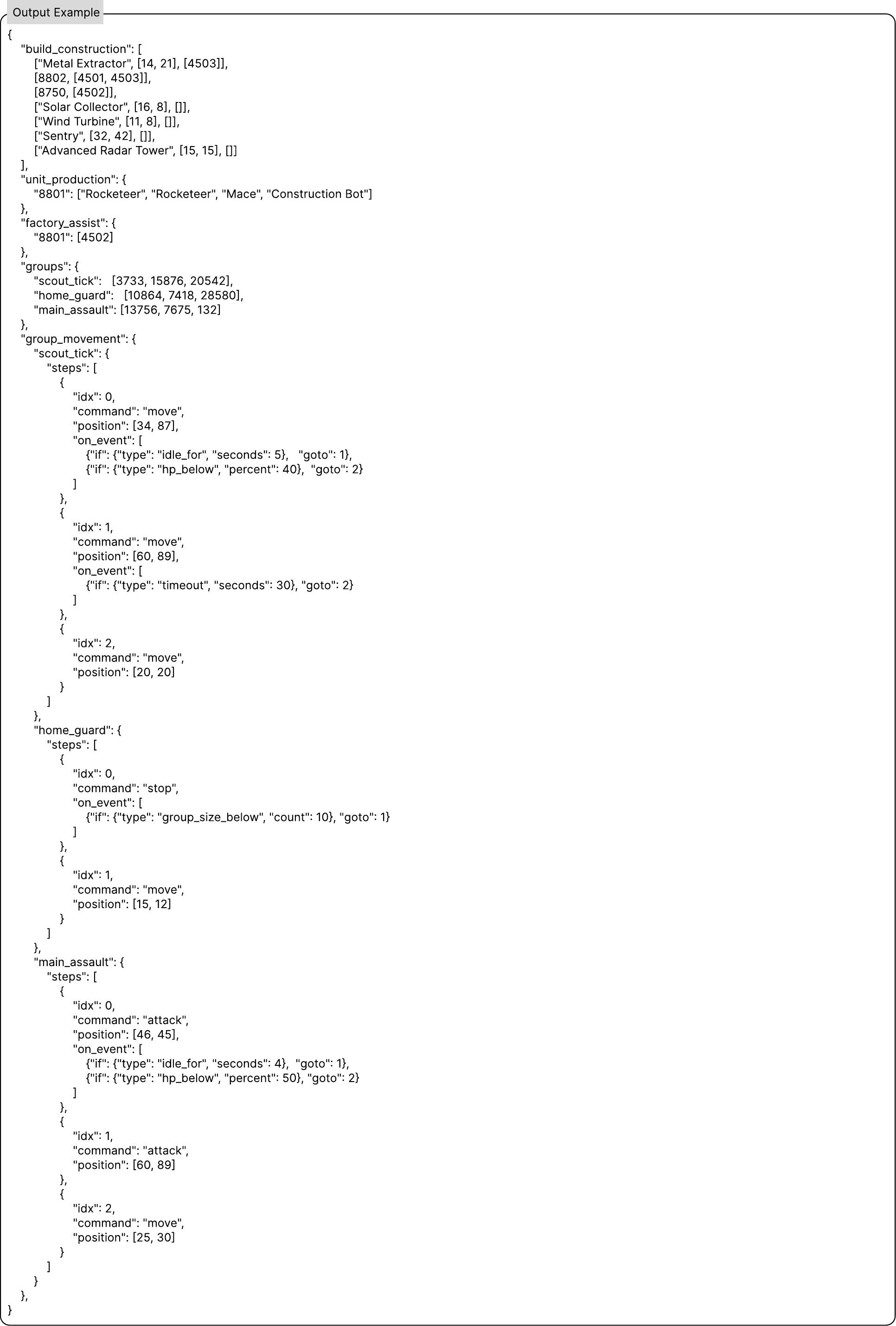}
    \vspace{-1.0em}
    \caption{\textbf{Output example for the decision phase.} The output of the decision phase specifies actions across five decision categories: constructing buildings, producing units, assigning builders to assist factories, organizing units into squads, and specifying multi-step trajectories for each squad.}
    \label{fig:decision_output}
\end{figure}

\noindent \textbf{Prompts and examples.}
\label{supp:agent_prompts}
Figures~\ref{fig:memory_system}, \ref{fig:memory_input}, and~\ref{fig:memory_output} show the system prompt, an input example, and an output example for the memory phase, respectively.
Figures~\ref{fig:decision_system}, \ref{fig:decision_input}, and~\ref{fig:decision_output} show the corresponding prompt and examples for the decision phase.

\noindent \textbf{Qualitative example.}
Figure~\ref{fig:fullgame_quali} shows a mid-game moment from a 1v1 full game, in which the agent expands its economy while preparing for upcoming combat.

\subsection{Results with Confidence Intervals}
\label{supp:fullgame_ci}
Tab.~\ref{tab:fullgame_results_ci} repeats the full-game win rates of Tab.~\ref{tab:main_results} together with 90\% Wilson score confidence intervals over 30 runs per setting.
A win rate of zero over 30 runs still admits a true win rate of up to about 8\%, so the many zero entries at higher difficulty levels indicate that wins are rare rather than impossible.

\subsection{Agent Comparison with Other Backbones}
\label{supp:agent_backbones}
The agent comparison in Tab.~\ref{tab:main_results} uses Gemini-3.7-Flash, which matches or exceeds all other backbones.
To check whether the comparison depends on this choice, we repeat it with GPT-5.6-Terra and GLM-5.3-Flash under the same protocol (Tab.~\ref{tab:agent_backbones}).
The trends remain similar, although absolute win rates are lower.
As with Gemini-3.7-Flash, WR declines as the built-in AI level increases and stays low in Asymmetric Team and Free-for-All.
\method also matches or outperforms the baselines in nearly every setting, except Symmetric Team Lv.1 and Lv.2 with GPT-5.6-Terra.

\begin{table}[t]
\centering
\caption{\textbf{Agent comparison with other backbones.}
WR of each agent with GPT-5.6-Terra and GLM-5.3-Flash as backbones, under the same protocol as Tab.~\ref{tab:main_results} (30 runs per setting).
Best and second-best non-zero results within each backbone are shown in \textbf{bold} and \underline{underlined}, respectively; ties share the same formatting.}
\vspace{0.5em}
\label{tab:agent_backbones}
\resizebox{\linewidth}{!}{%
\begin{tabular}{lcccccccccccc}
\toprule
\multirow{2}{*}{\textbf{Model}}
& \multicolumn{3}{c}{\textit{Duel} (1v1)}
& \multicolumn{3}{c}{\textit{Symmetric Team} (2v2)}
& \multicolumn{3}{c}{\textit{Asymm.\ Team} (3v4)}
& \multicolumn{3}{c}{\textit{Free-for-All} (1v1v1v1)} \\
\cmidrule(lr){2-4}
\cmidrule(lr){5-7}
\cmidrule(lr){8-10}
\cmidrule(lr){11-13}

& Lv.1 & Lv.2 & Lv.3
& Lv.1 & Lv.2 & Lv.3
& Lv.1 & Lv.2 & Lv.3
& Lv.1 & Lv.2 & Lv.3 \\

\midrule
\multicolumn{13}{l}{\textit{Backbone = GPT-5.6-Terra}} \\
\midrule

CoS~\citep{ma2024large}
& 0.03 & 0.00 & 0.00
& 0.00 & 0.00 & 0.00
& 0.00 & 0.00 & 0.00
& 0.00 & 0.00 & 0.00 \\

LLM-PySC2~\citep{li2025llmpysc2}
& \underline{0.13} & 0.00 & 0.00
& \textbf{0.37} & \textbf{0.07} & 0.00
& 0.00 & 0.00 & 0.00
& 0.03 & 0.00 & 0.00 \\

SwarmBrain~\citep{shao2024swarmbrain}
& 0.10 & 0.00 & 0.00
& 0.10 & 0.00 & 0.00
& 0.00 & 0.00 & 0.00
& \underline{0.07} & 0.00 & 0.00 \\

\method{} (Ours)
& \textbf{0.57} & \textbf{0.07} & 0.00
& \underline{0.30} & \underline{0.03} & \textbf{0.07}
& 0.00 & 0.00 & 0.00
& \textbf{0.13} & \textbf{0.07} & 0.00 \\

\midrule
\multicolumn{13}{l}{\textit{Backbone = GLM-5.3-Flash}} \\
\midrule

CoS~\citep{ma2024large}
& 0.00 & 0.00 & 0.00
& 0.10 & 0.07 & 0.00
& 0.00 & 0.00 & 0.00
& 0.00 & 0.00 & 0.00 \\

LLM-PySC2~\citep{li2025llmpysc2}
& 0.07 & 0.00 & 0.00
& \underline{0.37} & 0.07 & 0.00
& 0.00 & 0.00 & 0.00
& 0.00 & 0.00 & 0.00 \\

SwarmBrain~\citep{shao2024swarmbrain}
& \underline{0.53} & \underline{0.03} & 0.00
& \underline{0.37} & \underline{0.10} & 0.00
& 0.00 & 0.00 & 0.00
& 0.00 & 0.00 & 0.00 \\

\method{} (Ours)
& \textbf{0.67} & \textbf{0.13} & \textbf{0.10}
& \textbf{0.40} & \textbf{0.13} & \textbf{0.07}
& 0.00 & 0.00 & 0.00
& \textbf{0.03} & \textbf{0.03} & 0.00 \\

\bottomrule
\end{tabular}%
}
\end{table}

\section{Details of Diagnostic Mini-Games}
\label{supp:minigames}

\begin{figure}[t]
    \centering
    \includegraphics[width=0.9\linewidth]{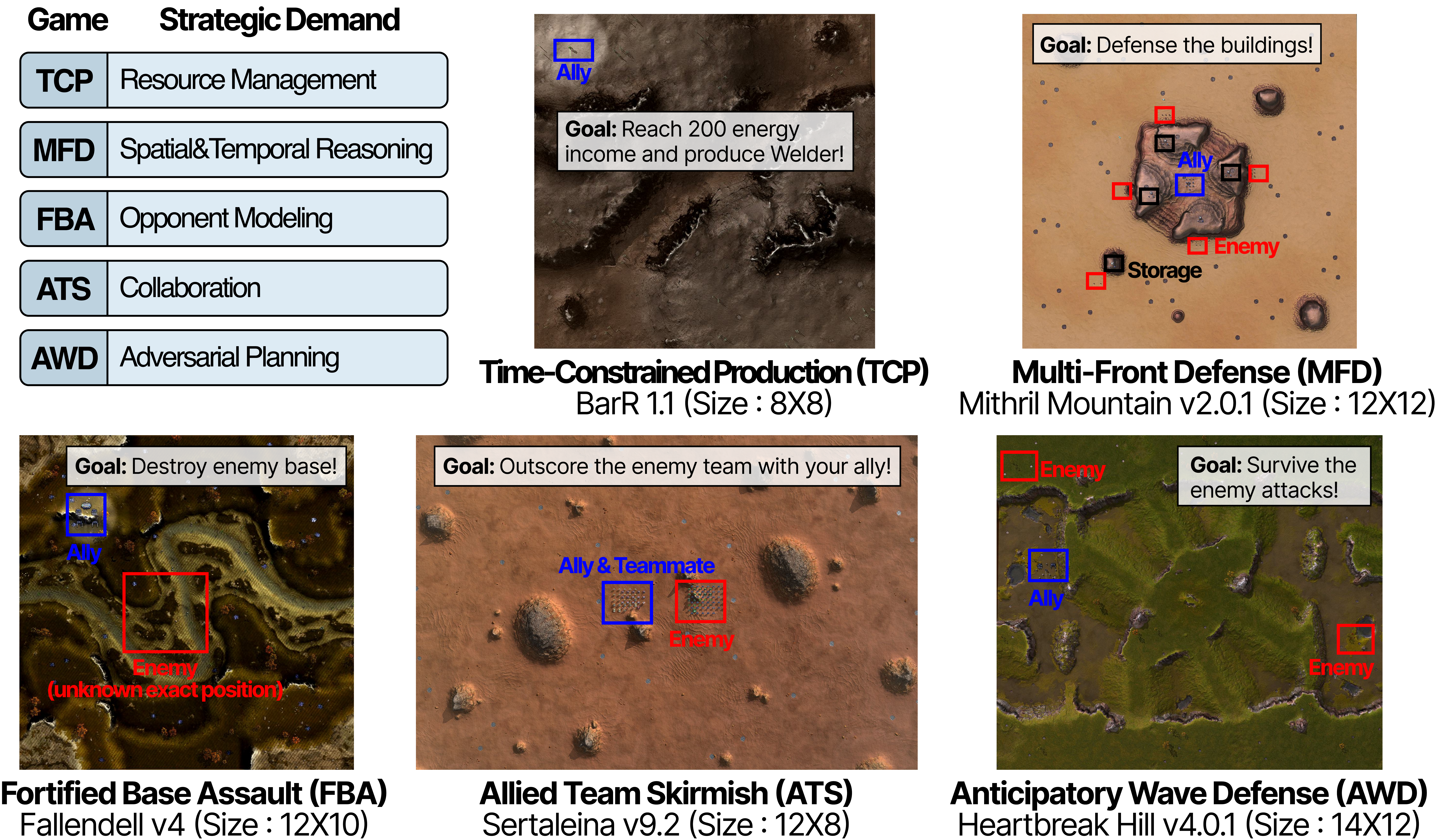}
    \caption{\textbf{Initial configurations of the diagnostic mini-games.} The figure illustrates the map layouts and initial unit placements for the five diagnostic mini-games used in our evaluation. Blue markers indicate the controlled agent and its teammates, while red markers denote enemy units or enemy base locations. In scenarios where the exact enemy position is unknown, the red box indicates the region where the enemy may appear.}
    \label{fig:minigame_layout}
    \vspace{-1.0em}
\end{figure}

We design five diagnostic mini-games, each targeting a distinct strategic demand from a taxonomy of RTS AI challenges~\citep{buro2003real}: \textit{resource management} (RM), \textit{spatial \& temporal reasoning} (S\&T), \textit{opponent modeling} (OM), \textit{collaboration} (COL), and \textit{adversarial planning} (AP).
Each mini-game emphasizes its focal demand through the objective, the available action types, and the information made visible to the agent, and we describe below what that demand amounts to in BAR.
Their map layouts and initial configurations are illustrated in Fig.~\ref{fig:minigame_layout}.
A sixth demand, \textit{decision making under uncertainty}, is incorporated selectively through Fog-of-War (FoW) rather than as a separate mini-game.
The taxonomy also includes pathfinding, which we do not target, as BAR's engine resolves it automatically once a destination is issued (Sec.~\ref{supp:bar_overview}).

Every mini-game is evaluated at three difficulty levels (Lv.1--3).
Rather than changing the objective, each level scales the difficulty of achieving it, mostly by increasing the \emph{power} of the enemy forces, BAR's internal value scale that converts energy into metal-equivalent units (Sec.~\ref{supp:metrics}).

\subsection{Human Ratings of Strategic Demands}
\label{supp:demand_ratings}

To verify that each mini-game emphasizes its intended strategic demand, we collect ratings from five human raters on how strongly every mini-game elicits each of the five focal demands targeted by the mini-games (Tab.~\ref{tab:eval_overview}), \ie, RM, S\&T, OM, COL, and AP.

\noindent \textbf{Demand definitions.}
Raters receive a brief definition of each demand, adapted from the taxonomy of RTS AI challenges~\citep{buro2003real} and identical across all scenarios.
RM concerns gathering resources and allocating them among economy, production, and technology; S\&T concerns reasoning about positions, terrain, and the timing of actions; OM concerns inferring the opponent's forces or strategy from in-game observations; COL concerns coordinating with allies whose actions the player does not control; and AP concerns planning ahead against the adversary's future actions.
Whereas \citet{buro2003real} frames opponent modeling as learning opponents' weaknesses across games, we define it as inference within an episode, since each mini-game is played as a single episode.
We exclude pathfinding, which BAR handles at the engine level (Sec.~\ref{supp:bar_game_characteristic}), and decision making under uncertainty, which we control through the FoW setting rather than scenario design (Tab.~\ref{tab:eval_overview}).

\noindent \textbf{Protocol.}
The five raters are experienced RTS players familiar with \textit{StarCraft~II} who were not involved in designing the benchmark.
To avoid revealing the intended focal demands, mini-games are presented in a randomized order under neutral labels (Scenario~A--E), and raters are not informed that each mini-game targets a particular demand.
For each scenario, a rater reads player-facing instructions that describe only the mechanics, \ie, the initial setup, win/loss conditions, and available actions, without the design rationale given in Sec.~\ref{sec:minigame_bench}.
The rater then plays one episode at Lv.1 before rating, so that ratings reflect the demands encountered in actual play rather than the instructions alone.
After playing, the rater scores all five demands on a 1--5 scale, where 1 indicates that the demand is irrelevant to winning and 5 that it is central.
Raters score independently, without access to other raters' scores.
Fig.~\ref{fig:minigame_demands} reports the mean over the five raters, and Tab.~\ref{tab:demand_ratings} additionally reports the standard deviation of each cell.

\noindent \textbf{Alignment with focal demands.}
Every mini-game receives its highest mean rating on its focal demand, ranging from 4.2 to 4.8, and this also holds for each rater individually: in all 25 rater--mini-game pairs, the focal demand receives the highest score, strictly so in 19 and tied for highest in the remaining 6.
Focal demands average 4.52 against 2.22 for the others, with per-mini-game margins of 1.0--2.4 over the next highest demand.

\noindent \textbf{Inter-rater agreement.}
Raters largely agree with one another, with a mean pairwise correlation of 0.83 between raters' 25 ratings (five mini-games $\times$ five demands) and a per-cell standard deviation of 0.51 on average (at most 0.89).

\noindent \textbf{Off-diagonal patterns.}
Off-diagonal ratings remain non-negligible, \eg, ATS scores 3.6 on S\&T and MFD scores 3.4 on AP, supporting that each mini-game emphasizes one demand while preserving the coupled decision-making of RTS gameplay rather than isolating a single capability.
Collaboration is the most sharply separated demand, scoring 4.6 on ATS but at most 1.2 elsewhere, as ATS is the only mini-game with an ally.
FBA and ATS show the smallest margins over the next highest demand (1.0 each), as assaulting a fortified base also draws on RM, S\&T, and AP (3.0--3.2), and fighting alongside an ally also draws on S\&T (3.6).

\begin{table}[t]
\centering
\caption{\textbf{Human ratings of strategic demands.}
Mean $\pm$ standard deviation over five raters, on a 1--5 scale.
Each mini-game's focal demand is in \textbf{bold}.}
\vspace{0.5em}
\label{tab:demand_ratings}
\footnotesize
\begin{tabular}{lccccc}
\toprule
\textbf{Mini-game} & \textbf{RM} & \textbf{S\&T} & \textbf{OM} & \textbf{COL} & \textbf{AP} \\
\midrule
TCP & \textbf{4.8}\,$\pm$\,0.45 & 1.8\,$\pm$\,0.84 & 1.2\,$\pm$\,0.45 & 1.0\,$\pm$\,0.00 & 2.4\,$\pm$\,0.89 \\
MFD & 1.2\,$\pm$\,0.45 & \textbf{4.6}\,$\pm$\,0.55 & 2.0\,$\pm$\,0.71 & 1.2\,$\pm$\,0.45 & 3.4\,$\pm$\,0.89 \\
FBA & 3.2\,$\pm$\,0.84 & 3.0\,$\pm$\,0.00 & \textbf{4.2}\,$\pm$\,0.45 & 1.0\,$\pm$\,0.00 & 3.0\,$\pm$\,0.71 \\
ATS & 1.8\,$\pm$\,0.45 & 3.6\,$\pm$\,0.55 & 2.6\,$\pm$\,0.89 & \textbf{4.6}\,$\pm$\,0.55 & 2.6\,$\pm$\,0.55 \\
AWD & 3.0\,$\pm$\,0.71 & 3.2\,$\pm$\,0.45 & 2.2\,$\pm$\,0.45 & 1.0\,$\pm$\,0.00 & \textbf{4.4}\,$\pm$\,0.55 \\
\bottomrule
\end{tabular}
\end{table}

\subsection{Time-Constrained Production (TCP) --- Resource Management}

\noindent \textbf{Strategic demand.}
BAR's streaming economy generates and consumes metal and energy as continuous flows, so income, tech progression, and unit production constantly compete for the same budget (Sec.~\ref{supp:bar_overview}).
Construction and technology dependencies further make the order of these investments consequential, as an advanced unit becomes available only after its prerequisites are complete.

\noindent \textbf{Objective.}
The agent must reach a metal income of at least 20 and an energy income of at least 200 as quickly as possible within a 10-minute limit, while fielding the units required by the difficulty level.
A run counts as a win if all targets are met before the time limit.
The available action categories are \textbf{Build}, \textbf{Prod.}, and \textbf{Action}, and FoW is \textbf{enabled}.

\noindent \textbf{Difficulty levels.}
All levels share the same income targets and time limit, and differ in the units the agent must field.
Lv.1 requires no units beyond the income targets, Lv.2 additionally requires at least three Pawns and two Stouts, and Lv.3 requires at least one Welder, a T2 unit that becomes available only after an additional tech step.

\noindent \textbf{Design rationale.}
At Lv.2 and Lv.3, the income targets and the unit requirements pull resources in different directions.
The former demand investment in economic structures, whereas the latter require production buildings and, at Lv.3, a further tech step before any unit can be produced.
Because BAR's production dependencies make each stage contingent on earlier ones, inefficient early decisions delay the entire chain and consume the limited time budget.

\noindent \textbf{Game description.}
At the beginning of each episode, the agent receives a natural-language description of the targets and the time constraint, \eg, at Lv.3:

\begin{quote}
\textit{Reach a metal income of 20 and an energy income of 200, and field at least one Welder, as quickly as possible within 10 minutes.}
\end{quote}

\subsection{Multi-Front Defense (MFD) --- Spatial \& Temporal Reasoning}

\noindent \textbf{Strategic demand.}
BAR's large maps make travel time a first-class constraint, so the agent must decide not only where forces should be but also when they must start moving to arrive in time.
Forces committed to one location are unavailable elsewhere until they redeploy, which couples spatial and temporal decisions.

\noindent \textbf{Objective.}
The agent must defend a set of Energy Storage buildings located on a mountain for 30 seconds, during which enemy forces attack from different directions at staggered times.
Losing any of the buildings results in defeat, so a run counts as a win only if all of them survive.
The only available action category is \textbf{Action}, as building construction and unit production are disabled, and FoW is \textbf{disabled}.

\noindent \textbf{Difficulty levels.}
Lv.1 and Lv.2 place five storage buildings, and Lv.2 increases the total power of the attacking forces by 50\% over Lv.1.
Lv.3 adds a sixth building and increases the attacking power by 100\% over Lv.1, so that a larger enemy force is spread over one more front.

\noindent \textbf{Design rationale.}
With a fixed force and no production, success depends entirely on where the forces are placed and when they are moved.
The staggered attacks require the agent to reason jointly about spatial distance, travel time, and the temporal order of incoming threats.
Forces committed to one building may have to be relocated before a later attack reaches another, while moving them too early leaves the current front exposed.
Disabling FoW keeps the task focused on positioning and redeployment rather than on gathering information.

\noindent \textbf{Game description.}
At the beginning of each episode, the agent receives a natural-language description of the objectives to defend and the defeat condition, \eg, at Lv.1:

\begin{quote}
\textit{Defend the five Energy Storage buildings on the mountain for 30 seconds. Losing any of them results in defeat.}
\end{quote}

\subsection{Fortified Base Assault (FBA) --- Opponent Modeling}

\noindent \textbf{Strategic demand.}
Under partial observability, the agent must infer the opponent's forces, defenses, and likely response from what it has observed, and adapt its own composition and approach accordingly.
BAR's counter relationships among unit types make this inference consequential, as a force effective against one composition can be ineffective against another.

\noindent \textbf{Objective.}
The agent must destroy the enemy Advanced Metal Storage, located inside a fortified base, within a 10-minute limit.
A run counts as a win if the target is destroyed before the time limit.
The available action categories are \textbf{Build}, \textbf{Prod.}, and \textbf{Action}, and FoW is \textbf{enabled}, so the defensive layout is not immediately visible.

\noindent \textbf{Difficulty levels.}
Lv.1 uses the base configuration, while Lv.2 and Lv.3 increase the total power of the base by 100\% and 200\%, respectively, so that the agent must assemble a stronger force and choose its approach more carefully.

\noindent \textbf{Design rationale.}
The target is protected by defensive structures and combat forces whose composition and placement determine which attack strategies succeed.
Since the layout is revealed only as the agent scouts, it must infer the defensive configuration from partial observations and adapt its production, force composition, and approach accordingly.
For example, concentrated anti-air defenses discourage an air assault, while a heavily fortified approach favors an alternative route.
As further parts of the base become visible, the agent may need to revise its assessment and plan.

\noindent \textbf{Game description.}
At the beginning of each episode, the agent receives a natural-language description of the target and the time constraint:

\begin{quote}
\textit{Destroy the enemy Advanced Metal Storage within 10 minutes.}
\end{quote}

\subsection{Allied Team Skirmish (ATS) --- Collaboration}

\noindent \textbf{Strategic demand.}
Team play requires acting jointly with allies the agent does not control, which entails dividing fronts, producing forces that complement allied ones, and timing engagements together.
Since outcomes are shared, decisions that benefit the agent alone do not necessarily benefit the team.

\noindent \textbf{Objective.}
The agent fights alongside an ally in a 2v2 team battle for 30 seconds.
The outcome is decided at the team level from the damage each team deals and the army strength it maintains over the episode.
Let $D$ denote the total damage, in hit points, that a team deals, and let $\text{AS}$ denote its army strength, \ie, the area under the curve of the total power of its surviving combat units over the episode, with power defined as in Sec.~\ref{supp:metrics}.
Unlike the auxiliary AS metric, which is normalized by each player's alive span (Sec.~\ref{supp:metrics}), AS here is accumulated over the whole episode, so that units lost during the battle lower the team's share.
With subscripts \textit{own} and \textit{opp} denoting the agent's team (the agent and its ally) and the opposing team, the two shares are
\begin{equation}
    s_{\text{dmg}} = \frac{D_{\text{own}}}{D_{\text{own}}+D_{\text{opp}}},
    \qquad
    s_{\text{AS}} = \frac{\text{AS}_{\text{own}}}{\text{AS}_{\text{own}}+\text{AS}_{\text{opp}}},
\end{equation}
and the team with the higher weighted sum $0.5\,s_{\text{dmg}}+0.5\,s_{\text{AS}}$ wins.
Since the two shares of the opposing teams sum to one, a run counts as a win when the agent's team exceeds 0.5, \ie, when it inflicts more damage and maintains a stronger force than the opponents on average.
The available action categories are \textbf{Prod.} and \textbf{Action}, and FoW is \textbf{disabled}.

\noindent \textbf{Difficulty levels.}
The levels vary the production budget of the agent's team relative to the opposing team: $+10\%$ at Lv.1, equal budgets at Lv.2, and $-10\%$ at Lv.3.
Since both teams fight with comparable forces, the outcome depends less on raw force and more on how well the agent coordinates with its ally.

\noindent \textbf{Design rationale.}
The agent does not control its ally and must therefore account for the ally's behavior when deciding how to produce and move its own units.
Effective collaboration may involve reinforcing an allied engagement, producing units that complement the allied composition, or covering regions the ally leaves undefended, whereas poor coordination leads to fragmented attacks and redundant force allocation.
Since both shares are measured over the whole team, preserving the agent's own units at the expense of the ally does not improve the outcome.
Disabling FoW lets the agent observe allied and enemy activity, so that the task measures coordination rather than information gathering.

\noindent \textbf{Game description.}
At the beginning of each episode, the agent receives a natural-language description of the team structure, the cooperative objective, and how the outcome is evaluated:

\begin{quote}
\textit{Fight alongside your ally in a 2v2 team battle for 30 seconds. Your performance is evaluated from the damage your team deals and the forces it preserves.}
\end{quote}

\begin{table}[p]
\centering
\caption{\textbf{Mini-game evaluation results with 90\% Wilson score confidence intervals.}
Each cell reports the win rate over 30 runs; the upper limit of the interval is shown in red superscript and the lower limit in green subscript.
We report Wilson score intervals rather than the normal approximation, which is unreliable for small samples and degenerates at win rates of 0 and 1.
CoS is evaluated only on TCP, as its heuristic army-wide control lacks the spatial unit assignment required by the other mini-games.}
\vspace{0.5em}
\label{tab:minigame_results_ci}
\footnotesize
\begin{tabular}{lccccc}
\toprule
\textbf{Model} & \textbf{TCP} & \textbf{MFD} & \textbf{FBA} & \textbf{ATS} & \textbf{AWD} \\
\midrule

\multicolumn{6}{l}{\textit{Lv.1}} \\
\cmidrule(lr){1-6}

CoS~\citep{ma2024large}
& \ci{0.63}{0.48}{0.76}
& ---
& ---
& ---
& --- \\

LLM-PySC2~\citep{li2025llmpysc2}
& \ci{0.93}{0.82}{0.98}
& \ci{0.13}{0.06}{0.27}
& \ci{0.73}{0.59}{0.84}
& \ci{0.37}{0.24}{0.52}
& \ci{0.27}{0.16}{0.41} \\

SwarmBrain~\citep{shao2024swarmbrain}
& \ci{0.97}{0.86}{0.99}
& \ci{0.67}{0.52}{0.79}
& \ci{1.00}{0.92}{1.00}
& \ci{0.87}{0.73}{0.94}
& \ci{0.10}{0.04}{0.23} \\

\method{} (Ours)
& \ci{0.93}{0.82}{0.98}
& \ci{0.93}{0.82}{0.98}
& \ci{0.90}{0.77}{0.96}
& \ci{1.00}{0.92}{1.00}
& \ci{0.77}{0.62}{0.87} \\

\cmidrule(lr){1-6}

GPT-5.6-Luna
& \ci{0.47}{0.33}{0.61}
& \ci{0.50}{0.36}{0.64}
& \ci{0.83}{0.70}{0.92}
& \ci{0.50}{0.36}{0.64}
& \ci{0.00}{0.00}{0.08} \\

Qwen3.8-27B
& \ci{0.60}{0.45}{0.73}
& \ci{0.63}{0.48}{0.76}
& \ci{0.83}{0.70}{0.92}
& \ci{0.20}{0.11}{0.34}
& \ci{0.20}{0.11}{0.34} \\

Qwen3.5-397B
& \ci{0.53}{0.39}{0.67}
& \ci{0.67}{0.52}{0.79}
& \ci{0.87}{0.73}{0.94}
& \ci{0.27}{0.16}{0.41}
& \ci{0.03}{0.01}{0.14} \\

Muse-Spark-1.2
& \ci{1.00}{0.92}{1.00}
& \ci{0.73}{0.59}{0.84}
& \ci{0.80}{0.66}{0.89}
& \ci{0.27}{0.16}{0.41}
& \ci{0.23}{0.13}{0.38} \\

GPT-5.6-Terra
& \ci{0.73}{0.59}{0.84}
& \ci{0.37}{0.24}{0.52}
& \ci{0.90}{0.77}{0.96}
& \ci{0.30}{0.18}{0.45}
& \ci{0.13}{0.06}{0.27} \\

GLM-5.3-Flash
& \ci{0.93}{0.82}{0.98}
& \ci{0.83}{0.70}{0.92}
& \ci{1.00}{0.92}{1.00}
& \ci{0.53}{0.39}{0.67}
& \ci{0.27}{0.16}{0.41} \\

Grok-4.6
& \ci{1.00}{0.92}{1.00}
& \ci{0.80}{0.66}{0.89}
& \ci{1.00}{0.92}{1.00}
& \ci{0.47}{0.33}{0.61}
& \ci{0.23}{0.13}{0.38} \\

Gemini-3.7-Flash
& \ci{0.93}{0.82}{0.98}
& \ci{0.93}{0.82}{0.98}
& \ci{0.90}{0.77}{0.96}
& \ci{1.00}{0.92}{1.00}
& \ci{0.77}{0.62}{0.87} \\

\midrule

\multicolumn{6}{l}{\textit{Lv.2}} \\
\cmidrule(lr){1-6}

CoS~\citep{ma2024large}
& \ci{0.10}{0.04}{0.23}
& ---
& ---
& ---
& --- \\

LLM-PySC2~\citep{li2025llmpysc2}
& \ci{0.43}{0.30}{0.58}
& \ci{0.10}{0.04}{0.23}
& \ci{0.10}{0.04}{0.23}
& \ci{0.30}{0.18}{0.45}
& \ci{0.00}{0.00}{0.08} \\

SwarmBrain~\citep{shao2024swarmbrain}
& \ci{0.40}{0.27}{0.55}
& \ci{0.00}{0.00}{0.08}
& \ci{0.33}{0.21}{0.48}
& \ci{0.43}{0.30}{0.58}
& \ci{0.00}{0.00}{0.08} \\

\method{} (Ours)
& \ci{0.67}{0.52}{0.79}
& \ci{0.60}{0.45}{0.73}
& \ci{0.47}{0.33}{0.61}
& \ci{0.40}{0.27}{0.55}
& \ci{0.43}{0.30}{0.58} \\

\cmidrule(lr){1-6}

GPT-5.6-Luna
& \ci{0.07}{0.02}{0.18}
& \ci{0.10}{0.04}{0.23}
& \ci{0.40}{0.27}{0.55}
& \ci{0.27}{0.16}{0.41}
& \ci{0.00}{0.00}{0.08} \\

Qwen3.8-27B
& \ci{0.00}{0.00}{0.08}
& \ci{0.13}{0.06}{0.27}
& \ci{0.03}{0.01}{0.14}
& \ci{0.17}{0.08}{0.30}
& \ci{0.03}{0.01}{0.14} \\

Qwen3.5-397B
& \ci{0.17}{0.08}{0.30}
& \ci{0.37}{0.24}{0.52}
& \ci{0.20}{0.11}{0.34}
& \ci{0.13}{0.06}{0.27}
& \ci{0.00}{0.00}{0.08} \\

Muse-Spark-1.2
& \ci{0.13}{0.06}{0.27}
& \ci{0.60}{0.45}{0.73}
& \ci{0.20}{0.11}{0.34}
& \ci{0.17}{0.08}{0.30}
& \ci{0.10}{0.04}{0.23} \\

GPT-5.6-Terra
& \ci{0.20}{0.11}{0.34}
& \ci{0.23}{0.13}{0.38}
& \ci{0.37}{0.24}{0.52}
& \ci{0.20}{0.11}{0.34}
& \ci{0.00}{0.00}{0.08} \\

GLM-5.3-Flash
& \ci{0.77}{0.62}{0.87}
& \ci{0.47}{0.33}{0.61}
& \ci{0.37}{0.24}{0.52}
& \ci{0.30}{0.18}{0.45}
& \ci{0.00}{0.00}{0.08} \\

Grok-4.6
& \ci{0.83}{0.70}{0.92}
& \ci{0.13}{0.06}{0.27}
& \ci{0.57}{0.42}{0.70}
& \ci{0.23}{0.13}{0.38}
& \ci{0.10}{0.04}{0.23} \\

Gemini-3.7-Flash
& \ci{0.67}{0.52}{0.79}
& \ci{0.60}{0.45}{0.73}
& \ci{0.47}{0.33}{0.61}
& \ci{0.40}{0.27}{0.55}
& \ci{0.43}{0.30}{0.58} \\

\midrule

\multicolumn{6}{l}{\textit{Lv.3}} \\
\cmidrule(lr){1-6}

CoS~\citep{ma2024large}
& \ci{0.00}{0.00}{0.08}
& ---
& ---
& ---
& --- \\

LLM-PySC2~\citep{li2025llmpysc2}
& \ci{0.03}{0.01}{0.14}
& \ci{0.00}{0.00}{0.08}
& \ci{0.00}{0.00}{0.08}
& \ci{0.10}{0.04}{0.23}
& \ci{0.00}{0.00}{0.08} \\

SwarmBrain~\citep{shao2024swarmbrain}
& \ci{0.07}{0.02}{0.18}
& \ci{0.00}{0.00}{0.08}
& \ci{0.30}{0.18}{0.45}
& \ci{0.13}{0.06}{0.27}
& \ci{0.00}{0.00}{0.08} \\

\method{} (Ours)
& \ci{0.20}{0.11}{0.34}
& \ci{0.37}{0.24}{0.52}
& \ci{0.20}{0.11}{0.34}
& \ci{0.23}{0.13}{0.38}
& \ci{0.13}{0.06}{0.27} \\

\cmidrule(lr){1-6}

GPT-5.6-Luna
& \ci{0.00}{0.00}{0.08}
& \ci{0.03}{0.01}{0.14}
& \ci{0.00}{0.00}{0.08}
& \ci{0.00}{0.00}{0.08}
& \ci{0.00}{0.00}{0.08} \\

Qwen3.8-27B
& \ci{0.00}{0.00}{0.08}
& \ci{0.00}{0.00}{0.08}
& \ci{0.10}{0.04}{0.23}
& \ci{0.00}{0.00}{0.08}
& \ci{0.00}{0.00}{0.08} \\

Qwen3.5-397B
& \ci{0.00}{0.00}{0.08}
& \ci{0.13}{0.06}{0.27}
& \ci{0.10}{0.04}{0.23}
& \ci{0.07}{0.02}{0.18}
& \ci{0.00}{0.00}{0.08} \\

Muse-Spark-1.2
& \ci{0.07}{0.02}{0.18}
& \ci{0.00}{0.00}{0.08}
& \ci{0.00}{0.00}{0.08}
& \ci{0.10}{0.04}{0.23}
& \ci{0.00}{0.00}{0.08} \\

GPT-5.6-Terra
& \ci{0.00}{0.00}{0.08}
& \ci{0.17}{0.08}{0.30}
& \ci{0.33}{0.21}{0.48}
& \ci{0.07}{0.02}{0.18}
& \ci{0.00}{0.00}{0.08} \\

GLM-5.3-Flash
& \ci{0.07}{0.02}{0.18}
& \ci{0.00}{0.00}{0.08}
& \ci{0.40}{0.27}{0.55}
& \ci{0.00}{0.00}{0.08}
& \ci{0.00}{0.00}{0.08} \\

Grok-4.6
& \ci{0.23}{0.13}{0.38}
& \ci{0.00}{0.00}{0.08}
& \ci{0.30}{0.18}{0.45}
& \ci{0.07}{0.02}{0.18}
& \ci{0.00}{0.00}{0.08} \\

Gemini-3.7-Flash
& \ci{0.20}{0.11}{0.34}
& \ci{0.37}{0.24}{0.52}
& \ci{0.20}{0.11}{0.34}
& \ci{0.23}{0.13}{0.38}
& \ci{0.13}{0.06}{0.27} \\

\bottomrule
\end{tabular}
\end{table}

\subsection{Anticipatory Wave Defense (AWD) --- Adversarial Planning}

\noindent \textbf{Strategic demand.}
Plans must anticipate an opponent that acts concurrently, so preparations have to be made before threats materialize and are often costly in the short term.
In BAR, counters to a specific threat, such as anti-air units or defensive structures, must be built in advance to be available when needed.

\noindent \textbf{Objective.}
The agent must withstand successive enemy waves converging on its commander from different directions for 30 seconds.
A run counts as a win if the commander survives all waves.
The available action categories are \textbf{Build}, \textbf{Prod.}, and \textbf{Action}, and FoW is \textbf{disabled}.

\noindent \textbf{Difficulty levels.}
The levels increase both the number of attack directions and the strength of the waves.
Lv.1 sends waves from two directions, Lv.2 from three with the total enemy power increased by 50\% over Lv.1, and Lv.3 from four with an increase of 100\%.
More directions leave fewer forces available per front, so the agent must spread its defenses rather than concentrate them.

\noindent \textbf{Design rationale.}
The composition of each wave is given to the agent in advance, so the central question is whether it uses this information proactively rather than reacting to each wave as it arrives.
Different compositions call for different unit types, defensive structures, and force placements, and preparation that has no immediate value, such as anti-air units before an air-heavy wave, pays off only if the agent plans ahead.

\noindent \textbf{Game description.}
At the beginning of each episode, the agent receives a natural-language description of the survival objective and the compositions of the upcoming waves:

\begin{quote}
\textit{Withstand the enemy waves converging on your commander from different directions for 30 seconds. The composition of each wave is known in advance.}
\end{quote}



\subsection{Results with Confidence Intervals}
\label{supp:minigame_ci}

Since each mini-game is evaluated over 30 runs per difficulty level, the reported win rates carry substantial sampling uncertainty.
Tab.~\ref{tab:minigame_results_ci} therefore repeats the mini-game results of the main paper together with 90\% Wilson score confidence intervals.
The intervals are wide enough that many of the differences between models at a single level are not statistically distinguishable, and we accordingly base our claims on consistent trends across levels and mini-games rather than on individual cells.


\section{Detailed Procedure of~\method}
\label{supp:agent_inference}

This section details the action space (Sec.~\ref{supp:agent_action_space}), trajectory planning and replanning (Sec.~\ref{supp:agent_trajectory}), and inference loop (Sec.~\ref{supp:agent_algo}) of~\method{} introduced in Sec.~\ref{sec:rtsagent}.

\subsection{Action Space}
\label{supp:agent_action_space}

\method{} keeps the building construction and unit production actions of the benchmark interface (Eq.~\ref{eq:rtsbench_policy}) and replaces per-unit \emph{unit action} for combat forces with two group-level action types, \emph{group assignment} and \emph{group command}, so that the action space at interval~$t$ is
\begin{equation}
\mathcal{A}_t
=
\mathcal{A}^{\text{build}}_t
\cup
\mathcal{A}^{\text{produce}}_t
\cup
\mathcal{A}^{\text{assign}}_t
\cup
\mathcal{A}^{\text{command}}_t.
\end{equation}

\paragraph{Building construction and unit production.}
Let $\mathcal{U}^{\text{ctor}}_t$ denote the available constructor units, $\mathcal{F}_t$ the active production buildings, and $\mathcal{O}(e)$ the build options of entity~$e$ in the encyclopedia of~$\mathcal{K}$, \ie, the building types a constructor can build or the unit types a production building can produce.
With $\mathcal{X}\times\mathcal{Y}$ denoting the coordinate grid spanning $(0,0)$ to $(100,100)$,
\begin{align}
\mathcal{A}^{\text{build}}_t
&=
\bigl\{(u,b,x,y) \bigm| u\in\mathcal{U}^{\text{ctor}}_t,\ b\in\mathcal{O}(u),\ (x,y)\in\mathcal{X}\times\mathcal{Y}\bigr\},\\
\mathcal{A}^{\text{produce}}_t
&=
\bigl\{(f,p) \bigm| f\in\mathcal{F}_t,\ p\in\mathcal{O}(f)\bigr\}.
\end{align}
Both action types are inherited unchanged from the base interface, and their low-level execution is handled by the game engine.

\paragraph{Group assignment.}
Let $\mathcal{U}^{\text{combat}}_t$ denote the active combat units and $\mathcal{G}_t$ the group identifiers available at interval~$t$, including newly created ones.
A group-assignment action associates a subset of combat units with a group:
\begin{equation}
\mathcal{A}^{\text{assign}}_t
=
\mathcal{G}_t
\times
2^{\mathcal{U}^{\text{combat}}_t},
\end{equation}
where each unit belongs to at most one group, so that assigning a unit removes it from its previous group.
Unlike predefined combat groups~\citep{li2025llmpysc2}, groups are created, updated, or dissolved by the VLM according to the current strategy and game state.
The textual observation~$\mathcal{W}(s_t)$ reports the current group membership, unit counts, and unassigned combat units, so that subsequent assignments can build on the existing grouping.

\paragraph{Group command.}
For each group $g\in\mathcal{G}_t$, a group command specifies a mode from $\mathcal{C}=\{\texttt{move},\texttt{move\_force},\texttt{stop}\}$ and, when needed, a target coordinate:
\begin{equation}
\mathcal{A}^{\text{command}}_t = \mathcal{G}_t\times\mathcal{C}\times\bigl((\mathcal{X}\times\mathcal{Y})\cup\{\varnothing\}\bigr),
\end{equation}
where $\varnothing$ denotes no target, as for \texttt{stop}.
Under \texttt{move}, a group travels toward its target, and under \texttt{stop}, it holds its current location.
In both modes, engagement is handled by the execution controller, which switches a group to a \texttt{fight} state upon enemy contact and resumes its command once the engagement ends.
Under \texttt{move\_force}, a group instead pushes through toward its target without engaging, which is useful for repositioning across contested terrain.
Since \texttt{fight} is never issued by the VLM, it is excluded from~$\mathcal{C}$.
Consequently, the number of commands the VLM issues scales with the number of groups~$|\mathcal{G}_t|$ rather than the number of combat units, while the VLM retains control over group membership, command modes, targets, and conditional trajectories.

\subsection{Trajectory Planning and Replanning}
\label{supp:agent_trajectory}

\paragraph{Command trajectories.}
Since the VLM is invoked only once per interval, a single command would leave a group unable to react until the next call.
Instead, the VLM outputs for each group~$g$ a condition-dependent command trajectory, represented as a directed acyclic graph (DAG) $\mathcal{T}_t^g=(\mathcal{V}_t^g,\mathcal{E}_t^g)$ whose nodes hold group commands and whose edges are labeled with transition conditions.
Execution starts at the root node.
Whenever the condition on an outgoing edge of the active node is satisfied, the controller activates the successor node without another VLM call; otherwise, the active command remains in effect.

\paragraph{Transition conditions.}
Rather than free-form text, each condition is one of four predefined primitives whose parameter is set by the VLM, allowing the controller to evaluate conditions unambiguously:
\begin{itemize}[leftmargin=1.5em]
\item \texttt{idle\_for}: all units in the group have no remaining orders, typically after completing the current command.
\item \texttt{hp\_below}: the average health of units in the group falls below a threshold.
\item \texttt{group\_size\_below}: the number of units in the group falls below a threshold.
\item \texttt{timeout}: a given amount of time has elapsed since the current node became active.
\end{itemize}
For example, a group can advance to a rally point and, once \texttt{idle\_for} holds, proceed to an enemy base, while an \texttt{hp\_below} edge redirects it to the home base if it sustains heavy damage.

\paragraph{Replanning triggers.}
Trajectories cover changes that can be anticipated at decision time.
For sudden changes that require a new strategic decision, three triggers with fixed thresholds are monitored over the entire game state:
\begin{itemize}[leftmargin=1.5em]
\item \texttt{enemy\_surge}: the number of visible enemy units increases by at least 50\% and at least five units since the most recent inference.
\item \texttt{army\_loss}: the number of allied combat units decreases by at least 30\% and at least five units since the most recent inference.
\item \texttt{commander\_hp}: the commander's health falls below 50\% of its maximum.
\end{itemize}
Combining relative and absolute thresholds keeps the first two triggers insensitive to small fluctuations regardless of army size.
When any trigger fires, the game is paused and a new inference cycle begins immediately, replacing the current group assignments and trajectories.

\subsection{Inference Loop}
\label{supp:agent_algo}

\begin{algorithm}[t]
\caption{Inference loop of \method.}
\label{alg:rtsagent}
\begin{algorithmic}[1]
\Require static game knowledge $\mathcal{K}$, VLM policy $\pi$, call interval $\Delta$
\Statex \textcolor{gray}{\textit{Notation:}} $\mathcal{T}^g_t$ is the trajectory of group $g$ at call $t$, a DAG whose nodes are commands $c \in \{\texttt{move}, \texttt{move\_force}, \texttt{stop}\}$ and whose edges carry transition conditions.
\State $\mathcal{L}_0 \gets \varnothing$;\ $\mathcal{S}_1 \gets \varnothing$
\For{$t = 1, 2, \dots$ until the game ends}
    \State Pause the game and read the engine state $s_t$
    \State $o_t \gets \bigl(v_t, \mathcal{W}(s_t)\bigr)$ \Comment{Observation, restricted to allied line of sight under FoW}
    \State $(m_t, \mathcal{L}_t) \gets \mathrm{VLM}(\mathcal{S}_t, \mathcal{L}_{t-1})$ \Comment{Memory phase, text-only}
    \State $\mathcal{S}_{t+1} \gets \varnothing$
    \State $a_t \gets \pi(o_t, m_t \mid \mathcal{K})$ \Comment{Decision phase}
    \State Dispatch building and production orders to the game engine
    \State Update groups and trajectories $\{\mathcal{T}^g_t\}$; activate each root node
    \State Resume the game
    \Repeat \Comment{Trajectory execution without further VLM calls}
        \State Advance the game; append triggered event logs to $\mathcal{S}_{t+1}$
        \For{each group $g$}
            \If{a condition on an outgoing edge of the active node of $\mathcal{T}^g_t$ holds}
                \State Activate the corresponding successor node
            \EndIf
        \EndFor
    \Until{$\Delta$ has elapsed \textbf{or} a critical trigger fires} \Comment{The latter initiates replanning}
\EndFor
\end{algorithmic}
\end{algorithm}

Algorithm~\ref{alg:rtsagent} summarizes the inference loop.
Following the benchmark interface, the game advances for a fixed interval~$\Delta$ between scheduled inferences and is paused during inference.
Before the game begins, the agent receives the static game knowledge~$\mathcal{K}$, consisting of the scenario description, team configuration, and unit and building encyclopedia, which remains fixed throughout the episode.
The long-term memory, short-term memory, and set of groups all start empty.

\paragraph{Observation.}
At each interval~$t$, the agent constructs $o_t=\bigl(v_t,\mathcal{W}(s_t)\bigr)$ from the engine state~$s_t$.
The visual observation~$v_t$ consists of a global minimap and up to four local camera views, selected from candidate locations in the following fixed order of priority:
(i)~recent combat locations,
(ii)~the commander's location when its health is below 50\%, the same threshold as the \texttt{commander\_hp} trigger,
(iii)~visible enemy force clusters,
(iv)~allied group clusters,
(v)~the commander's location when its health is 50\% or above, and
(vi)~the allied home base.
Spatially close candidates are merged, and up to four of the highest-priority locations are selected.
With FoW enabled, both the minimap and the local views are restricted to allied line of sight.
The textual observation~$\mathcal{W}(s_t)$, extracted by the Python wrapper, summarizes resources, buildings and production states, allied units and groups, visible enemies, and pending actions.

\paragraph{Memory phase.}
Between VLM calls, event-triggered logs accumulate in the short-term memory~$\mathcal{S}_t$.
A log entry is recorded when
(i)~enemy units are observed,
(ii)~allied units, including the commander, take damage or are destroyed,
(iii)~a new unit is completed,
(iv)~a unit receives a new order, such as moving to a target coordinate,
(v)~a resource reaches its storage cap, indicating a surplus, or
(vi)~a resource is depleted, indicating a shortage.
At each inference cycle, the policy VLM, prompted with text only, consolidates $\mathcal{S}_t$ with the previous long-term memory~$\mathcal{L}_{t-1}$, producing the updated memory~$\mathcal{L}_t$ and the decision-relevant context~$m_t$:
\begin{equation}
    (m_t,\mathcal{L}_t) = \mathrm{VLM}\bigl(\mathcal{S}_t, \mathcal{L}_{t-1}\bigr).
\end{equation}
$\mathcal{S}_t$ is then cleared, so that transient events do not accumulate indefinitely, while the persistent summaries in~$\mathcal{L}_t$ retain strategically relevant information throughout the game.

\paragraph{Decision phase.}
The VLM policy then produces the action plan $a_t=\pi(o_t,m_t\mid\mathcal{K})$, comprising building and production orders, group assignments, and a command trajectory for each group (Sec.~\ref{supp:agent_action_space} and Sec.~\ref{supp:agent_trajectory}).
Building and production orders are executed through the base interface, while group assignments and trajectories are passed to the execution controller, which carries them out without further model calls until~$\Delta$ elapses or a critical trigger fires.

\subsection{Adapting StarCraft~II LLM Agents to BAR}
\label{supp:adaptation_agents}

CoS~\citep{ma2024large}, SwarmBrain~\citep{shao2024swarmbrain}, and LLM-PySC2~\citep{li2025llmpysc2} were designed for StarCraft~II (SC2) and depend on its interfaces, race-specific mechanics, and worker-harvesting economy, so running them on BAR requires re-implementation. This section describes each re-implementation component by component, showing that it preserves the mechanisms its original paper identifies as its contribution and stating every deviation from the original explicitly.

\paragraph{Shared protocol.}
All agents, including ours, share the observation interface, action executor, unit-level control rules, backbone LLM, decoding configuration, and decision interval of the main experiments.
On this shared substrate, each baseline follows its original paper in prompt structure, LLM roles, memory, grouping, and reflex rules, except for the deviations noted below (Table~\ref{tab:baseline_adaptation}).
No baseline uses the memory or replanning modules of our agent, and race-specific prompt content is rewritten for BAR while the prompt structure is kept.

\begin{table}[t]
\centering
\caption{Structural summary of the adapted baselines. SwarmBrain calls the Overmind and the SC2 Brain; LLM-PySC2 calls the Commander, the Developer, and one CombatGroup per squad, where $N$ is the number of combat squads.}
\label{tab:baseline_adaptation}
\resizebox{\linewidth}{!}{%
\begin{tabular}{lccc}
\toprule
 & CoS~\citep{ma2024large} & SwarmBrain~\citep{shao2024swarmbrain} & LLM-PySC2~\citep{li2025llmpysc2} \\
\midrule
LLM calls per step & 1 & 2 & $N{+}2$ \\
Memory & last 4 frames & last 5 directives & none \\
Unit grouping & none & collective, label-addressed & per unit archetype \\
Unit-level control & shared rules & ReflexNet + shared rules & shared rules \\
Army engagement & scripted push & SC2 Brain & CombatGroups \\
\bottomrule
\end{tabular}%
}
\end{table}

\noindent \textbf{CoS}~\citep{ma2024large}\textbf{.}
In TextStarCraft~II, the LLM selects only macro actions while scripts handle all unit control.
Each observation is summarized, $K$ summaries are aggregated, and the LLM reasons through six phases, from situation overview to decision-making, before extracting actions into a queue.
Our port keeps these elements.
The structured text observation serves as the single-frame summary, the LLM receives the current frame with the preceding $K{=}4$ frames, and the prompt enforces the six phases in order.
Its action space is restricted to position-free construction and production, as research has no BAR counterpart, where technology advances through higher-tier factories.
The scripted engagement becomes a fixed-timing push: at 20 in-game minutes, all combat units attack a target sampled from the enemy Commanders' spawn positions, re-issued every 5 minutes so that new units join.
\emph{Deviation:} Instead of querying every $K$ steps and executing a queue of $K$ actions, we query at every decision step over a sliding window of $K$ frames, keeping CoS on the shared decision interval.

\noindent \textbf{SwarmBrain}~\citep{shao2024swarmbrain}\textbf{.}
SwarmBrain pairs an LLM-based Overmind Intelligence Matrix with the Swarm ReflexNet, rule-based state machines that act without waiting for the LLM.
Within the Matrix, the Overmind Brain formulates a strategy, a text memory keeps it consistent, the SC2 Brain translates it into commands, and a Command Center validates them, with locations referenced by mineral-field labels.
Our port keeps this structure.
Each cycle runs the Overmind and then the SC2 Brain on its directive, giving two calls per step, and the five most recent directives serve as the strategy memory.
Metal spots, the BAR counterpart of mineral fields, are labeled at every frame and referenced by both roles, and our JSON validator replaces the Command Center.
Since BAR has neither worker harvesting nor supply units, we port only the ReflexNet's flee transition, retreating a damaged combat unit near an enemy.
\emph{Deviation:} The original SC2 Brain is text-only and forms collectives by unit type and location; ours receives the shared visual observation and commands the groups of our interface.

\noindent \textbf{LLM-PySC2}~\citep{li2025llmpysc2}\textbf{.}
LLM-PySC2 defines a Commander for combat deployment, a Developer for economy, CombatGroups for micro-operations, and a Builder for building placement, all queried in independent threads.
The Commander directs the CombatGroups through messages delivered at the next step.
Our port keeps these elements.
The Commander only sends messages, the Developer handles construction and production, and one CombatGroup per unit-archetype group moves its own group, giving $N{+}2$ concurrent calls per step.
Folding the Builder into the Developer corresponds to Easy Build mode, so our setting matches the paper's Standard Control + Easy Build configuration.
Each role observes what it needs, namely group centroids and alerts for the Commander, base state and army totals for the Developer, and nearby units with a screen crop for each CombatGroup, and all roles receive the minimap.
\emph{Deviation:} The original CombatGroups act over the full PySC2 action space, including unit skills; ours issue group-level movement, and unit-level engagement follows the shared rules.

\begin{figure}[t]
    \centering
    \includegraphics[width=0.7\linewidth]{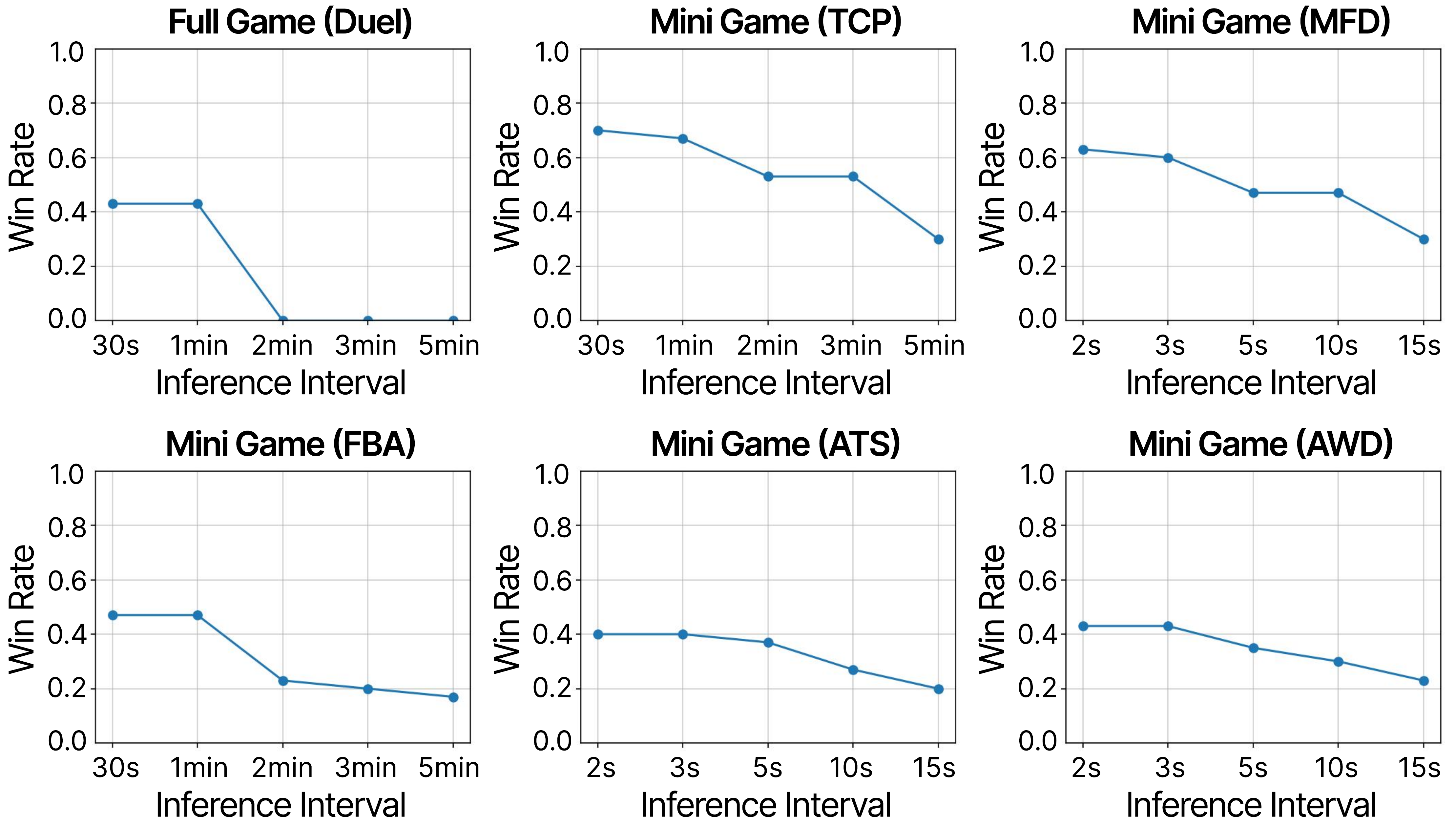}
    \caption{\textbf{Sensitivity to the inference interval.} Win rate of \method with Gemini-3.7-Flash in six Lv.2 scenarios under different inference intervals, \ie, the game time between VLM calls (30 games per point). Shorter intervals allow more frequent decisions but require more VLM calls; win rate saturates at the two shortest intervals and declines beyond them. We use 1\,min for full games, TCP, and FBA, and 3\,s for MFD, ATS, and AWD.}
    \label{fig:interval_sensitivity}
\end{figure}

\subsection{Inference Interval Analysis}
\label{supp:interval}

The inference interval, \ie, the game time between consecutive VLM calls, trades off decision frequency against inference cost.
Since the game is paused during inference, a shorter interval does not penalize slow responses but lets the agent revise its decisions more often, at the cost of more VLM calls.
We therefore set the interval according to the temporal demands of each setting, sharing one interval within each group below.

\noindent \textbf{Full games (1\,min).}
Full games span long horizons dominated by economic expansion and strategic planning.
Sudden events within an interval are handled by command trajectories and replanning triggers (Sec.~\ref{supp:agent_trajectory}), so frequent VLM calls are unnecessary.

\noindent \textbf{Planning-oriented mini-games (1\,min).}
TCP requires meeting production and economy targets, and FBA requires destroying a fortified enemy base.
Both depend on long-term planning, such as sequencing builds or preparing an assault, rather than on moment-to-moment reactions.

\noindent \textbf{Defense-oriented mini-games (3\,s).}
MFD, ATS, and AWD require repelling incoming attacks, during which the situation changes rapidly and forces must be redeployed frequently.

\noindent \textbf{Sensitivity results.}
In Fig.~\ref{fig:interval_sensitivity}, win rate in every setting saturates at the two shortest intervals and declines as the interval grows further, most sharply in full games.
The chosen intervals all lie within this saturated range and thus retain near-best performance, whereas longer intervals degrade it.


\section{Details about Evaluation Metrics for \benchmark}
\label{supp:metrics}
\label{supp:metrics_agent} 

We use win rate (WR) as the primary metric for agent evaluation, together with six auxiliary metrics for a detailed analysis of agent behavior in full games.

\noindent \textbf{Win rate (WR).}
WR is the primary metric for all full-game match-ups and mini-games, defined as the fraction of evaluation runs in which the agent satisfies the win condition:
\begin{equation}
    \text{WR} = \frac{\#\,\text{wins}}{\#\,\text{runs}}.
\end{equation}
In team match-ups, a run counts as a win when the agent's team wins.
The win condition of each mini-game is described in Sec.~\ref{supp:minigames}.

\noindent \textbf{Motivation for auxiliary metrics.}
WR reflects only the final outcome and barely distinguishes agents when it is close to zero or one.
To analyze \emph{how} agents play in full games, we measure four dimensions of gameplay: \emph{resources} (resources gathered, RG; resources spent, RS), \emph{army} (army strength, AS; unit quality, UQ), \emph{battle} (enemies destroyed, ED), and \emph{visibility} (vision coverage, VC).
Raw in-game statistics, however, are hard to compare across runs.
Their scale depends on the match-up, and since they accumulate over time, they conflate performance with how long a player stays in the game, so that players eliminated early would appear weak in every dimension and inflate the others' shares.
We therefore express each quantity as a rate over each player's alive span and report the agent's normalized share among all players, leaving survival itself to WR.
Per-unit averages, which are already independent of match length, are used without this conversion.

\noindent \textbf{Share normalization.}
Let $T_i=[t_0,t_i^{\text{end}}]$ with $t_i^{\text{end}}=\min(t_{\text{final}},t_i^{\text{death}})$ denote the alive span of player~$i$, where $t_0$ and $t_{\text{final}}$ are the start and end of the match, and let $x_i^{\mathrm{M}}$ denote its per-player quantity for metric~$\mathrm{M}$, defined below.
The agent's share among all $N$ players, counting each ally and opponent separately, is
\begin{equation}
    p^{\mathrm{M}} = \frac{x_1^{\mathrm{M}}}{\sum_{i=1}^{N} x_i^{\mathrm{M}}},
\end{equation}
with player~1 denoting the agent.
Since the even share $1/N$ varies across match-ups, we report the odds-normalized share
\begin{equation}
    \mathrm{M} = f\bigl(p^{\mathrm{M}};N\bigr),
    \qquad
    f(p;N)=\frac{\rho}{1+\rho},
    \quad
    \rho=\frac{(N-1)\,p}{1-p},
\end{equation}
which maps $p=0$, $1/N$, and $1$ to $0$, $0.5$, and $1$, so that 0.5 denotes an even share in every match-up.
Equivalently, $\mathrm{M}=x_1^{\mathrm{M}}/(x_1^{\mathrm{M}}+\bar{x}_{-1}^{\mathrm{M}})$, where $\bar{x}_{-1}^{\mathrm{M}}$ is the mean over the other $N-1$ players, so each metric compares the agent with an average other player.
For $N=2$, $f$ reduces to the identity, \ie, a share between two sides is already anchored at 0.5 and is reported as is.

\noindent \textbf{Per-player quantities.}
All quantities are expressed in \emph{power}, BAR's internal value scale, which converts energy into metal-equivalent units at a ratio of 60 to 1.
The power of an entity~$u$ is thus $\beta(u)=c_{\text{metal}}(u)+c_{\text{energy}}(u)/60$, where $c_{\text{metal}}$ and $c_{\text{energy}}$ denote its metal and energy costs, and the same conversion is applied to resource income and spending.
For player~$i$, let $\phi_i(t)$ denote its power income rate, $S_i(t)$ its cumulative power spent, $\mathcal{U}^{\text{combat}}_i(t)$ its surviving combat units at time~$t$, $\mathcal{P}_i$ the combat units it produced within~$T_i$, and $\mathcal{D}_i$ the enemy units and buildings it destroyed within~$T_i$ according to kill attribution.
Quantities that accumulate over time are divided by~$|T_i|$ to obtain rates over the alive span:
\begin{align}
    x_i^{\text{RG}} &= \frac{1}{|T_i|}\int_{T_i}\phi_i(t)\,dt,
    &
    x_i^{\text{RS}} &= \frac{S_i(t_i^{\text{end}})}{|T_i|},
    \\
    x_i^{\text{AS}} &= \frac{1}{|T_i|}\int_{T_i}\sum_{u\in\mathcal{U}^{\text{combat}}_i(t)}\beta(u)\,dt,
    &
    x_i^{\text{ED}} &= \frac{1}{|T_i|}\sum_{u\in\mathcal{D}_i}\beta(u),
    \\
    x_i^{\text{UQ}} &= \frac{1}{|\mathcal{P}_i|}\sum_{u\in\mathcal{P}_i}\beta(u).
\end{align}
AS measures the average power of the combat force a player maintains, whereas UQ measures the average power per produced unit and thus reflects how advanced its unit composition is.
We measure UQ with power rather than BAR's discrete tech tiers (T1--T3), because units within the same tier differ considerably in cost and capability.
Tiers register only transitions between tiers and miss shifts toward stronger units within a tier, whereas power places every unit on a continuous scale that still increases with tier and is shared with the other metrics.
AS and UQ count only combat units, as the army dimension concerns the force the agent can deploy, whereas investment in economic and defensive structures is reflected in RS.

\noindent \textbf{Vision coverage.}
Since allies share vision, VC is computed as a share among alliances rather than players.
Let $K$ denote the number of alliances in a match, \ie, $K=2$ in Duel and team match-ups and $K=4$ in Free-for-All, and let $\bar{\eta}_k$ denote the fraction of the map within the vision of alliance~$k$, averaged over the part of the agent's alive span~$T_1$ during which alliance~$k$ has surviving members, with $k=1$ denoting the agent's alliance.
Restricting each average to the period in which the alliance survives prevents alliances eliminated earlier in Free-for-All from inflating the agent's share, consistent with the alive-span normalization of the other metrics.
VC is defined as
\begin{equation}
    \text{VC} = f\Biggl(\frac{\bar{\eta}_1}{\sum_{k=1}^{K}\bar{\eta}_k};\,K\Biggr),
\end{equation}
which reduces to the alliance share itself when $K=2$.

\begin{algorithm}[!ht]
    \caption{Self-Evolving Game Generation Framework}
    \label{alg:self_evolving}
    \begin{algorithmic}[1]
        \Require User query $q$; shared knowledge database $\mathcal{D}$ of GDDs, rule sets, and stage-wise rubrics $\mathcal{R}$, all initially empty, and fixed reference scenarios $\mathcal{S}_{\text{ref}}$; rollback budget $B{=}3$
        \Ensure Mini-games $\mathcal{G}_{1:3}$ at Lv.1--3 with status \texttt{ok}, or, if aborted, the outputs produced so far with status \texttt{aborted}; updated $\mathcal{D}$ and $\mathcal{R}$

        \State $\mathcal{A} \leftarrow \emptyset$, $\mathcal{F} \leftarrow \emptyset$, $n \leftarrow 0$, $\text{fb}^\star \leftarrow \emptyset$ \Comment{Artifacts, feedback log, rollback count, guidance}

        \Statex \textcolor{gray}{\textit{// Stage 1: Scenario Brief Generation}}
        \State $b \leftarrow \Call{Designer.Clarify}{q}$ \Comment{Brief with map, composition, and demand ratings}
        \State $k \leftarrow 2$ \Comment{Current stage}
        \If{$\Call{PM.Match}{b, \mathcal{D}}$ finds a reusable GDD} \Comment{Compare $b$ with all stored GDDs}
            \State Retrieve $\{\text{GDD}, \mathcal{C}, \text{Rules}\}$ from $\mathcal{D}$ into $\mathcal{A}$; $k \leftarrow 4$ \Comment{Fast-track past Stages 2--3}
        \EndIf

        \Statex \textcolor{gray}{\textit{// Stages 2--4 with a Global Rollback Budget}}
        \While{$k \le 4$}
            \If{$k = 2$} \Comment{GDD generation}
                \State $(\text{GDD}, \mathcal{C}) \leftarrow \Call{Designer.Expand}{b, \text{fb}^\star}$ \Comment{Finalize demand ratings}
                \State $\mathcal{R}_g \leftarrow \Call{PM.SelectRubrics}{\mathcal{R}^{(2)}, b}$
                \State $(\text{pass}, \text{fb}) \leftarrow \Call{Analyst.Validate}{\text{GDD}, \mathcal{R}_g}$
            \ElsIf{$k = 3$} \Comment{Rule set construction}
                \State $\text{Rules}_{\text{reuse}} \leftarrow \Call{PM.RetrieveRules}{\text{GDD}, \mathcal{D}}$ \Comment{Verified implementations, if any}
                \State $(\text{Rules}, \mathcal{T}_r) \leftarrow \Call{Developer.Implement}{\text{GDD}, \text{Rules}_{\text{reuse}}, \text{fb}^\star}$ \Comment{New rules only}
                \State $\mathcal{R}_r \leftarrow \Call{PM.SelectRubrics}{\mathcal{R}^{(3)}, \text{GDD}}$
                \State $(\text{pass}, \text{fb}) \leftarrow \Call{Analyst.Validate}{\text{Rules}, \mathcal{R}_r, \mathcal{T}_r}$ \Comment{Rule-based tests in engine}
            \Else \Comment{Game implementation and verification}
                \State $(\mathcal{G}, \mathcal{T}_s) \leftarrow \Call{Developer.Configure}{b, \mathcal{A}, \text{Assets}, \text{fb}^\star}$ \Comment{Medium difficulty, map in $b$}
                \State $\mathcal{R}_s \leftarrow \Call{PM.SelectRubrics}{\mathcal{R}^{(4)}, \mathcal{A}}$
                \State $(\text{pass}, \text{fb}) \leftarrow \Call{Analyst.Validate}{\mathcal{G}, \mathcal{R}_s \cup \mathcal{C}, \mathcal{T}_s}$ \Comment{Rubrics, checklist, and tests}
            \EndIf
            \State $\mathcal{F} \leftarrow \mathcal{F} \cup \{\text{fb}\}$
            \If{pass}
                \State Add the stage output to $\mathcal{A}$ if $k \le 3$; $k \leftarrow k + 1$; $\text{fb}^\star \leftarrow \emptyset$
            \ElsIf{$n < B$}
                \State $(k, \text{fb}^\star) \leftarrow \Call{PM.MetaFeedback}{\mathcal{F}}$ \Comment{Roll back to the failing or an earlier stage}
                \State $\mathcal{A} \leftarrow \mathcal{A}_{<k}$; $n \leftarrow n + 1$ \Comment{Discard artifacts of stages $\ge k$}
            \Else
                \State \Return (outputs produced so far, \texttt{aborted}) \Comment{Budget exhausted}
            \EndIf
        \EndWhile

        \Statex \textcolor{gray}{\textit{// Difficulty Scaling (No Further Validation)}}
        \State $\mathcal{G}_{1:3} \leftarrow \Call{Developer.Scale}{\mathcal{G}, \mathcal{S}_{\text{ref}}}$ \Comment{Lv.1--3 at once; $\mathcal{G}$ as medium}

        \Statex \textcolor{gray}{\textit{// Self-Evolution Phase}}
        \State $\mathcal{D} \leftarrow \mathcal{D} \cup \mathcal{A}$ \Comment{Commit GDD, checklist, and rule set}
        \State $\mathcal{R} \leftarrow \Call{PM.Retrospective}{\mathcal{R}, \mathcal{F}}$ \Comment{Update rubrics within each stage's hierarchy}
        \State \Return $(\mathcal{G}_{1:3}, \texttt{ok})$
    \end{algorithmic}
\end{algorithm}


\section{Details of the Self-Evolving Game Generation Framework}
\label{supp:self_evolving}

This section details the self-evolving game generation framework introduced in Sec.~\ref{sec:minigame_gen}; its full procedure (Sec.~\ref{supp:self_evolving_algo}), the evaluation of generated games (Sec.~\ref{supp:eval_generated_games}), an ablation of its self-evolution mechanisms (Sec.~\ref{supp:abl_self_evol_gen_framework}), representative generation examples (Sec.~\ref{supp:selfevolving_examples}), and agent evaluation on the generated games (Sec.~\ref{supp:exp_generated_games}).

\subsection{Algorithm Details}
\label{supp:self_evolving_algo}

Algorithm~\ref{alg:self_evolving} presents the full procedure of the framework, and we detail each component below.

\begin{figure}[t]
    \centering
    \resizebox{0.9\linewidth}{!}{
        \includegraphics{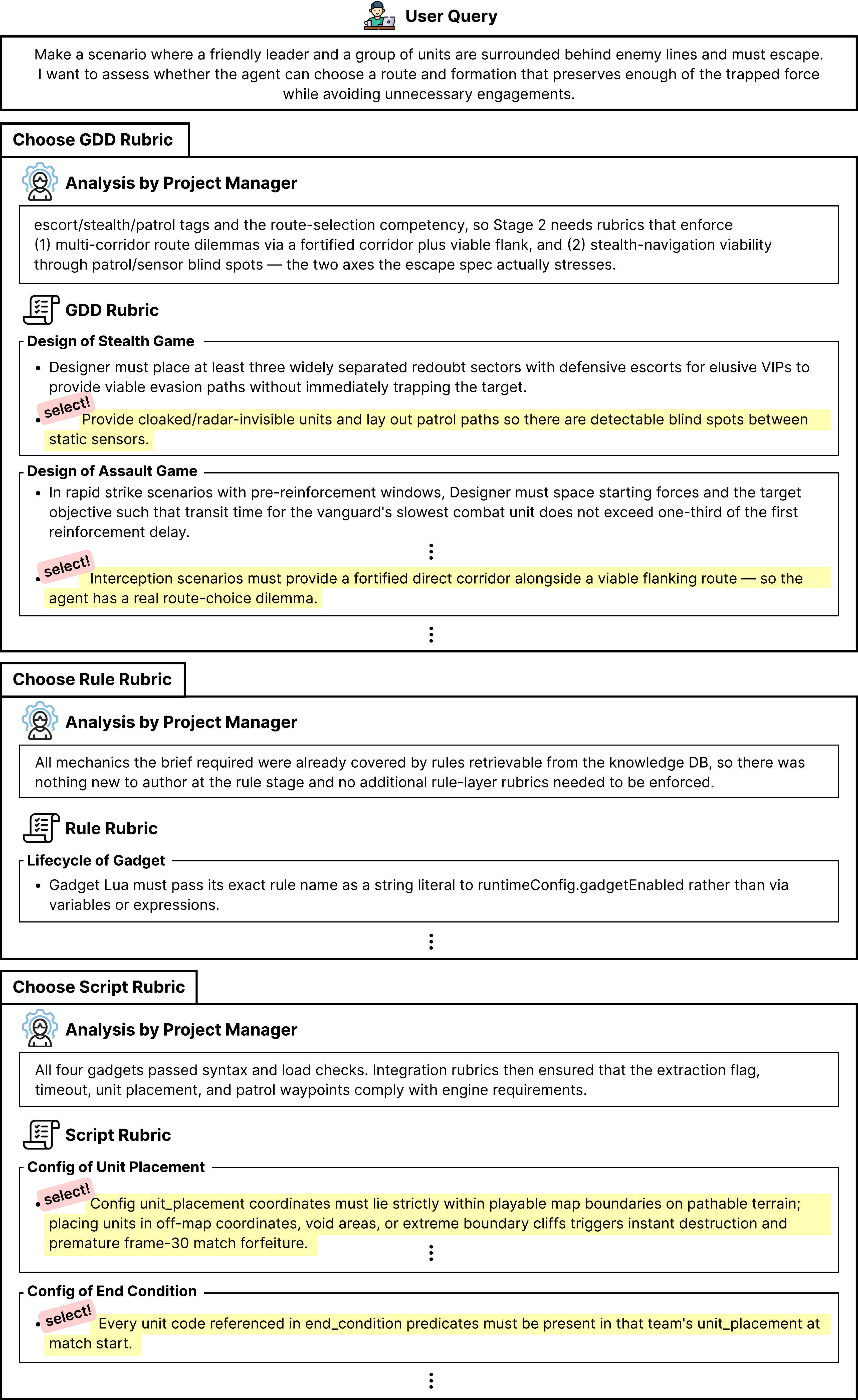}
        }
    \caption{\textbf{Rubric selection by the Project Manager.} For a user query, project manager chooses scope-appropriate rubrics at three authoring stages — GDD, Rule, and Script — and provides a short rationale for each selection.}
    \label{fig:rubric_select}
\end{figure}

\noindent \textbf{Inputs and initialization.}
All four agents, \ie, the PM, designer, developer, and analyst, are driven by Gemini-3.8-Flash.
The framework takes as input a user query $q$ and a shared knowledge database $\mathcal{D}$, which holds validated GDDs and rule sets, the validation rubrics $\mathcal{R}$, the game assets available in BAR, and the five human-designed diagnostic mini-games (Sec.~\ref{sec:minigame_bench}) at their three difficulty levels, which serve as reference scenarios $\mathcal{S}_{\text{ref}}$ for difficulty scaling.
GDDs, rule sets, and rubrics start empty and accumulate only through self-evolution, whereas the game assets and $\mathcal{S}_{\text{ref}}$ remain fixed.
Because few validation criteria apply across stages, the rubrics are maintained separately for each stage, $\mathcal{R} = \{\mathcal{R}^{(2)}, \mathcal{R}^{(3)}, \mathcal{R}^{(4)}\}$, each organized as a two-level hierarchy that groups individual rubrics under broad categories.
Before each validation, the PM inspects both the categories and their rubrics for that stage and selects those relevant to the current game (Fig.~\ref{fig:rubric_select}), denoted $\mathcal{R}_g \subseteq \mathcal{R}^{(2)}$, $\mathcal{R}_r \subseteq \mathcal{R}^{(3)}$, and $\mathcal{R}_s \subseteq \mathcal{R}^{(4)}$.

An artifact buffer $\mathcal{A}$ and a feedback log $\mathcal{F}$ start empty; $\mathcal{A}$ accumulates validated outputs for downstream reuse, and $\mathcal{F}$ records all analyst feedback, including failed attempts, for rollbacks and the final retrospective analysis.

\noindent \textbf{Rollbacks and aborts.}
Stages~2--4 share the same control flow for validation failures.
When a stage fails, the PM invokes \textsc{MetaFeedback}, which reviews $\mathcal{F}$, including the failure history and the approaches attempted so far, and rolls the pipeline back to the stage it identifies as the source of the failure, either the failing stage or an earlier one, with corrective guidance $\text{fb}^\star$ for the next attempt.
Artifacts of the stages being redone are discarded from $\mathcal{A}$.
The PM can perform at most three rollbacks per generation; if validation still fails, generation is aborted and outputs what it has produced so far, \ie, the latest game script if Stage~4 has been reached and the earlier artifacts otherwise.
Such \textit{aborted} generations are included in the evaluation (Sec.~\ref{supp:eval_generated_games}), and experience-driven evolution is expected to reduce their number over successive cycles.

\noindent \textbf{Stage 1: scenario brief generation.}
The designer engages in multi-turn dialogue with the user to clarify the intent of $q$, producing a structured scenario brief $b$ that specifies the map, game composition, enemy behavior rules, win/loss conditions, and proposed ratings of the five strategic demands (Sec.~\ref{sec:minigame_bench}) on a 0--5 scale.
The PM then compares $b$ with all GDDs in $\mathcal{D}$ and judges whether any is similar enough to reuse, deciding case by case whether a partial match suffices.
If so, the pipeline \textit{fast-tracks} to Stage~4 by retrieving that GDD, its checklist, and its rule set from $\mathcal{D}$, bypassing Stages~2 and~3.

\noindent \textbf{Stage 2: GDD generation.}
The designer expands $b$ into a full game design document (GDD) that finalizes the demand ratings, with the focal demand rated highest, and specifies the rule components governing game behavior, \eg, unit spawning conditions and termination criteria, together with a checklist $\mathcal{C}$ for validating the final game.
The analyst validates the GDD against $\mathcal{R}_g$, returning a binary pass signal $\text{pass}_g$ and structured feedback $\text{fb}_g$.
Once the GDD passes, it is stored in $\mathcal{A}$ together with $\mathcal{C}$.

\noindent \textbf{Stage 3: rule set construction.}
The PM first retrieves verified implementations of the rules required by the GDD from $\mathcal{D}$, when available.
The developer then implements the remaining rules as Lua scripts~\citep{ierusalimschy2006lua} for in-engine execution and writes test code $\mathcal{T}_r$ for the rule set.
These simulation tests run in the BAR engine without a playing agent: each sets up a predefined situation and checks, in a rule-based manner, whether the specified behavior occurs, \eg, whether units spawn at the specified time.
The analyst validates the rule set by running $\mathcal{T}_r$ and checking it against $\mathcal{R}_r$, returning $\text{pass}_r$ and $\text{fb}_r$, and the passing rule set is appended to $\mathcal{A}$.

\noindent \textbf{Stage 4: game implementation and verification.}
Based on $b$ and the artifacts in $\mathcal{A}$, the developer retrieves game assets, \eg, unit configurations, from $\mathcal{D}$ and assembles an executable game script $\mathcal{G}$ at medium difficulty on the map specified in $b$, setting unit placement, end conditions, and rule parameters.
It also writes simulation tests $\mathcal{T}_s$ for the assembled game in the same form as in Stage~3.
The analyst validates $\mathcal{G}$ by running $\mathcal{T}_s$ and checking it against $\mathcal{R}_s$ and the checklist $\mathcal{C}$, returning $\text{pass}_s$ and $\text{fb}_s$.

\noindent \textbf{Difficulty scaling.}
Once $\mathcal{G}$ passes validation, the developer generates the scripts at all three difficulty levels, $\mathcal{G}_{1:3}$ for Lv.1--3, at once, taking $\mathcal{G}$ as the medium-difficulty reference and using $\mathcal{S}_{\text{ref}}$ as in-context examples that pair each reference scenario's three levels with the rationale for their difficulty.
Elements shared across levels, \ie, the map, match format, active mechanics, and inference interval, are kept fixed, and only difficulty parameters, \eg, the number of enemy waves, unit counts, resource caps, time limits, and enemy AI aggressiveness, are varied within the ranges observed in $\mathcal{S}_{\text{ref}}$.
Scaling concludes generation without further validation, as the scaled scripts share the rule set validated for $\mathcal{G}$ and differ from it only in difficulty parameters.
The resulting games are evaluated as described in Sec.~\ref{supp:eval_generated_games}.

\noindent \textbf{Self-evolution phase.}
Upon successful generation, two self-evolution mechanisms are triggered.
First, the validated artifacts in $\mathcal{A}$, \ie, the GDD, its checklist, and the rule set, are committed to the database, $\mathcal{D} \leftarrow \mathcal{D} \cup \mathcal{A}$, so that future queries can reuse them to fast-track or partially skip stages.
The final scripts at the three difficulty levels are not stored, so $\mathcal{S}_{\text{ref}}$ remains the only reference for difficulty scaling.
Second, the PM conducts a \textit{retrospective analysis} over $\mathcal{F}$, including failed attempts, identifying systematic discrepancies between verification outcomes and quality expectations and updating the rubrics accordingly, $\mathcal{R} \leftarrow \textsc{PM.Retrospective}(\mathcal{R}, \mathcal{F})$.
Because updates are applied within the relevant category of each stage's hierarchy, validation experience accumulates where it applies instead of in a single monolithic rubric.
Together, these mechanisms improve both generation efficiency and output quality over successive cycles (Sec.~\ref{sec:eval_self-evol}), and each contributes to both gains (Sec.~\ref{supp:abl_self_evol_gen_framework}).

\begin{algorithm}[!ht]
    \caption{Behavioral Fidelity Evaluation}
    \label{alg:evaluator}
    \begin{algorithmic}[1]
        \Require User query $q$; game design document GDD; game scripts $\mathcal{G}_1, \mathcal{G}_2, \mathcal{G}_3$ for the three difficulty levels, or the latest (possibly incomplete) $\mathcal{G}$ if aborted during Stage~4; test agent $\pi$ (\method, playing to win); rollout budget $M{=}6$
        \Ensure Behavioral fidelity score $y \in [0, 1]$, or $0$ if no game script exists

        \If{no game script exists} \Comment{Aborted before Stage 4}
            \State \Return $y = 0$ \Comment{No game to play}
        \EndIf

        \Statex \textcolor{gray}{\textit{// Checklist Construction}}
        \State $\mathcal{E} \leftarrow \Call{BuildChecklist}{q, \text{GDD}}$ \Comment{Behaviors the game must exhibit}
        \Statex \hspace{\algorithmicindent}$\mathcal{E}_{\text{init}}$: the game starts with the specified rosters and map/zones
        \Statex \hspace{\algorithmicindent}$\mathcal{E}_{\text{rule}}$: per required rule, it fires on time and takes its intended effect
        \Statex \hspace{\algorithmicindent}$\mathcal{E}_{\text{end}}$: victory (defeat) is declared when its condition holds
        \Statex \hspace{\algorithmicindent}$\mathcal{E}_{\text{pace}}$: up to three expected events on the pacing timeline occur

        \Statex \textcolor{gray}{\textit{// Rollouts with Privileged Logging}}
        \State $\Gamma \leftarrow \{\mathcal{G}_1, \mathcal{G}_2, \mathcal{G}_3\}$, or $\{\mathcal{G}\}$ if aborted
        \State $R \leftarrow M / |\Gamma|$ \Comment{2 per level, or 6 if aborted}
        \ForAll{$\mathcal{G}_\ell \in \Gamma$ and $i = 1, \dots, R$ \textbf{in parallel}}
            \State $\tau_{\ell,i} \leftarrow \Call{Rollout}{\mathcal{G}_\ell, \pi}$ \Comment{Full simulator state until the game ends or crashes}
        \EndFor

        \Statex \textcolor{gray}{\textit{// Checklist Matching}}
        \ForAll{$e \in \mathcal{E}$ and rollouts $\tau_{\ell,i}$}
            \State $v_{e,\ell,i} \leftarrow \Call{Judge}{e, \tau_{\ell,i}} \in \{0, 1\}$ \Comment{1 if the expected behavior is found}
        \EndFor
        \State $y \leftarrow \frac{1}{|\mathcal{E}|} \sum_{e \in \mathcal{E}} \max_{\ell,i} v_{e,\ell,i}$ \Comment{An item passes if found in any rollout}
        \State \Return $y$
    \end{algorithmic}
\end{algorithm}

\subsection{Evaluation of Generated Games}
\label{supp:eval_generated_games}

Evaluating generated games is challenging for several reasons.
First, there is no reference against which a generated game can be compared: each game is created on demand from a free-form query, so its correctness can only be judged against the intent expressed in that query.
Second, whether a game realizes this intent becomes observable only in play.
A game script can pass static inspection and even run without errors while failing to produce the intended dynamics, and many intended behaviors, such as victory and defeat, arise only in particular situations whose occurrence depends on how the game is played, so a single play-through may never reach them.
Third, human judgment captures intent most faithfully, but it is costly to scale and does not by itself indicate which behaviors fail.
Finally, the framework itself can fail: a generation may abort before producing a complete game, and one that succeeds may do so only after costly rollbacks.
We therefore evaluate generated games along four complementary criteria:
\textit{playability}, a coarse check of whether generation yields a game that can be launched;
\textit{behavioral fidelity}, a fine-grained check of whether the behaviors expected from the query and GDD actually occur in play;
\textit{token usage}, the inference cost of producing a game;
and \textit{human preference}, a judgment of query alignment, fun and balance, and completeness by human raters.
All four are measured for generation both with and without experience-driven evolution (Sec.~\ref{sec:eval_self-evol}) and reported per batch of five consecutive generation cycles, so that the effect of accumulated experience can be traced over successive cycles.
Aborted generations are included in every criterion, so that failures are reflected in the results rather than excluded.

\subsubsection{Playability}
\label{supp:playability}

Playability is a coarse check of whether generation yields a game that can be launched at all.
For each batch, it is computed as
\[
\text{Playability} = \frac{\#\,\text{games that launch successfully}}{\#\,\text{generation attempts}}.
\]
Each game is launched in the BAR engine once at each of the three difficulty levels, or once from its latest script if aborted, and counts as launching successfully if the game starts in every launch; what happens after the game starts is not considered.
The denominator includes every generation attempt, including aborted ones: a generation aborted before Stage~4 produces no game and counts as a failure, whereas an aborted game whose latest script still launches counts as playable.
Playability does not check whether the game behaves as intended; a game may launch even if its enemy waves never spawn or a rule script fails midway.
Such failures are captured by behavioral fidelity.

\subsubsection{Behavioral Fidelity}
\label{supp:evaluator}

\begin{figure}[t]
    \centering
    \resizebox{\linewidth}{!}{
        \includegraphics{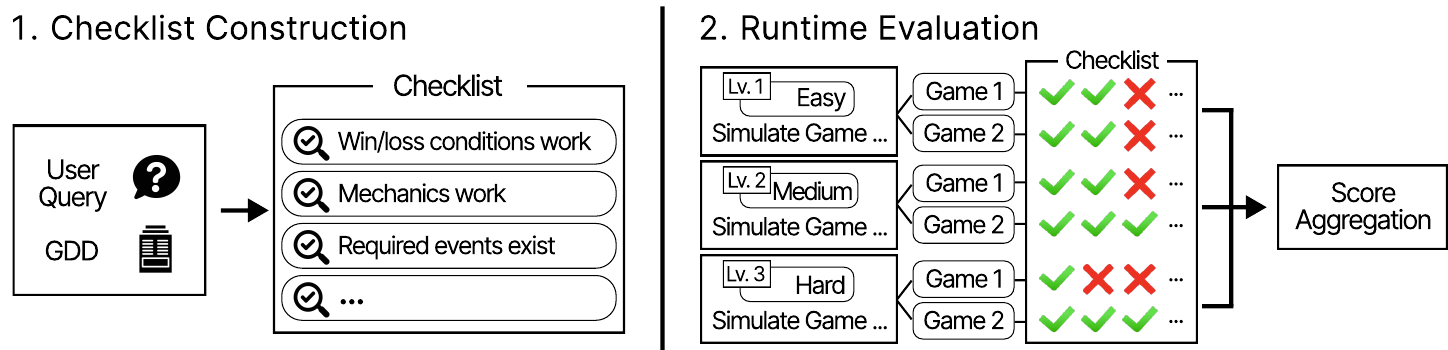}
        }
    \caption{\textbf{Behavioral fidelity evaluation.} A checklist of expected behaviors is built from the user query and GDD, and each game is played twice at each difficulty level. A judge checks each item against the log of every rollout, and an item passes if it is observed in any rollout; the score is the fraction of passing items.}
    \label{fig:evaluator}
\end{figure}

Behavioral fidelity starts from the premise that a user's intent is ultimately realized through the in-game behaviors of the generated game.
A game aligned with its query exhibits the behaviors that the query and its GDD call for, and it can do so only if its implementation functions correctly.
We check whether these expected behaviors actually occur when the game is played and summarize the result as a score from 0 to 1, \ie, the checklist pass rate, which reflects both alignment with the user's intent and functionality (Fig.~\ref{fig:evaluator}; Algorithm~\ref{alg:evaluator}).
Aborted generations are scored in the same way whenever their latest script launches, as described below; only generations that yield no launchable game receive a score of 0.
We report the mean and standard deviation of the scores over the five generations in each batch, including aborted ones.

\noindent \textbf{Checklist construction.}
Gemini-3.8-Flash reads the user query and the GDD and constructs an evaluation checklist $\mathcal{E}$, typically of around ten behaviors that the game must exhibit in play, separately from the checklist used for validation during generation.
Because it is derived from the GDD, the same checklist applies to all three difficulty levels.
It covers the following categories:
\textit{rosters and map/zones}, checking whether the game starts with the specified units, placements, and zones;
\textit{mechanics}, with two items per rule required by the GDD, one checking whether the rule fires at the intended time and the other whether it takes its intended effect;
the \textit{victory} and \textit{defeat} specifications, checking whether the game is won or lost when the corresponding condition holds;
and the \textit{pacing timeline}, with up to three items checking whether events expected at points on the timeline occur.
For example, for a game that restricts the player to static defenses, the mechanics items check whether the production restriction is in force from the start and whether the player's construction units are indeed unable to build factories or air units, and a pacing item checks whether the first enemy wave arrives through the western ramp at around 30\,s.

\noindent \textbf{Rollouts.}
Each game is played twice at each of the three difficulty levels, for six rollouts in total, all run in parallel.
The rollouts are played by \method (Sec.~\ref{sec:rtsagent}) with replanning and memory enabled and Gemini-3.8-Flash as its backbone, playing to win.
Covering all levels is necessary to exercise the whole checklist: victory is typically observed at easier levels, whereas defeat may only be reached at harder ones.
Conversely, if defeat is never observed even at Lv.3, the defeat item fails, so the checklist also reveals games that remain too easy even at their hardest level.
Aborted games have no difficulty levels, so their latest script $\mathcal{G}$, which may be incomplete, is played six times to keep the number of rollouts equal across games; behaviors that are already implemented can still be verified even if other parts of the game do not work.

\noindent \textbf{Privileged logging.}
Since the rollouts serve evaluation rather than play, we log the full simulator state at every step rather than only what the agent observes, including the positions and states of all units of every team, even those outside the agent's field of view, together with engine logs such as rule-script outputs and Lua errors.
Logging continues until the game ends or crashes; a crash truncates the log, leaving the behaviors expected afterwards unobserved.

\noindent \textbf{Checklist matching.}
A judge, Gemini-3.8-Flash, examines the log of each rollout separately and determines, for each checklist item, whether the expected behavior is found.
An item passes if it is found in at least one of the six rollouts, regardless of difficulty level, and fails if it never occurs as expected; the score of a game is the fraction of its items that pass.
The score therefore measures whether each expected behavior is realized by the implemented game, not how consistently it recurs across rollouts.
Although the judge shares its backbone with the generation agents, both conditions compared in Sec.~\ref{sec:eval_self-evol}, \ie, generation with and without experience-driven evolution, use the same backbone, so any self-preference of the judge applies equally to both.

\subsubsection{Token Usage}
\label{supp:token_usage}
\label{supp:gen_time} 

Token usage measures the cost of producing a game as the total number of input and output tokens consumed by all agents in the framework, from receiving a user query until it outputs a game, including all stages, rollbacks, and difficulty scaling.
Unlike wall-clock time, token counts do not depend on API latency or hardware, so costs remain comparable across generations.
Token usage is measured for every generation, including aborted ones, whose count covers all calls until the abort; since aborted generations keep running until the rollback budget is exhausted, they tend to consume more tokens, so aborts also raise the average.
We report the mean over the five generations in each batch, together with its standard deviation.
Experience-driven evolution can reduce this cost in two ways: later queries can fast-track to Stage~4 or reuse verified rule implementations, and refined rubrics help avoid recurring mistakes and the rollbacks they cause.
This criterion thus reflects how much accumulated experience reduces redundant work and repeated failures.

\subsubsection{Human Preference}
\label{supp:human_pref_gen_game}

Human preference assesses the generated games as a whole, beyond the individual behaviors covered by the checklist.
We conduct a blind pairwise study with four RTS-experienced raters, none involved in the benchmark design.
From each of the ten batches, we sample two queries for which generation completed both with and without experience-driven evolution, yielding 20 game pairs.
Each pair is presented side by side in random order without revealing the source, and every rater plays both games at Lv.2.
Since aborted generations, which occur more often without evolution, are excluded, the comparison is, if anything, conservative for the evolved games.

For each pair, raters judge which game is better, or declare a tie, on three criteria:
\textit{query alignment}, whether the game realizes the requested scenario, with the described units, objectives, and threats present and the decision the query intends to assess actually arising in play;
\textit{fun and balance}, whether the game is enjoyable and fairly challenging, with difficulty neither trivial nor impossible, outcomes attributable to the player's decisions, and sustained tension;
and \textit{completeness}, whether the game is functionally well built, running without errors, triggering rules and win/loss conditions as intended, and forming a coherent whole of map, units, and rules with understandable objectives.
Each rater's final preference for a pair is the game that wins more criteria, or a tie if both win equally many.
For each batch, we report the win, tie, and loss rates of the evolved game over eight judgments (four raters on two pairs).

Inter-rater agreement, measured by Krippendorff's $\alpha$~\citep{krippendorff2011computing} with win $>$ tie $>$ loss as an ordinal scale, is 0.62 for query alignment, 0.51 for fun and balance, 0.59 for completeness, and 0.58 for the final preference, indicating moderate agreement.
Fun and balance, the most subjective criterion, shows the lowest agreement.

\begin{figure}[t]
    \centering
    \resizebox{0.6\linewidth}{!}{
        \includegraphics{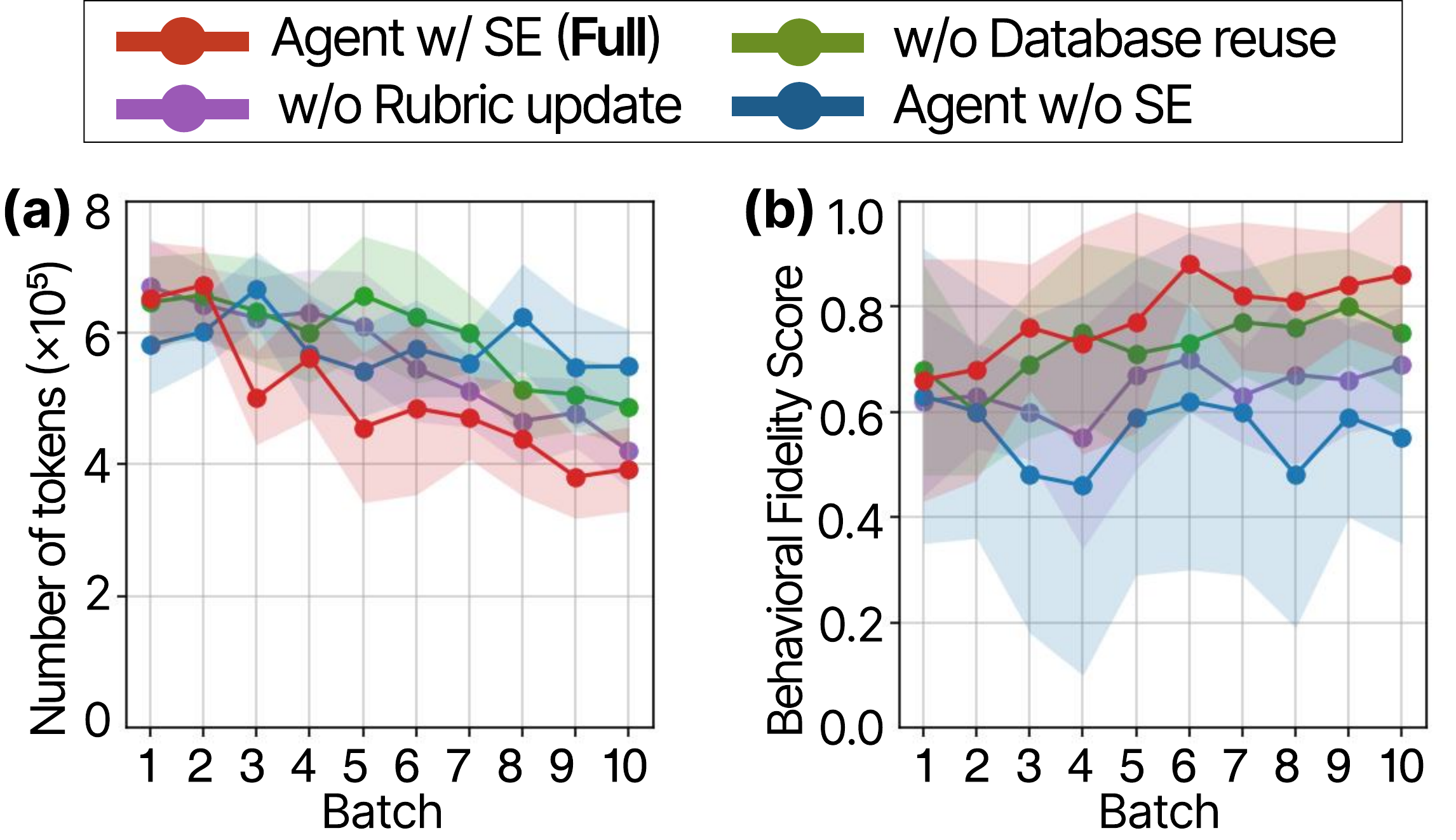}
        }
    \vspace{-1em}
    \caption{\textbf{Ablation of self-evolution mechanisms.} (a)~Token usage and (b)~behavioral fidelity across ten batches when removing database reuse, rubric updates, or both from \textit{Agent w/ SE}. Shaded areas denote standard deviation.}
    \label{fig:generated_ablation}
    \vspace{-1em}
\end{figure}

\subsubsection{Implementation of the End-to-End Baselines}
\label{supp:e2e_impl}

The end-to-end (E2E) baselines generate a complete game in a single LLM call instead of the staged pipeline, using the same backbone (Gemini-3.8-Flash).
Along with the user query, this call receives the reference information available to the pipeline:
(i)~a manual that concatenates the role prompts for GDD writing, Lua rule writing, and game configuration, specifying the format and writing rules of each output;
(ii)~twelve candidate maps relevant to the query, with their terrain information;
(iii)~a catalog mapping Armada unit names to their in-game IDs;
and (iv)~examples of how configurations differ across difficulty levels.
As with \textit{Agent w/o SE}, the baselines receive no rubrics, stored rules, or previous generations, and the prompt states that the rule catalog is empty, so every mechanic must be implemented from scratch.
The call outputs a GDD, one Lua rule script per mechanic, and game configurations for the three difficulty levels.
The GDD plays the same role as in the pipeline; in particular, the behavioral fidelity checklist is derived from it (Appendix~\ref{supp:evaluator}).

\textit{E2E} applies only the framework's non-LLM checks, \ie, rule syntax checking and configuration validation, before storing the result, so each generation uses exactly one LLM call.
\textit{E2E w/ verify} additionally plays one game in the BAR engine. 
If the run shows Lua errors, a crash, or a rule that never fires, the logs are fed back and all outputs are regenerated, up to three times, for at most four LLM calls per generation.

\subsection{Ablation of Self-Evolution Mechanisms}
\label{supp:abl_self_evol_gen_framework}

We ablate the two self-evolution mechanisms of the framework ($\S$\ref{sec:minigame_gen}) on the same ten batches of five queries, measuring token usage and behavioral fidelity as in $\S$\ref{sec:eval_self-evol}.
Without database reuse, validated GDDs and rule sets are not reused, so every query passes through all stages and implements every rule anew; without rubric updates, the retrospective analysis is disabled and the rubrics remain at their initial state.
As shown in Fig.~\ref{fig:generated_ablation}, \textit{Agent w/ SE} reaches the lowest token usage and the highest behavioral fidelity in later batches, whereas \textit{Agent w/o SE} shows neither trend.
Each single-mechanism variant retains part of the gains, with token usage still declining and fidelity still rising, but less than with \textit{Agent w/ SE}.
The two mechanisms also appear to contribute differently.
Removing database reuse leaves the highest token usage among the single-mechanism variants, suggesting that reuse mainly reduces cost by bypassing redundant stages, whereas removing rubric updates yields lower fidelity, suggesting that refined rubrics mainly improve quality by preventing recurring design mistakes.
As the standard deviations overlap in several batches, we regard this distinction as indicative rather than conclusive.

\subsection{Examples of Self-Evolving Game Generation}
\label{supp:selfevolving_examples}

\begin{figure}[t]
    \centering
    \resizebox{\linewidth}{!}{
        \includegraphics{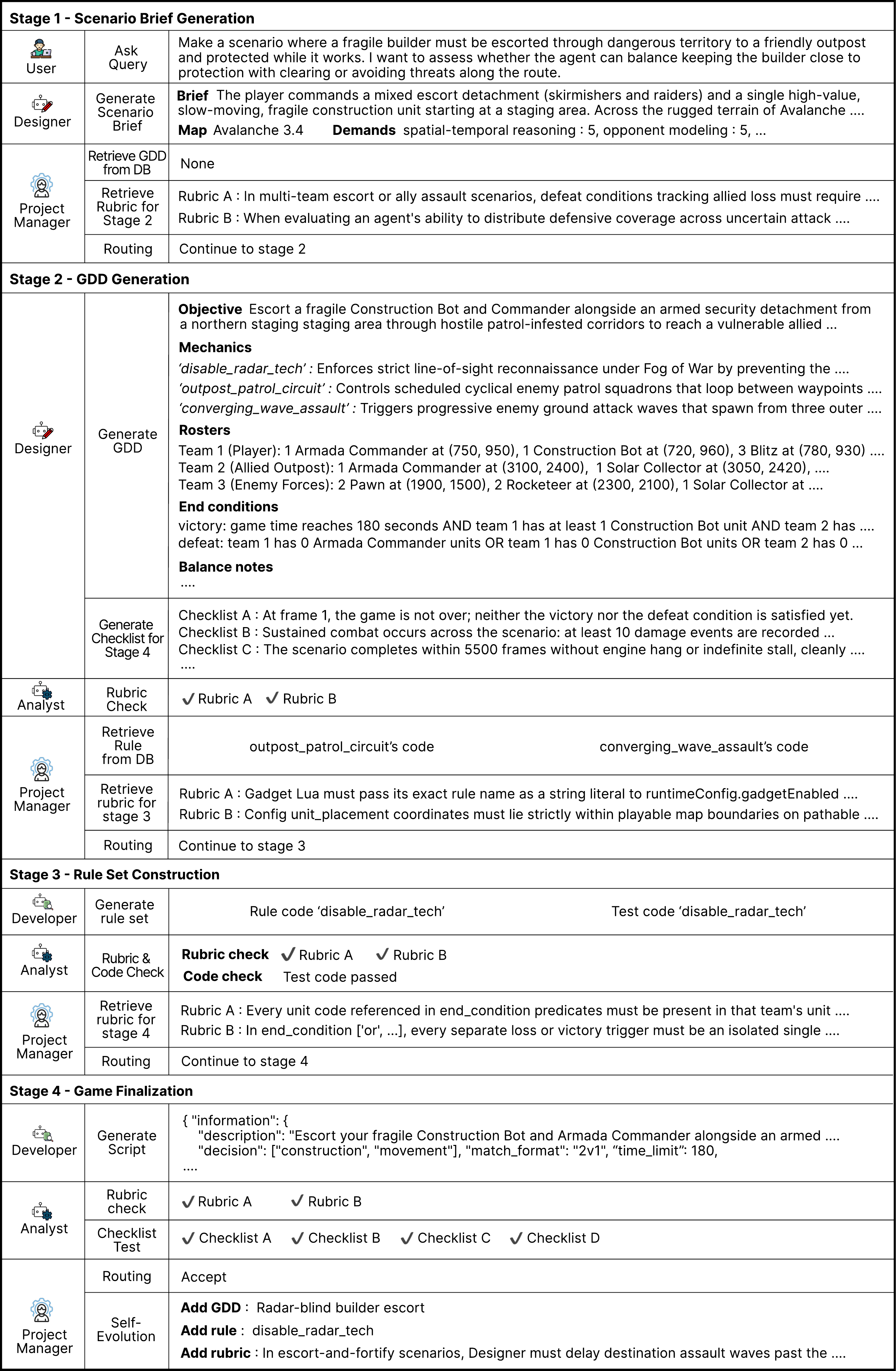}
        }
    \vspace{-1em}
    \caption{\textbf{Example of the game generation process.}}
    \label{fig:full_game_gen_process}
\end{figure}

\noindent \textbf{Procedure for game generation.}
Fig.~\ref{fig:full_game_gen_process} illustrates one generation cycle of Algorithm~\ref{alg:self_evolving}.
The user requests a scenario in which a fragile builder must be escorted to an allied outpost and protected while it works, testing whether the agent can balance keeping it close with clearing or avoiding threats.
In Stage~1, the designer turns the query into a scenario brief specifying the escort force, map, and targeted strategic demands; finding no matching GDD in the database, the PM routes generation to Stage~2 instead of fast-tracking.
In Stage~2, the designer writes a GDD with the objective, three mechanics, team rosters, and end conditions, together with a checklist for the final game, and the GDD passes the retrieved rubrics, \eg, one requiring defeat conditions in escort scenarios to track allied losses.
The PM then retrieves verified implementations of two mechanics from the database, so in Stage~3 the developer writes and tests only the remaining rule, which disables radar to enforce line-of-sight reconnaissance.
In Stage~4, the game script passes the rubric check and checklist tests and is scaled to three difficulty levels (omitted from the figure).
Finally, the self-evolution phase stores the new GDD, the new rule, and a new rubric for escort-and-fortify scenarios in the database.

\begin{figure}[t]
    \centering
    \includegraphics[width=\linewidth]{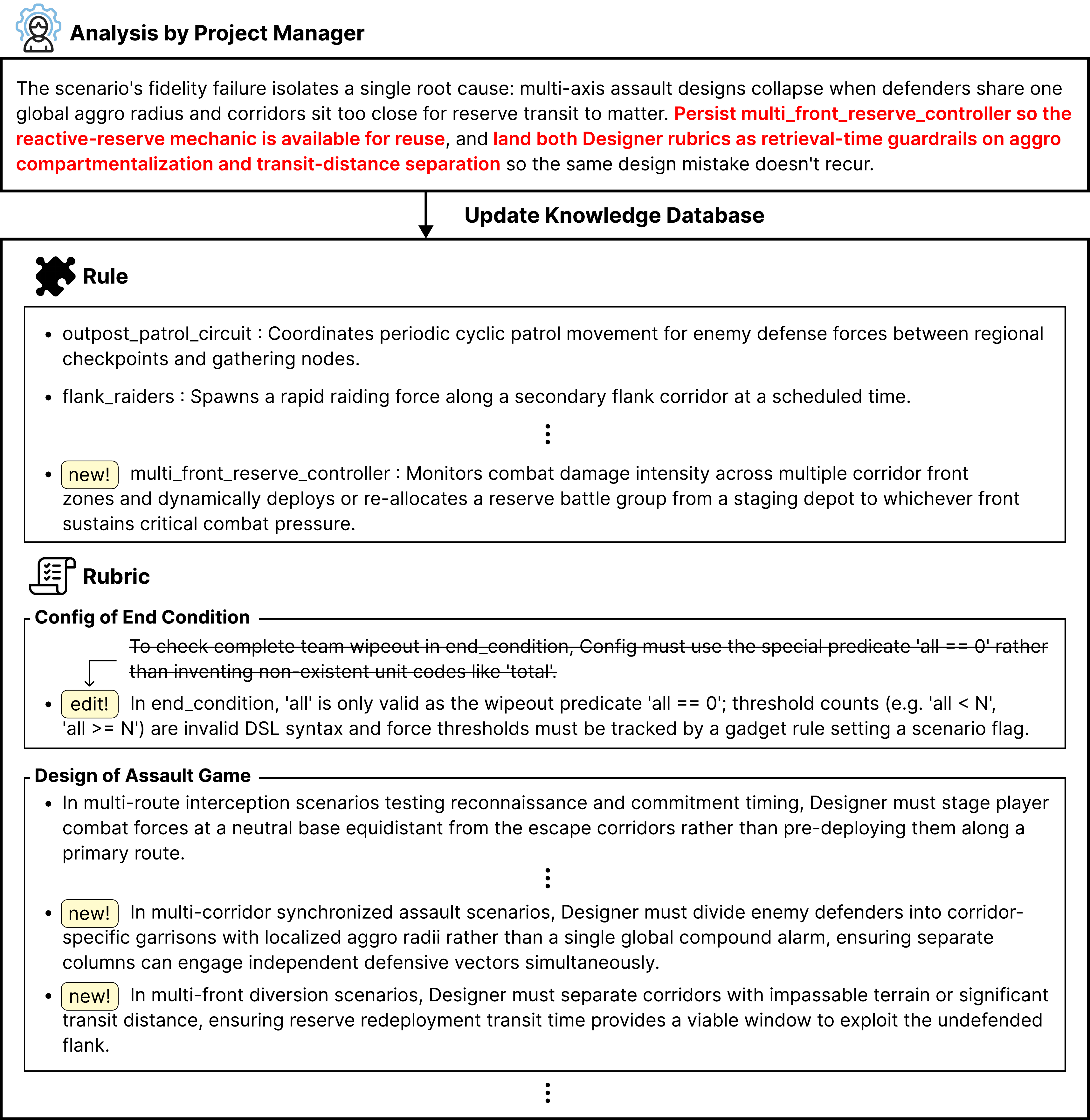}
    \caption{\textbf{Example of rule and rubric updates during Self-Evolving.} From the Project Manager's judgment highlighted in \textbf{red bold text}, reusable mechanics are appended to the rule catalog for later scenarios, and new rubrics are added to reduce the recurrence of the observed failure mode.}
    \label{fig:rubric_update_example}
\end{figure}

\noindent \textbf{Rule and rubric updates during self-evolution.}
Fig.~\ref{fig:rubric_update_example} shows how self-evolution updates the knowledge database.
Newly validated rules are committed for reuse, and through the retrospective analysis, the PM traces the failures in the feedback log to their root causes and distills the lessons into rubrics.
These updates either add new criteria that guard against the observed failure mode or refine existing ones that proved imprecise, and each is placed under the relevant category of the stage's hierarchy, \eg, end-condition configuration or assault game design.
Since later cycles retrieve these rubrics before validation, a recurring design mistake is checked for rather than rediscovered.

\noindent \textbf{Generated diagnostic games.}
Figs.~\ref{fig:generated_games1}--\ref{fig:generated_games5} present three representative mini-games generated for each of the five strategic demands.
Each example shows the free-form user query, the strategic demand ratings (0--5) recorded in its GDD, the resulting game scenario, and screenshots sampled from a full play-through.
The queries state only what the user wants to assess, and the framework grounds each into a concrete scenario with its own map, forces, objectives, and end conditions.
In every example, the focal demand receives the highest rating while the others are rated lower, mirroring the design of the human-designed mini-games (Sec.~\ref{sec:minigame_bench}).
Scenarios sharing the same focal demand also differ substantially, \eg, in whether the player acts alone or alongside an allied AI and whether it attacks, defends, or escorts, showing that the framework produces functionally diverse scenarios from free-form queries.

\subsection{Agent Evaluation on Generated Games}
\label{supp:exp_generated_games}

\begin{figure}[t]
    \centering
    \resizebox{\linewidth}{!}{
        \includegraphics{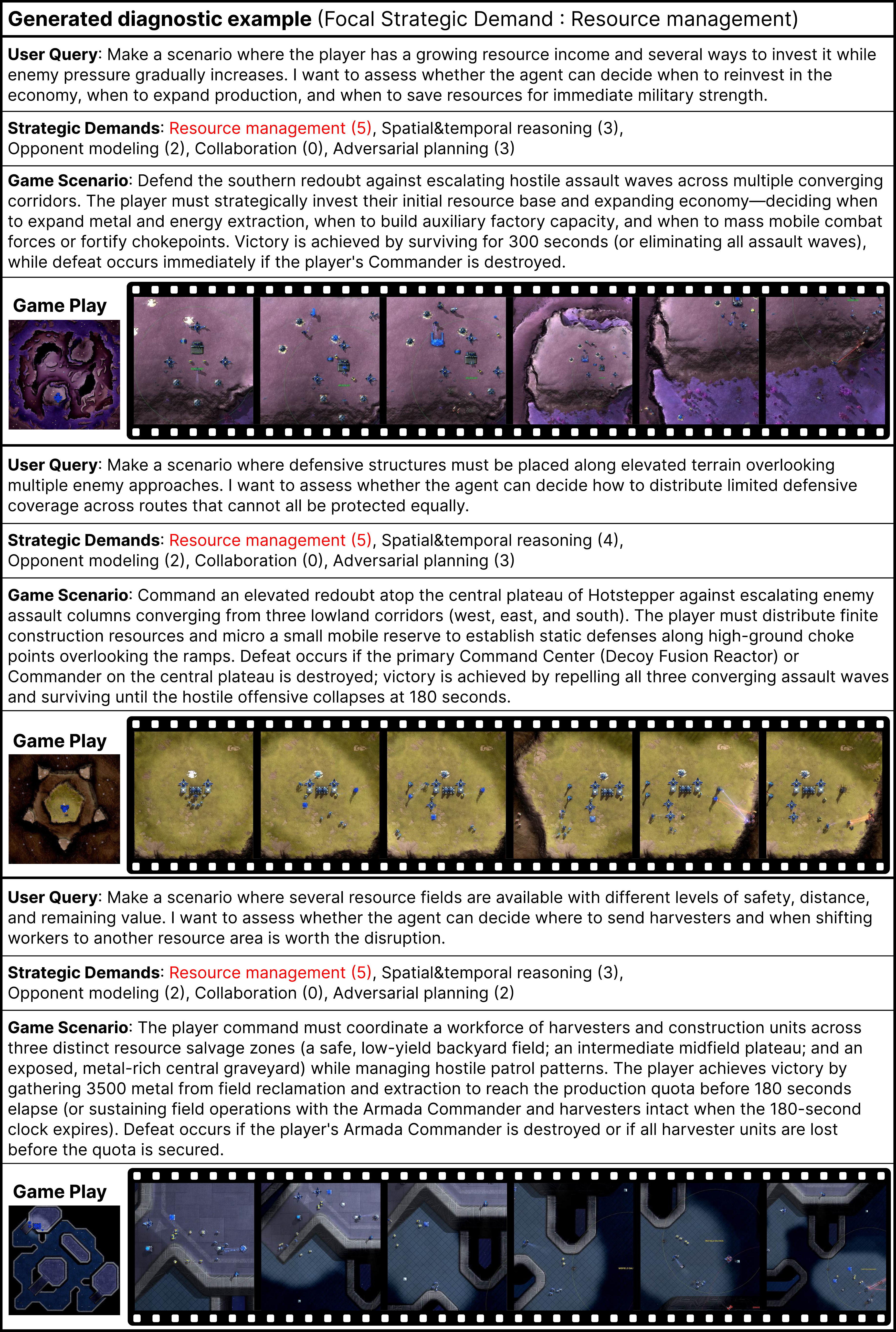}
        }
    \vspace{-1em}
    \caption{\textbf{A mini-game focused on resource management, generated by the self-evolving framework.}
    Each example shows the free-form user query, the strategic demands the generated scenario targets, the resulting game scenario, and screenshots sampled from a full play-through.}
    \label{fig:generated_games1}
\end{figure}

\begin{figure}[t]
    \centering
    \resizebox{\linewidth}{!}{
        \includegraphics{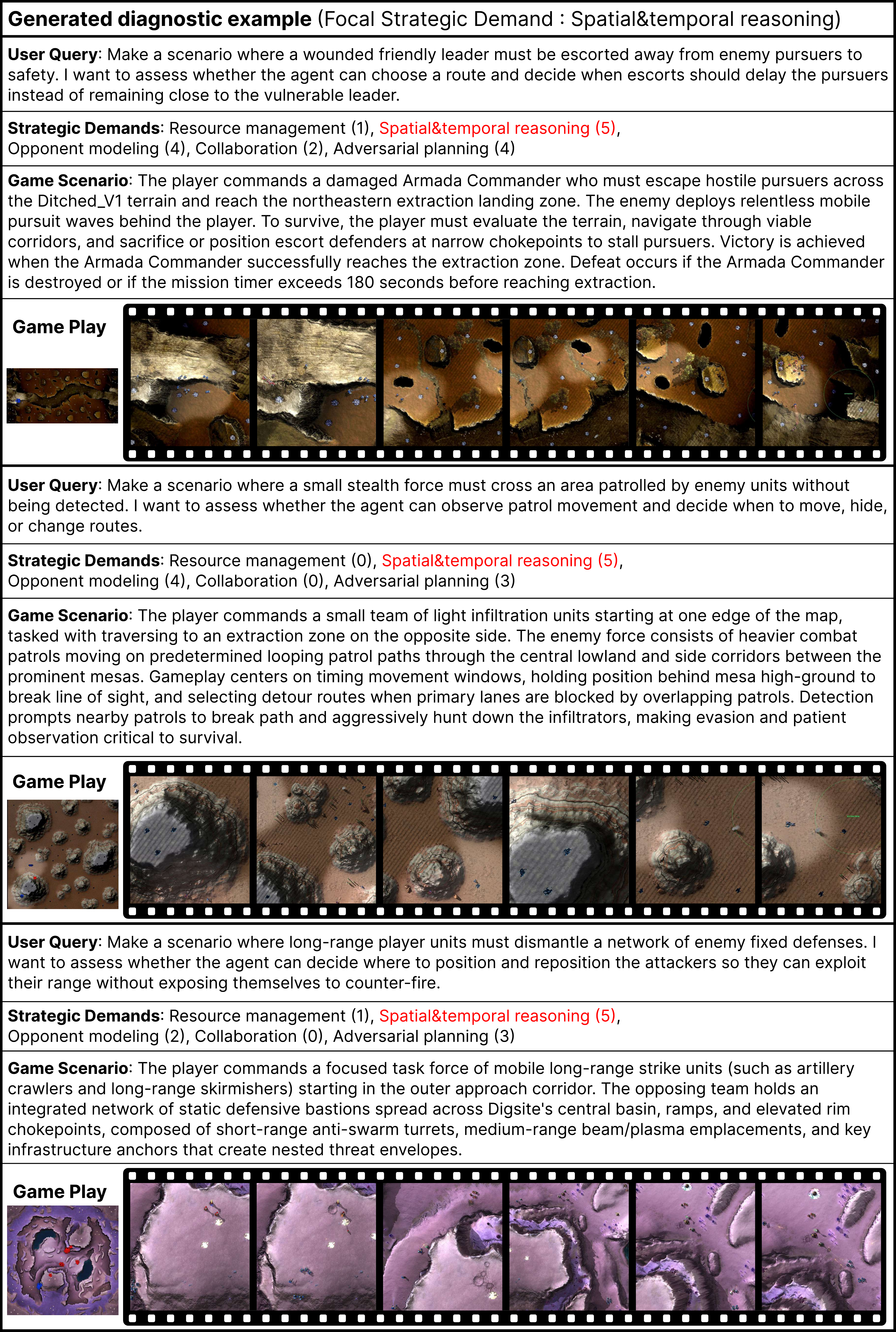}
        }
    \vspace{-1em}
    \caption{\textbf{A mini-game focused on spatial \& temporal reasoning, generated by the self-evolving framework.}
    Each example shows the free-form user query, the strategic demands the generated scenario targets, the resulting game scenario, and screenshots sampled from a full play-through.}
    \label{fig:generated_games2}
\end{figure}

\begin{figure}[t]
    \centering
    \resizebox{\linewidth}{!}{
        \includegraphics{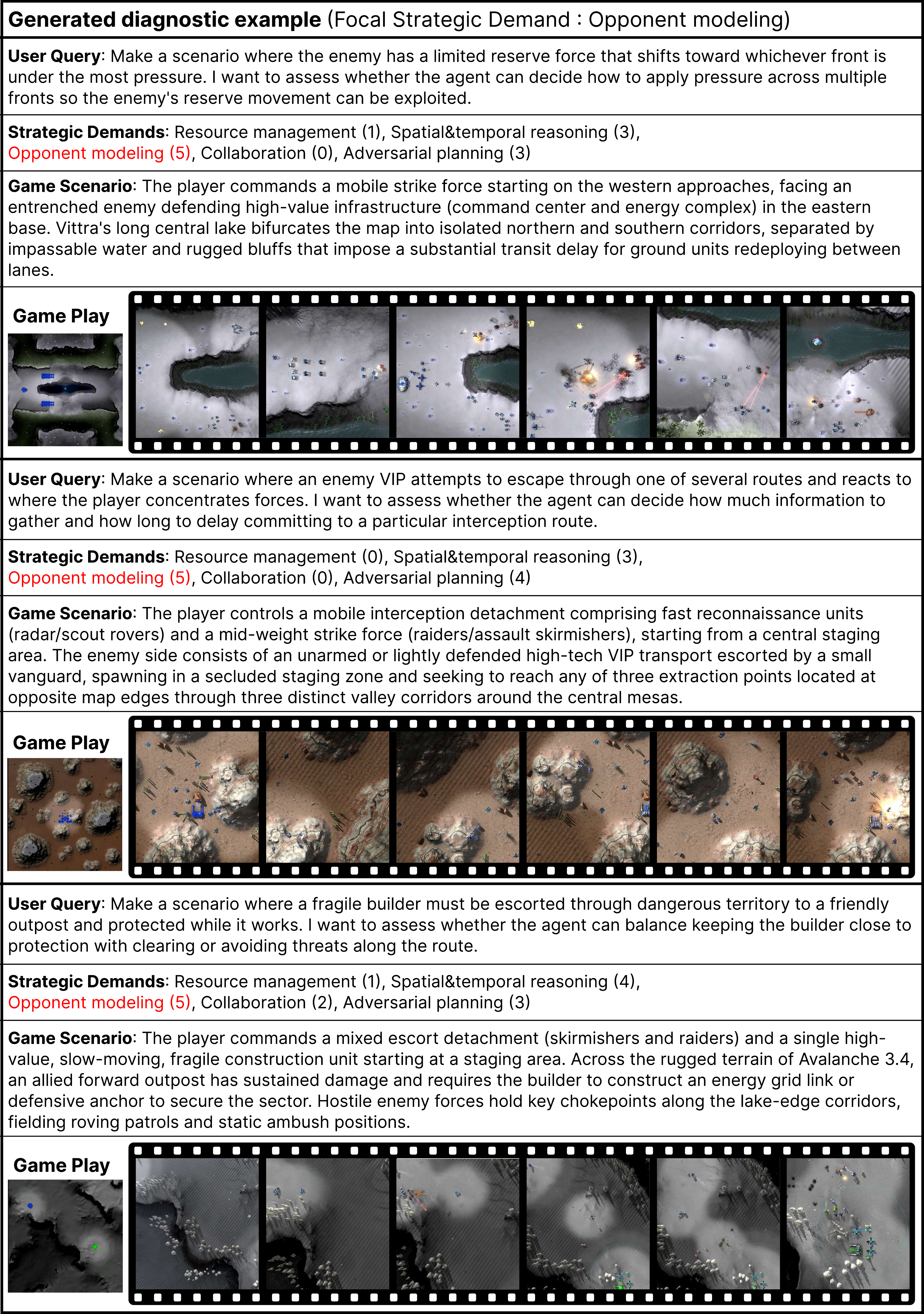}
        }
    \vspace{-1em}
    \caption{\textbf{A mini-game focused on opponent modeling, generated by the self-evolving framework.}
    Each example shows the free-form user query, the strategic demands the generated scenario targets, the resulting game scenario, and screenshots sampled from a full play-through.}
    \label{fig:generated_games3}
\end{figure}

\begin{figure}[t]
    \centering
    \resizebox{\linewidth}{!}{
        \includegraphics{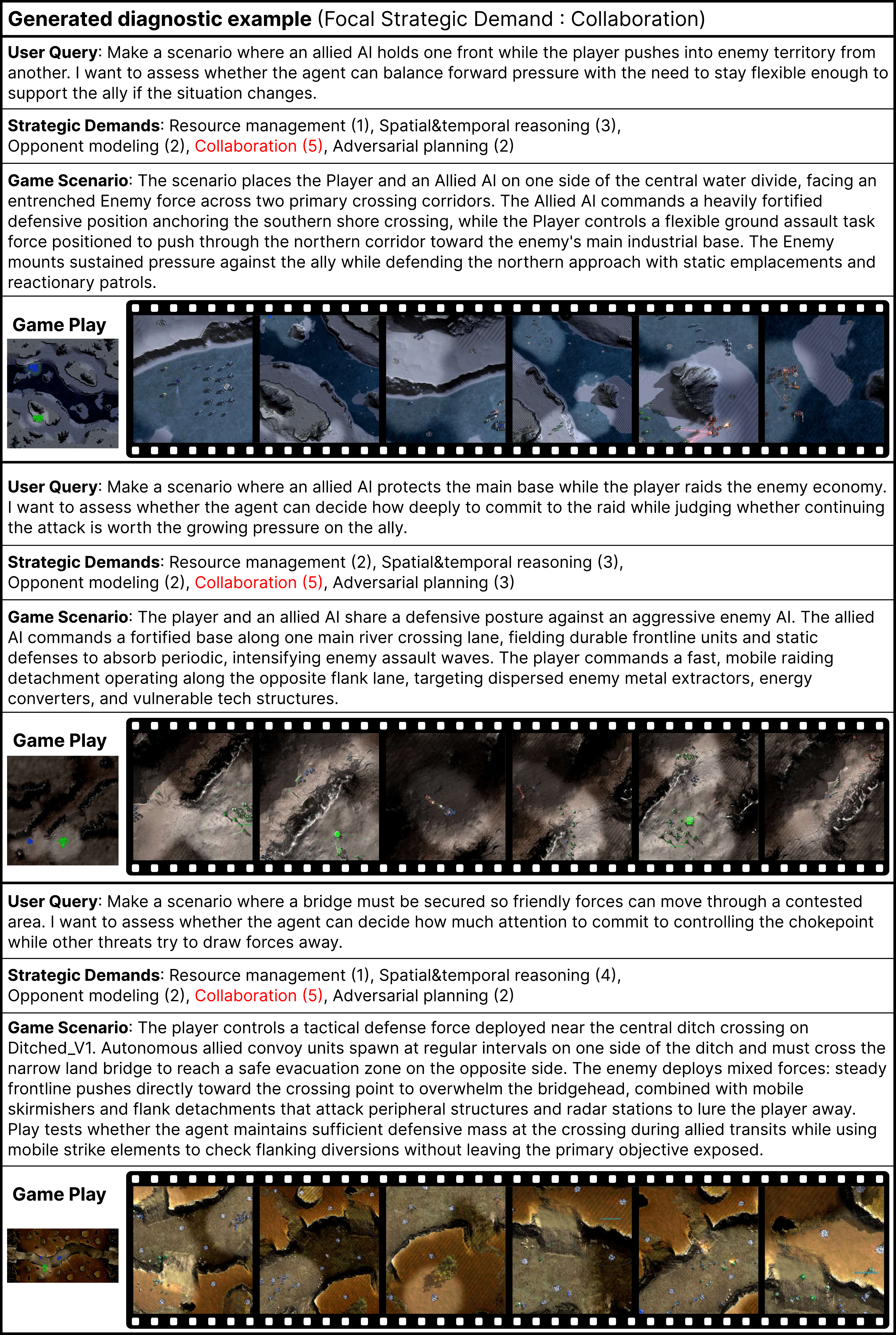}
        }
    \vspace{-1em}
    \caption{\textbf{A mini-game focused on collaboration, generated by the self-evolving framework.}
    Each example shows the free-form user query, the strategic demands the generated scenario targets, the resulting game scenario, and screenshots sampled from a full play-through.}
    \label{fig:generated_games4}
\end{figure}

\begin{figure}[t]
    \centering
    \resizebox{\linewidth}{!}{
        \includegraphics{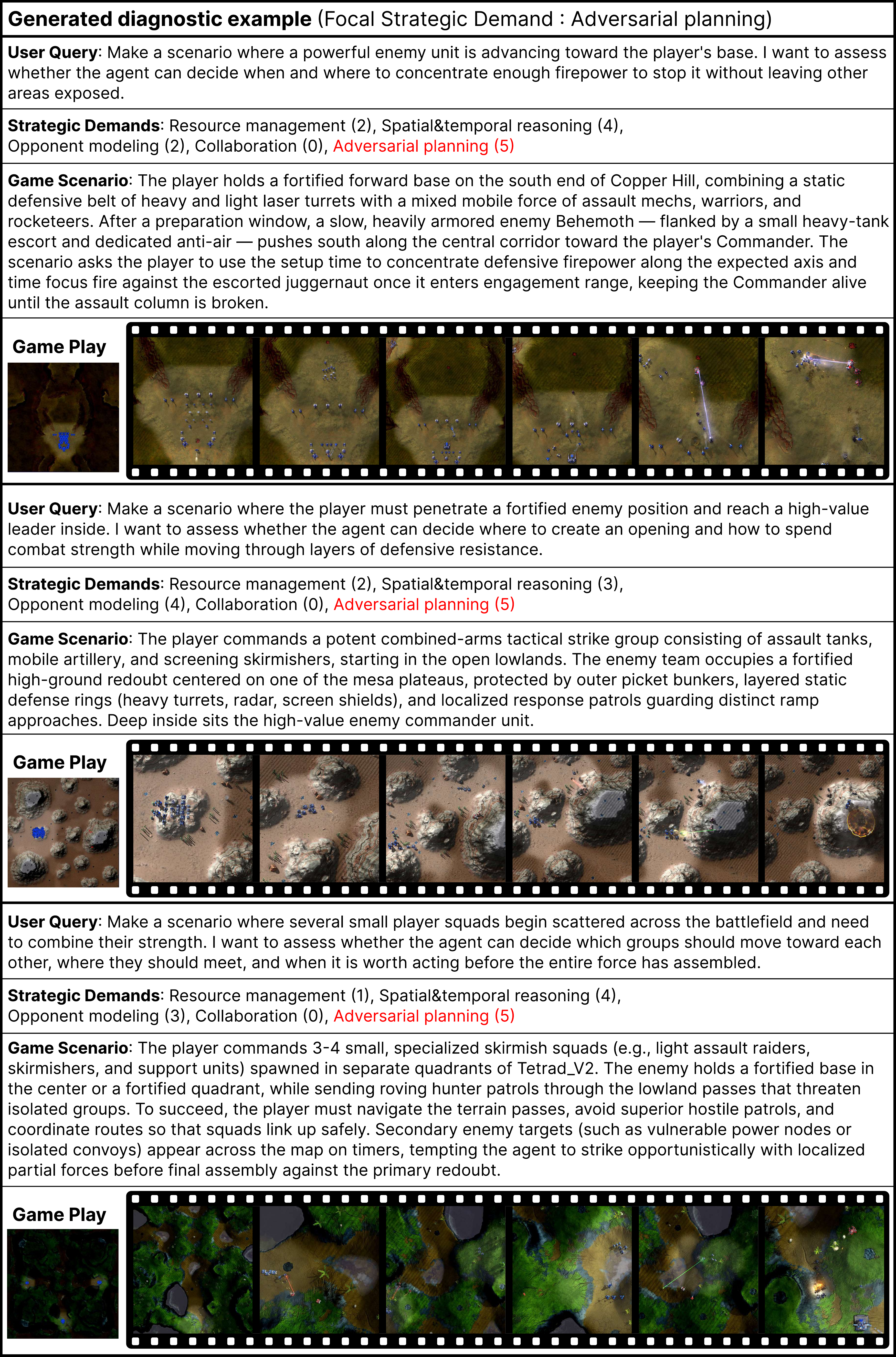}
        }
    \vspace{-1em}
    \caption{\textbf{A mini-game focused on adversarial planning, generated by the self-evolving framework.}
    Each example shows the free-form user query, the strategic demands the generated scenario targets, the resulting game scenario, and screenshots sampled from a full play-through.}
    \label{fig:generated_games5}
\end{figure}

As a preliminary check of generated games in agent evaluation, we evaluate \method with Gemini-3.7-Flash and GLM-5.3-Flash on five generated games, each created using the query from the top example in one of Figs.~\ref{fig:generated_games1}--\ref{fig:generated_games5}.
The five games are Economy Under Pressure (EUP, resource management), Wounded Leader Escape (WLE, spatial and temporal reasoning), Reserve Force Diversion (RFD, opponent modeling), Allied Front Coordination (AFC, collaboration), and Threat Response Allocation (TRA, adversarial planning), each played five times per level and evaluated by WR (Tab.~\ref{tab:generated_game_eval}).
For both backbones, WR never increases with the level, indicating that difficulty scaling yields graded difficulty.
No run wins at Lv.3, suggesting that it may be harder than Lv.3 of the human-designed mini-games, although five runs cannot rule out occasional wins.
Gemini-3.7-Flash matches or exceeds GLM-5.3-Flash in every game except EUP, where GLM-5.3-Flash wins one more game at Lv.2, and the gap is largest on AFC and TRA.
This pattern loosely mirrors the human-designed mini-games, where GLM-5.3-Flash is competitive on TCP but falls behind on ATS and AWD (Tab.~\ref{tab:minigame_results}), although the small number of games and runs limits this comparison.

\begin{table}[t]
\caption{\textbf{Agent evaluation on generated games.} WR of \method with two backbones on one generated game per strategic demand, over five runs per level.}
\label{tab:generated_game_eval}
\vspace{0.5em}
\centering
\small
\begin{tabular}{llcccccc}
\toprule
& & \multicolumn{3}{c}{Gemini-3.7-Flash} & \multicolumn{3}{c}{GLM-5.3-Flash} \\
\cmidrule(lr){3-5} \cmidrule(l){6-8}
Game & Focal demand & Lv.1 & Lv.2 & Lv.3 & Lv.1 & Lv.2 & Lv.3 \\
\midrule
EUP & Resource management          & 0.80 & 0.20 & 0.00 & 0.80 & 0.40 & 0.00 \\
WLE & Spatial \& temporal reasoning & 0.60 & 0.40 & 0.00 & 0.40 & 0.40 & 0.00 \\
RFD & Opponent modeling            & 0.80 & 0.20 & 0.00 & 0.60 & 0.20 & 0.00 \\
AFC & Collaboration                & 1.00 & 0.40 & 0.00 & 0.40 & 0.20 & 0.00 \\
TRA & Adversarial planning         & 0.80 & 0.20 & 0.00 & 0.20 & 0.00 & 0.00 \\
\bottomrule
\end{tabular}
\end{table}

\clearpage


\end{document}